\documentclass[10pt,twocolumn,letterpaper]{article}

\usepackage{cvpr}              

\usepackage{graphicx}
\usepackage{amsmath}
\usepackage{amssymb}
\usepackage{booktabs}

\usepackage[pagebackref,breaklinks,colorlinks]{hyperref}

\usepackage{hyperref}
\usepackage{textcomp}
\usepackage{stfloats}
\usepackage{url}
\usepackage{verbatim}
\usepackage{graphicx}
\usepackage{caption}
\usepackage{cite}
\usepackage{color} 
\usepackage{multirow}
\usepackage{booktabs}
\usepackage{makecell}
\usepackage{xcolor}
\newcommand{\dashuline}[1]{\textbf{#1}}
\renewcommand\fbox{\fcolorbox{red}{white}}

\usepackage[capitalize]{cleveref}
\crefname{section}{Sec.}{Secs.}
\Crefname{section}{Section}{Sections}
\Crefname{table}{Table}{Tables}
\crefname{table}{Tab.}{Tabs.}

\definecolor{black}{rgb}{0,0,0}
\definecolor{blue}{rgb}{0,0,1}

\begin{document}

\title{Parts2Words: Learning Joint Embedding of Point Clouds and Texts by Bidirectional Matching between Parts and Words}

\author{
Chuan~Tang$^1$ \qquad Xi~Yang$^{1,4 *}$ \qquad Bojian~Wu$^2$ \qquad Zhizhong~Han$^3$ \qquad Yi~Chang$^{1, 4}$\thanks{\,Corresponding authors.} \smallskip \\
{\small \textsuperscript{1}School of Artificial Intelligence, Jilin University, China}  \qquad 
{\small \textsuperscript{2}Zhejiang University} \\
{\small \textsuperscript{3}The Department of Computer Science, Wayne State University, USA}\\
{\small \textsuperscript{4}Engineering Research Center of Knowledge-Driven Human-Machine Intelligence, MoE, China
} 
}


\maketitle

\begin{abstract}
    Shape-Text matching is an important task of high-level shape understanding. Current methods mainly represent a 3D shape as multiple 2D rendered views, which obviously can not be understood well due to the structural ambiguity caused by self-occlusion in the limited number of views. To resolve this issue, we directly represent 3D shapes as point clouds, and propose to learn joint embedding of point clouds and texts by bidirectional matching between parts from shapes and words from texts. Specifically, we first segment the point clouds into parts, and then leverage optimal transport method to match parts and words in an optimized feature space, where each part is represented by aggregating features of all points within it and each word is abstracted by its contextual information. We optimize the feature space in order to enlarge the similarities between the paired training samples, while simultaneously maximizing the margin between the unpaired ones. 
    Experiments demonstrate that our method achieves a significant improvement in accuracy over the SOTAs on multi-modal retrieval tasks under the Text2Shape dataset. 
    Codes are available at \href{https://github.com/JLUtangchuan/Parts2Words}{here}.
\end{abstract}


\begin{figure*}
\begin{center}
\centering
  \includegraphics[width=\linewidth]{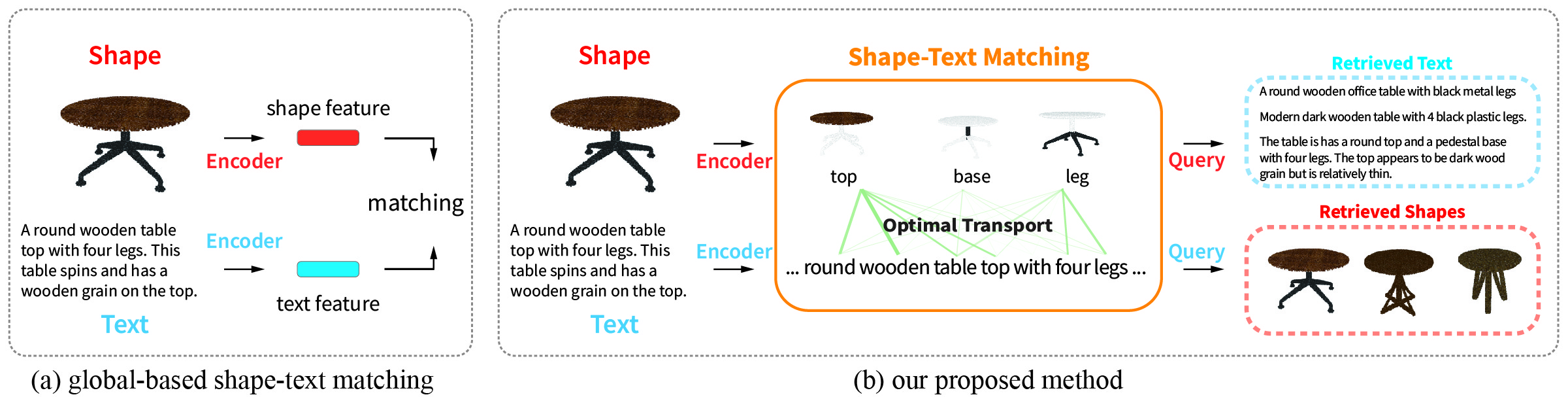}
  \captionof{figure}{Comparison between the global-based matching method and our proposed method. The proposed end-to-end framework aims to learn the joint embedding of point clouds and text by matching parts to words. It can either retrieve shapes using text or vice versa. Our novelty lies in the way of jointly learning embeddings of point clouds and texts.}

  \label{fig:teaser}
\end{center}
\end{figure*}

\vspace{-0.5cm}
\section{Introduction}

Interaction scenarios, such as metaverse, and computer-aided design (CAD), create a larger number of 3D shapes and text descriptions. To enable a more intelligent process of interaction, it is important to bridge the gap between 3D data and linguistic data. Recently, 3D shapes with rich geometric details have been available in large-scale 3D deep learning benchmark datasets\cite{2015ShapeNet,mo2019partnet}. Beyond 3D shapes themselves, text descriptions can also provide additional information. However, it is still hard to jointly understand 3D shapes and texts, since representing different modalities in a common semantic space is still very challenging.

The existing methods aim at learning a joint embedded space to connect various 3D representations with texts, such as voxel grids~\cite{chen2018text2shape} and multi-view rendered images~\cite{han2019y2seq2seq,han2020shapecaptioner}. However, due to the low resolution and self-occlusions, it is hard for those methods mentioned above to improve the ability of joint understanding of shapes and texts. On the other hand, previous shape-text matching methods~\cite{chen2018text2shape,han2019y2seq2seq,TriCoLo} usually take the global features of the entire 3D shape for text matching, making it challenging to capture the local geometries, and thus are not suitable for matching detailed geometric descriptions.

Regional-based matching approaches are commonly employed in the image-text matching task~\cite{lee2018stacked, karpathy2014deep, karpathy2015deep, SMAN}, whereby visual-text alignment is established at the semantic level to enhance the performance of retrieval.
These models compute the local similarities between regions and words and then aggregate the local information to obtain the global metrics between the heterogeneous pairs. However, these two-stage methods based on the pre-trained segmentation networks split the connection between matching embeddings and segmentation prior information.

In this paper, we introduce an optimal transport based shape-text matching method to achieve fine-grained alignment and retrieval of 3D shapes and texts, as shown in Figure~\ref{fig:teaser}. To mitigate the influence of low-resolution or self-occlusions, we directly represent the shape as point clouds and learn a part-level segmentation prior. Afterward, we leverage optimal transport to build the regional cross-modal correspondences and achieve more precise retrieval results. Our main contributions are summarized as follows:

\begin{itemize}
\setlength{\itemsep}{0pt}
\setlength{\parsep}{0pt}
\setlength{\parskip}{0pt}
    \item We propose a novel end-to-end network framework to learn the joint embedding of point clouds and texts, which enables the bidirectional matching between parts from point clouds and words from texts. 

    \item We leverage optimal transport theory to obtain the best matches between parts and words and incorporate Earth Mover's Distance (EMD) to describe the matching score.

    \item To the best of our knowledge, our proposed network achieves SOTA results in joint 3D shape/text understanding tasks in terms of various evaluation metrics.
\end{itemize}

\section{Related Work}

\subsection{Joint embedding of 3D shapes and text}
In recent pioneering work, Chen \textit{et al.}~\cite{chen2018text2shape} introduce a novel 3D-Text cross-modal dataset by annotating each 3D shape from ShapeNet~\cite{2015ShapeNet} with natural language descriptions. In order to understand the inherent connections between text and 3D shapes, they employ CNN+RNN and 3D-CNN to extract features from text and 3D voxelized shapes respectively. They use a full multi-modal loss to learn the joint embedding and calculate the similarity between both modalities. 
Han \textit{et al.}~\cite{han2019y2seq2seq} propose $\rm{Y^{2}Seq2Seq}$, which is a view-based method, to learn cross-modal representations by joint reconstruction and prediction of view and word sequences. Although this method can extract texture information from multiple rendered views by CNN and acquire global shape representation by RNN, it ignores local information aggregation such as part-level features of 3D shapes, which proves to be useful for 3D-Text tasks. To take a step further, ShapeCaptioner~\cite{han2020shapecaptioner} detects shape parts on 2D rendered images, but it is still struggling to fully understand 3D shapes due to the inaccurate boundaries and self-occlusion.
TriCoLo~\cite{TriCoLo} learn the joint embedding space from three modalities by contrastive learning.

In addition, other work is attempting to establish connections between 3D shapes and natural language in other ways.
Liu \textit{et al.}~\cite{ISG} design a new approach for high-fidelity text-guided 3D shape generation.
Text4Point~\cite{huang2023joint} implements implicit alignment between 3D and text modalities using 2D images.
ShapeGlot~\cite{achlioptas2019shapeglot} explores how fine-grained differences between the shapes of common objects are expressed in language, grounded on 2D and/or 3D object representations. They build a dataset of human utterances to develop neural language understanding (listening) and production (speaking) models.
ChangeIt3D~\cite{achlioptas2022changeIt3D} addresses the task of language-assisted 3D shape edits and deformations, which involves modifying or deforming a 3D shape with the assistance of natural language descriptions.
VLGrammar\cite{VLGrammar21} employs compound probabilistic context-free grammars to induce grammars for both image and language within a joint learning framework.
PartGlot\cite{PartGlot22} learns the semantic part segmentation of 3D shape geometry, exclusively relying on part referential language. 

\subsection{Point-based 3D deep learning}
Point clouds have been important representations of 3D shapes due to their simplicity and compactness. PointNet~\cite{qi2017pointnet} and PointNet++~\cite{qi2017pointnet++} are the pioneer works to understand this kind of irregular data. After that, lots of studies~\cite{wu2019pointconv,li2018pointcnn} are proposed to improve the interpretability of network for point clouds in different tasks, such as 3D segmentation~\cite{wenxinacmmm2020,l2g2019,p2seq18}, 3D classification~\cite{wenxinacmmm2020,l2g2019,p2seq18}, 3D reconstruction~\cite{MAPVAE19,InsafutdinovD18,Groueix_2018_CVPR,handrwr2020} and 3D completion~\cite{Hu20193DSC,Hu2019Render4CompletionSM,wenxin_2020_CVPR}.

\subsection{Image-text matching}
The image-text matching task allows the image or text to mutually find the most relevant instance from the multi-modal database. Most existing methods can be roughly categorized into two types: global matching methods and regional matching methods.

\noindent\textbf{Global matching methods. } m-RNN~\cite{DBLP:journals/corr/MaoXYWY14a} aims to extract the global representation from both images and texts and then calculate the similarity score. VSE~\cite{kiros2014unifying} learns to map images and text to the same embedding space by optimizing a pairwise ranking loss. VSE++~\cite{DBLP:conf/bmvc/FaghriFKF18} tries to improve the performance by exploiting the hard negative mining strategy during training.

\noindent\textbf{Regional image-text matching. } These methods extract image region representation from existing detectors and then take latent visual-semantic correspondence at the level of image regions and words into consideration. DeFrag~\cite{karpathy2014deep} and DVSA~\cite{karpathy2015deep} propose visual semantic matching by inferring their inter-modal alignment, these methods first detect object regions and then acquire the region-word correspondence, finally aggregate the similarity of all possible pairs of image regions and words in the sentence to infer the global image-text similarity. Inspired by Up-Down~\cite{DBLP:conf/cvpr/00010BT0GZ18}, SCAN ~\cite{lee2018stacked} takes a step towards attending to important image regions and words with each other as context for inferring the image-text similarity. Recently, some works SMAN~\cite{SMAN}, CASC~\cite{CASC}, R-SCAN~\cite{R-SCAN}, RDAN~\cite{DBLP:conf/ijcai/HuLLYC19}, DP-RNN~\cite{DBLP:conf/aaai/ChenL20a}, PFAN~\cite{DBLP:conf/ijcai/WangYQMLLF19} attempt to improve SCAN and try to achieve better performance.

\begin{figure*}[t]
    \begin{center}
    \includegraphics[width=1.0\linewidth]{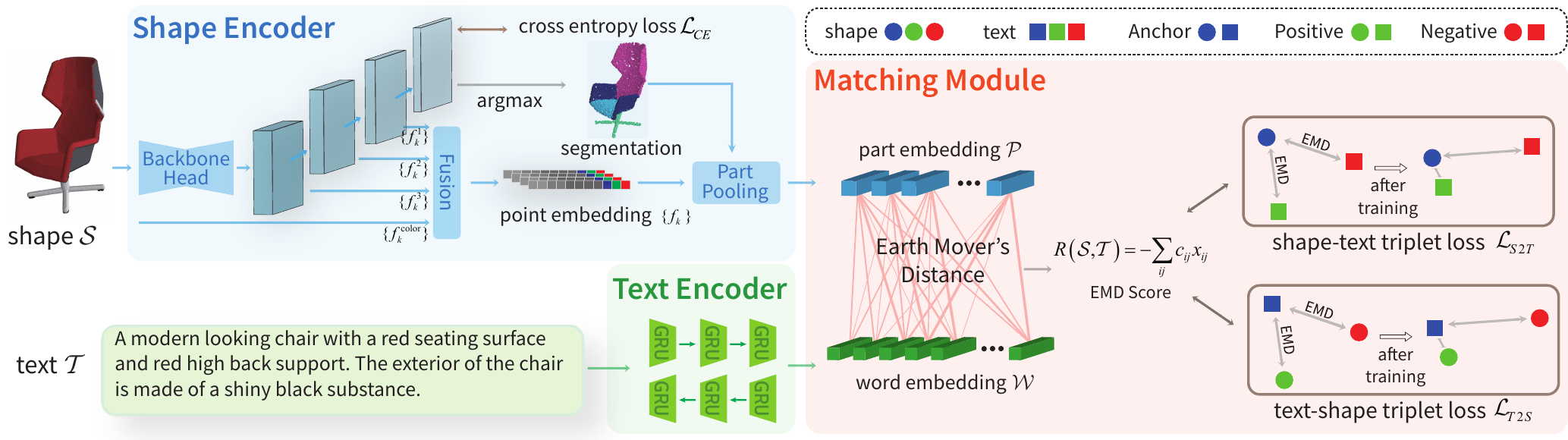}
    \end{center}
    \vspace{-0.4cm}
    \caption{Overview. The proposed network includes three modules: shape encoder, text encoder, and matching module. The shape encoder learns the part embedding from the input 3D shape, and the text encoder learns the word embedding from the corresponding text description. Then we utilize Earth Mover's Distance to measure the discrepancy between parts and words in order to obtain the best matches. To achieve the goal, we also leverage triplet loss to enhance the similarity between the paired training samples, and push the unmatched ones far apart. }
    \label{fig:overview}
\end{figure*}

\section{Our Method}

\noindent\textbf{Overview} As shown in Figure~\ref{fig:overview}, our proposed network includes three modules: a shape encoder, a text encoder, and a matching module. To encode a 3D shape $\mathcal{S}$, we first uniformly sample points and then use a point-based backbone network that aims to predict point labels and concatenate hierarchical hidden features to acquire the representation of each point. Then, we aggregate these representations in the same part to extract its embedding $\mathcal{P} \in \{p_i|i\in[1,n]\}$ of the input shape $\mathcal{S}$ with $n$ parts. For the text encoder, we use the Bi-directional Gate Recurrent Unit (GRU) to learn context-sensitive embedding $\mathcal{W} = \{w_j|j\in[1,m]\}$ of each word in the text $\mathcal{T}$, where the text $\mathcal{T}$ has $m$ words.

To achieve the matching between $\mathcal{P}$ and $\mathcal{W}$, 
we elaborate an optimal transport-based matching module to generate the similarity score between $\mathcal{S}$ and $\mathcal{T}$. Finally, we use both segmentation loss and matching loss to train our model.

\subsection{The Shape Encoder}

We use 3D point cloud data as visual input and use a semantic segmentation network as the shape encoder. 
Our shape encoder extracts the embedding of parts with different semantic types on each input shape by aggregating the features of corresponding points in the segmented parts. We feed a shape $\mathcal{S}$ to the segmentation backbone (using PointNet~\cite{qi2017pointnet} or PointNet++~\cite{qi2017pointnet++}) to acquire the semantic prediction for part assignment $\{a_i|i\in[1,n]\}$ and extract the point-wise features $f_k, k\in[1,l]$, where $l$ is the number of points in $\mathcal{S}$.
We then aggregate these point-wise features into embedding of parts, which can be instantly sent into the matching module.
The outputs $f^1_k, f^2_k, f^3_k, k\in[1,l]$ of the last three layers of the backbone are extracted to form the point features $f_k$. Besides, we also explicitly utilize the color representation $f^\mathrm{color}_k, k\in[1,l]$ of the input shape for better performance, our combined feature fusion module makes full use of the color information and semantic information of 3D shape at the same time. Thus, the point-wise feature representation $f_k$ is computed by Equation~(\ref{equ:point feature}):

\begin{equation}
\label{equ:point feature}
f_k=\operatorname{fc}(\operatorname{fc}(f^1_k)+\operatorname{fc}(f^2_k)+\operatorname{fc}(f^3_k) \oplus \operatorname{mlp}(f^\mathrm{color}_k)), 
\end{equation}
where $\operatorname{mlp}$ stacks multiple fully connection layers $\operatorname{fc}$ to perform nonlinear mapping on color feature $f^\mathrm{color}_k$. Then, $\operatorname{pooling}$ is applied to the point-wise features within the same semantic part, in order to extract the embedding of each part $p_i$, as shown in Equation~(\ref{equ:gap}).

\begin{equation}
p_i=\underset{k \in a_i}{pooling}(f_k), i=1, \ldots, n
\label{equ:gap}
\end{equation}

\subsection{The Text Encoder}
The text encoder aims to extract local features at the word level, as shown in Equation~(\ref{equ:GRU}), we use Bi-directional GRU to extract the context-sensitive word embedding $\mathcal{W}$. Each text description $\mathcal{T}$ is firstly represented by the embedding of every single word $e_{j}$ in the text through a word embedding layer. Then, we encode the context of $w_{j}$ in the bi-directional GRU. 
For the forward $\overrightarrow{G R U}$, the hidden state $\overrightarrow{h_j}$ can be calculated from the embedding $e_j$ of the current word at each time-step, and the hidden state $h_{j-1}$ from the previous time-step. Similarly, for the backward $\overleftarrow{G R U}$, the current hidden state $\overleftarrow{h_j}$ is calculated from the embedding $e_j$ of the current word and the hidden state $h_{j+1}$ from the next. 
Finally, the context-sensitive word embedding is obtained by the averages of the hidden states in the two directions.

\begin{equation}
\label{equ:GRU}
\begin{split}
&\overrightarrow{h_{j}}=\overrightarrow{G R U}\left(e_{j}, h_{j-1}\right), j \in[1, m] 
\\
&\overleftarrow{h_{j}}=\overleftarrow{G R U}\left(e_{j}, h_{j+1}\right), j \in[1, m] 
\\
&w_{j}=\frac{\overrightarrow{h_{j}}+\overleftarrow{h_{j}}}{2}, j \in[1, m]
\end{split}
\end{equation}

\subsection{The Matching Module}

We introduce the optimal transport method to evaluate the similarity of 3D shape and text using their embedding $\mathcal{P}$ and $\mathcal{W}$. Similar to the transport of goods between producers (part $p_i$) and consumers (word $w_j$), $p_i$ contains $u_i$ stock of goods and $w_j$ has $v_j$ capacity. Then, the transport cost between $p_i$ and $w_j$ is defined as $c_{ij}$, and the matching flow is defined as $x_{ij}$. 
We formulate our matching problem in Equation~(\ref{equ:emd}):
\begin{equation}
\label{equ:emd}
\begin{array}{ll}\underset{x_{i j}}{\operatorname{min}} & \sum_{i=1}^{n} \sum_{j=1}^{m} c_{i j} x_{i j} \\ 
\text { s.t. } & x_{i j} \ge 0, i=1, \ldots, n, j=1, \ldots, m \\ 
& \sum_{j=1}^{n} x_{i j}=u_i, \quad i=1, \ldots n \\ 
& \sum_{i=1}^{m} x_{i j}=v_j, \quad j=1, \ldots m \\ ~ \\
&  c_{i j}=1-\frac{{p_i}^T w_j}{\left\|p_i\right\|\left\|w_j\right\|}
\end{array}
\end{equation}
where the cost $c_{ij}$ between $p_i$ and $w_j$ is measured by cosine distance. 
We adopt a simple discrete uniform distribution for the EMD node weight settings of $u_i$ and $v_j$, which constrains the upper limit of the summation of matching flow.

Then, we use the Sinkhorn algorithm to compute the optimal matching flow $\hat{x}_{i j}$ which is the solution of the earth mover's distance function in Equation~(\ref{equ:emd}). After optimization, we calculate the similarity $R_{EMD}$ between shape $\mathcal{S}$ and text $\mathcal{T}$ as shown by Equation~(\ref{equ:score})
\begin{equation}
\label{equ:score}
R_{EMD}(\mathcal{S},\mathcal{T})=-\sum_{i=1}^{n} \sum_{j=1}^{m} c_{ij} \hat{x}_{ij}
\end{equation}

\subsection{Objective Function}

As shown in Equation~(\ref{equ:total loss}), 
we train the proposed network using the multi-task learning strategy, and learn the part segmentation and matching simultaneously.
The weight $\beta$ is employed to balance the two tasks.

The segmentation network is optimized by the cross-entropy loss $L_{CE}$ only. While, for the matching task, we adopt the paired ranking loss with the semi-hard negative sampling mining strategy~\cite{DBLP:conf/cvpr/SchroffKP15} to facilitate the network to better converge and avoid getting into a collapsed model, as shown in Equation~(\ref{equ:loss}). Specifically, for a positive sampling pair $(\mathcal{S},\mathcal{T})$, we select the semi-hard negative sampling pair $(\hat{\mathcal{S}}_{semi}, \hat{\mathcal{T}}_{semi})$ which has a smaller similarity score than $(\mathcal{S},\mathcal{T})$, and calculate the triplet loss for the input shape and text respectively. Similarly, the triplet loss between the sampling pair $(\mathcal{T},\mathcal{S})$ can also be calculated in the same way. 

\begin{equation}
  \label{equ:total loss}
  L=L_{CE}+\beta L_{EMD}
\end{equation}
\begin{gather}
\begin{aligned}
L_{EMD} &= L_{S2T}(\mathcal{S},\mathcal{T}) + L_{T2S}(\mathcal{T},\mathcal{S}), 
\\
L_{S2T}(\mathcal{S}{,}\mathcal{T}){ =} &max(\alpha{-}R_{EMD}(\mathcal{S}{,}\mathcal{T}){ + }R_{EMD}(\mathcal{S}{,} \hat{\mathcal{T}}_{semi}){,}0) 
\\
L_{T2S}(\mathcal{T}{,}\mathcal{S}){ =} &max(\alpha{-}R_{EMD}(\mathcal{T}{,}\mathcal{S}){ + }R_{EMD}(\mathcal{T}{,} \hat{\mathcal{S}}_{semi}){,}0)
\end{aligned}
\label{equ:loss}
\raisetag{45pt}
\end{gather}

Here, $\alpha$ is a margin that is enforced between the positive and negative pairs.

\section{Experiments}

\begin{table}
\centering
\caption{Retrieval results on Text2shape compared to the SOTAs.}
\label{tab:cmp_sota}

\centering
\resizebox{0.98\columnwidth}{!}{%
\begin{tabular}{lcccccc}
\toprule
\multicolumn{1}{c}{\multirow{2}{*}{Method}} & \multicolumn{3}{c}{S2T} & \multicolumn{3}{c}{T2S} \\
\cmidrule{2-7}
& RR@1 & RR@5 & NDCG@5 & RR@1 & RR@5 & NDCG@5 \\

\midrule
\textbf{Text2Shape}~\cite{chen2018text2shape} & 0.83 & 3.37 & 0.73 & 0.40 & 2.37 & 1.35 \\
$\mathbf{Y^{2}Seq2Seq}$~\cite{han2019y2seq2seq} & 6.77 & 19.30 & 5.30 & 2.93 & 9.23 & 6.05 \\
\textbf{TriCoLo}~\cite{TriCoLo} & 16.33 & 45.52 & 12.73 & 10.25 & 29.07 & 19.85 \\
\midrule
\textbf{Global-Max} (Our) & 12.60 & 32.96 & 9.48 & 8.53 & 24.09 & 16.46 \\
\textbf{Global-Avg} (Our) & 7.63 & 24.07 & 6.23 & 8.60 & 24.82  & 16.83 \\
\textbf{LocalBaseline}  (Our) & 12.81 & 34.71 & 9.82 & 7.87 & 23.55 & 15.84 \\
\textbf{LocalBaseline+} (Our) & 17.77 & 44.58 & 13.91 & 11.94 & 31.62 & 21.92 \\
\midrule
\textbf{Parts2Words (Our)} & \textbf{19.38} & \textbf{47.17} & \textbf{15.30}  & \textbf{12.72} & \textbf{32.98} & \textbf{23.13} \\ 
\bottomrule
\end{tabular}}
\end{table}

\begin{figure}[t]
    \includegraphics[width=\linewidth]{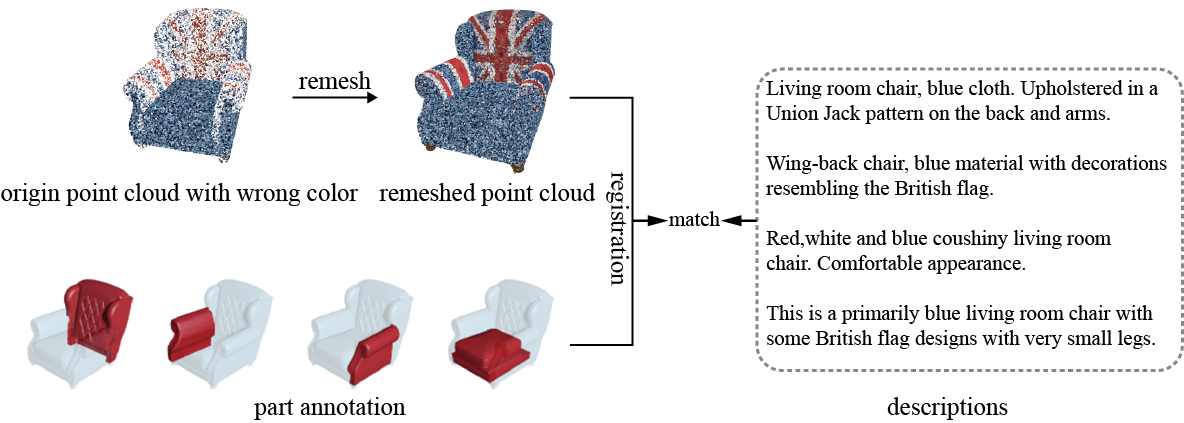}
    \caption{Data pre-process. We rectify the incorrect color input provided by ShapeNet~\cite{2015ShapeNet} and add the part segmentation annotation from PartNet~\cite{mo2019partnet} by registration. Finally, the new segmentation dataset will be combined with the Text2Shape~\cite{chen2018text2shape}.}
    \label{fig:data}
\end{figure}

\begin{figure}
\setkeys{Gin}{width=\linewidth}
\newcommand{\shape}[1]{\begin{minipage}{0.08\textwidth} \includegraphics[]{#1} \end{minipage}}
\newcommand{\selectedshape}[1]{\begin{minipage}{0.08\textwidth} \fbox{ \includegraphics[]{#1}} \end{minipage}}
\newcommand{\textbox}[1]{\begin{minipage}{6cm} \footnotesize{#1} \end{minipage}}
\centering
\resizebox{0.98\columnwidth}{!}{
\begin{tabular}{p{6cm}ccccc}
    \toprule
    Query text & top1 & top2 & top3 & top4 & top5 \\
    \midrule
    \textbox{\normalsize{This \dashuline{glass} table is excellent, beautiful if displayed at living room.}} &
    \selectedshape{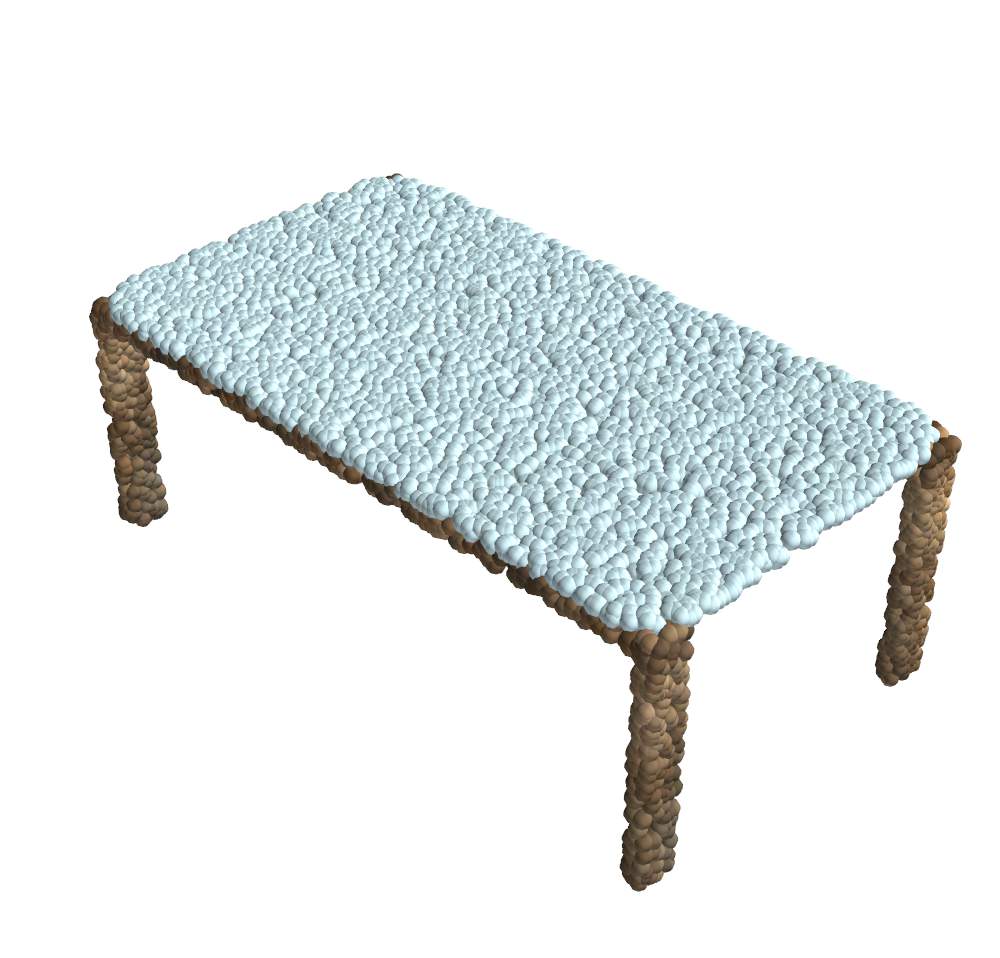} &
    \shape{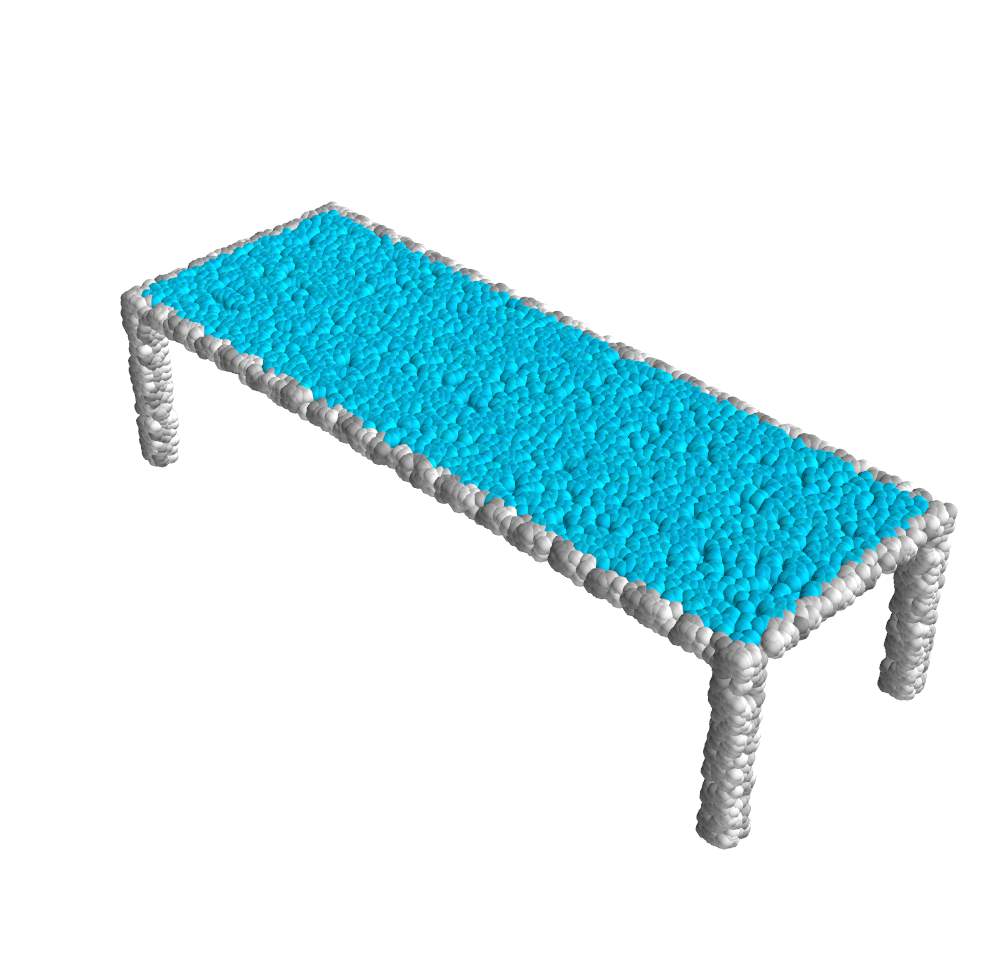} &
    \shape{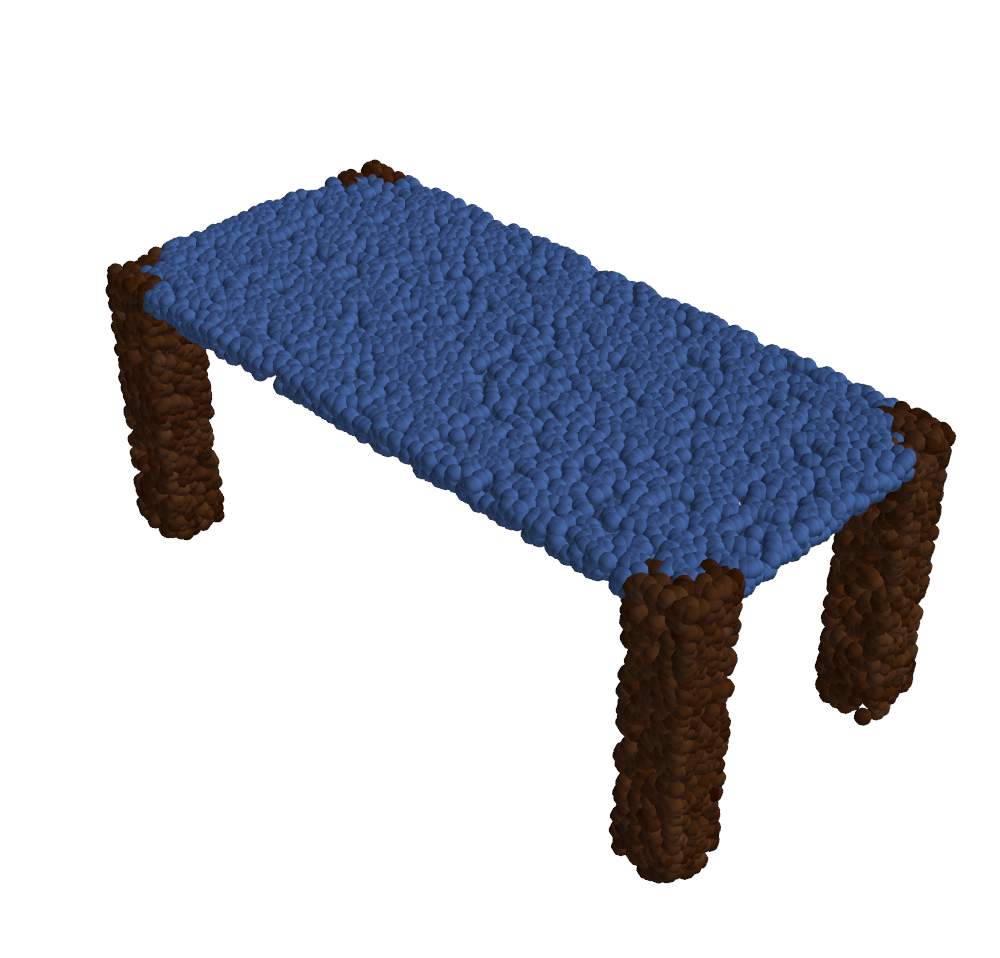} &
    \shape{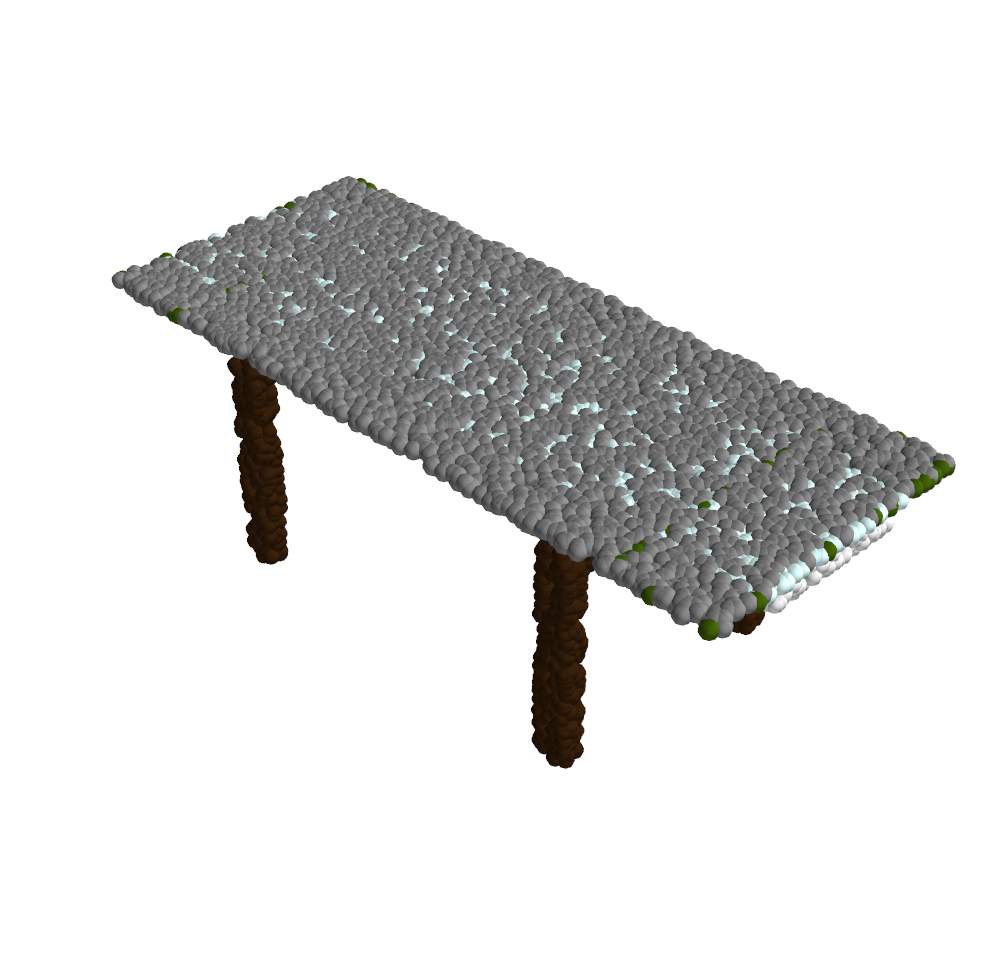} &
    \shape{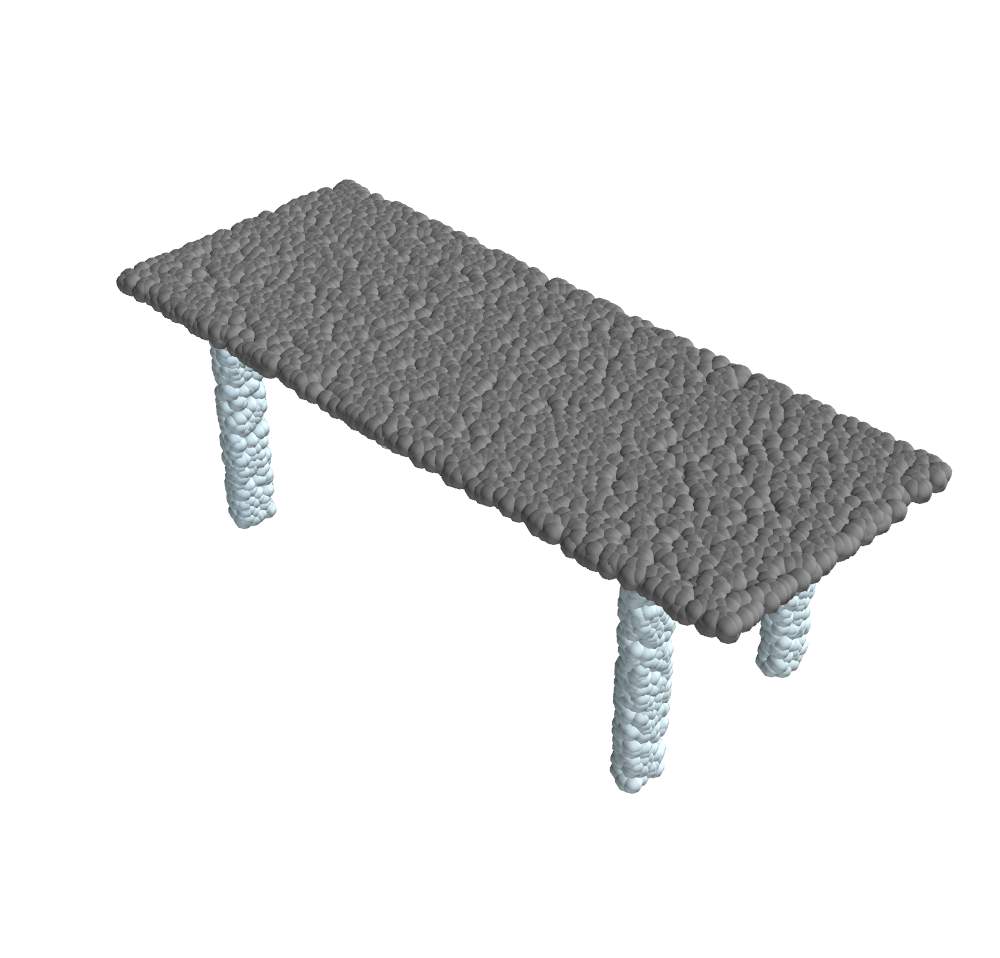} \\
    
    \midrule
    \textbox{\normalsize{it is \dashuline{long} chair. it can be able to sit comfortable. it has \dashuline{white} in color.}} &
    \shape{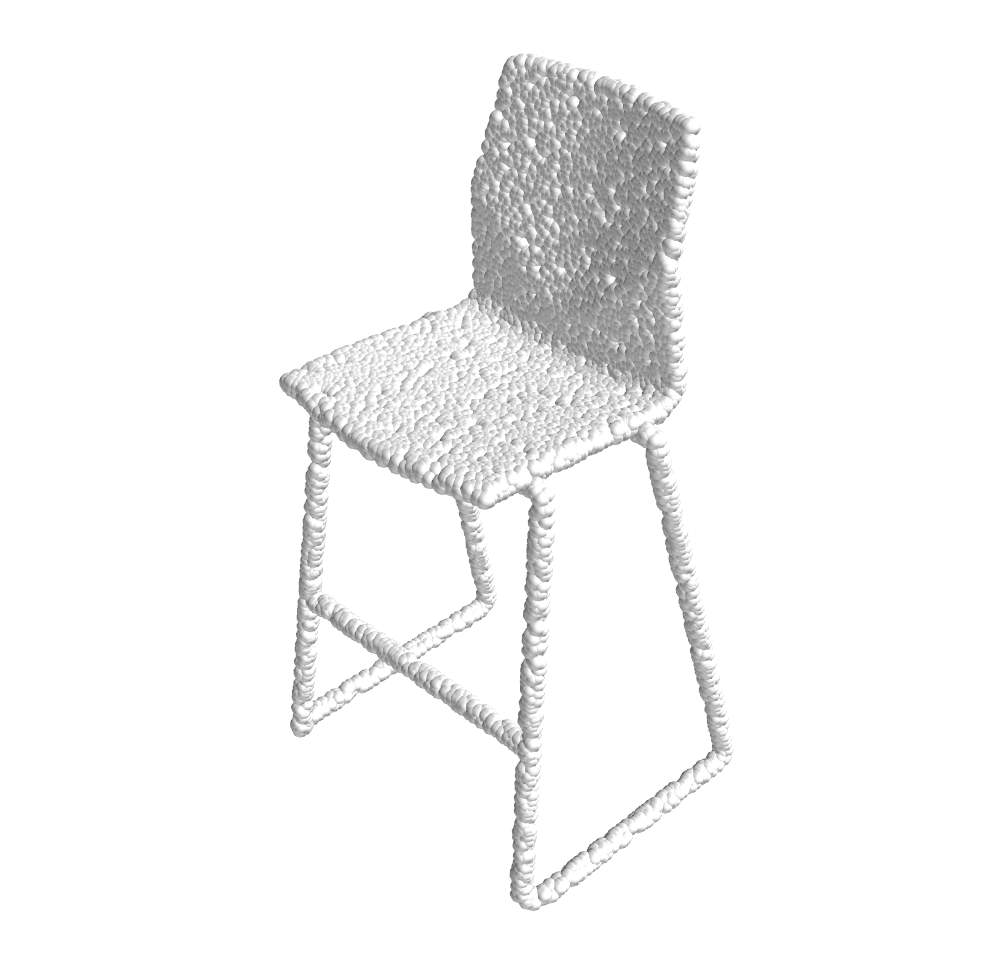} &
    \shape{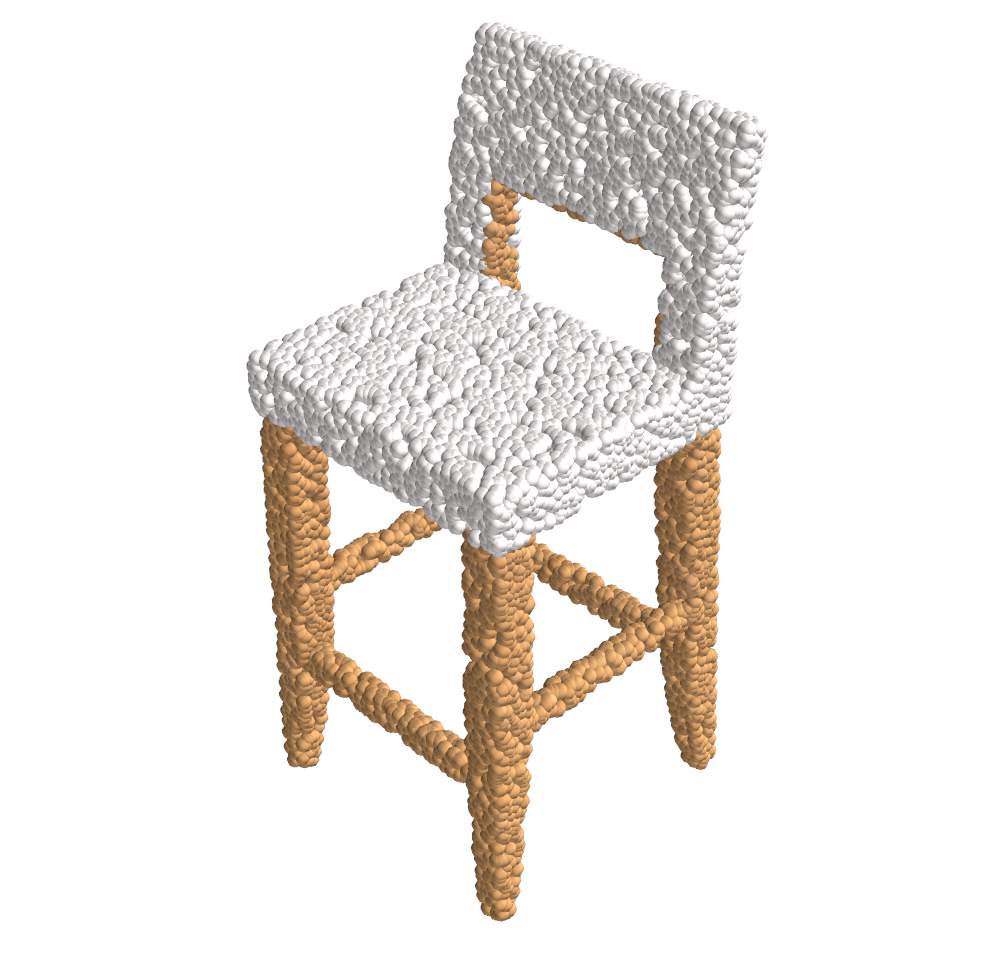} &
    \selectedshape{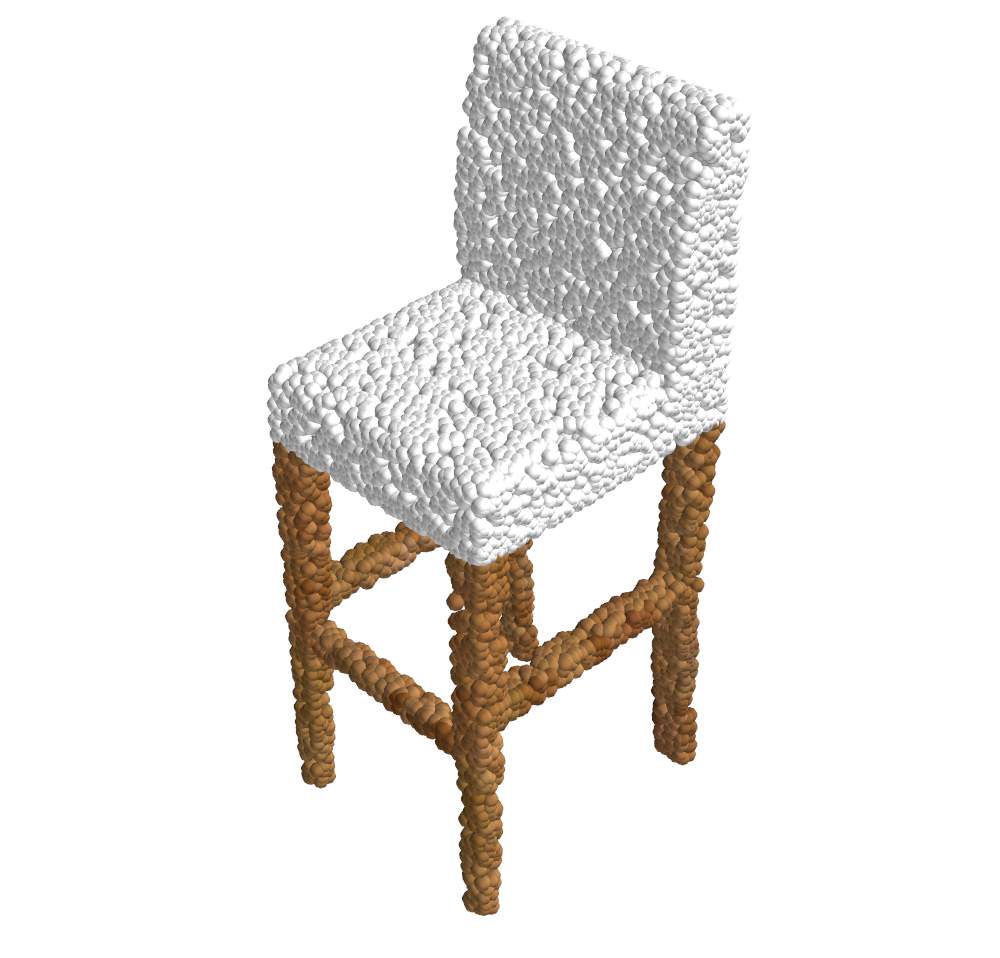} &
    \shape{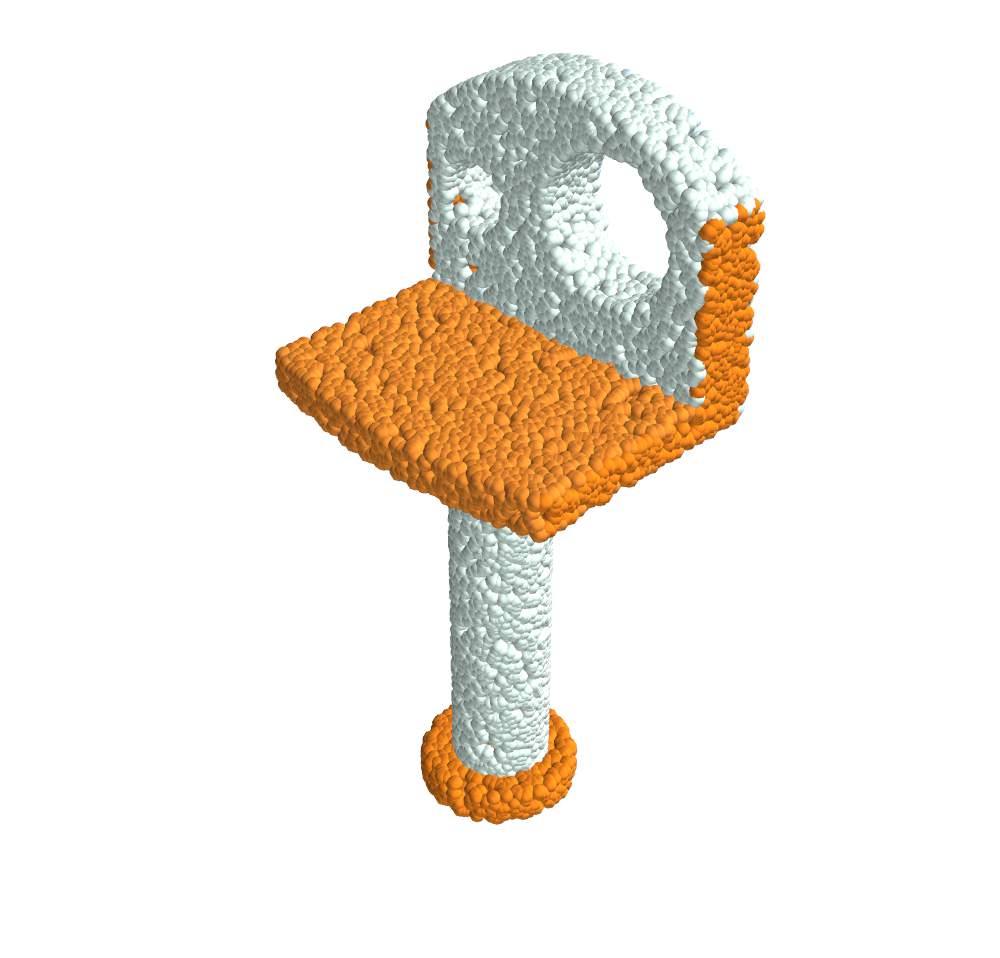} &
    \shape{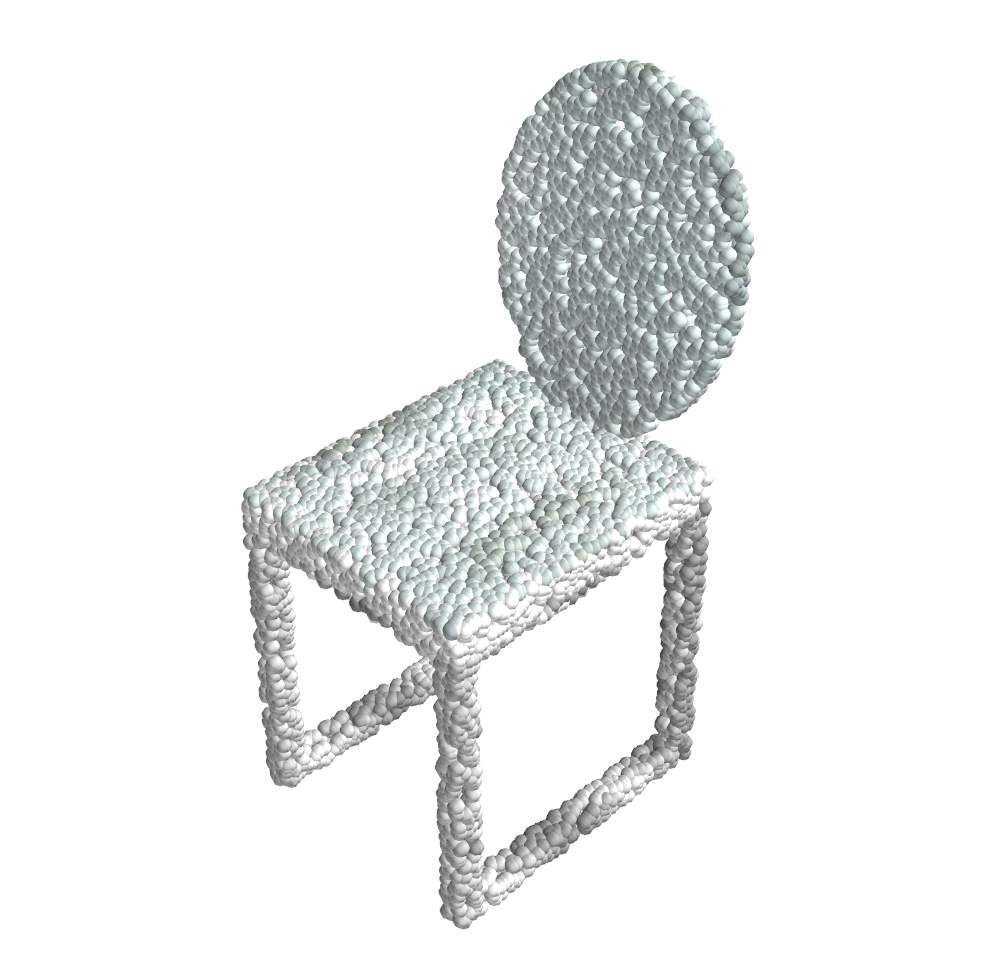} \\
    
    \midrule
    \textbox{\normalsize{chair with \dashuline{brown wooden structure, armrests} and both seat and back cover with \dashuline{red} and pastel fabric}} &
    \shape{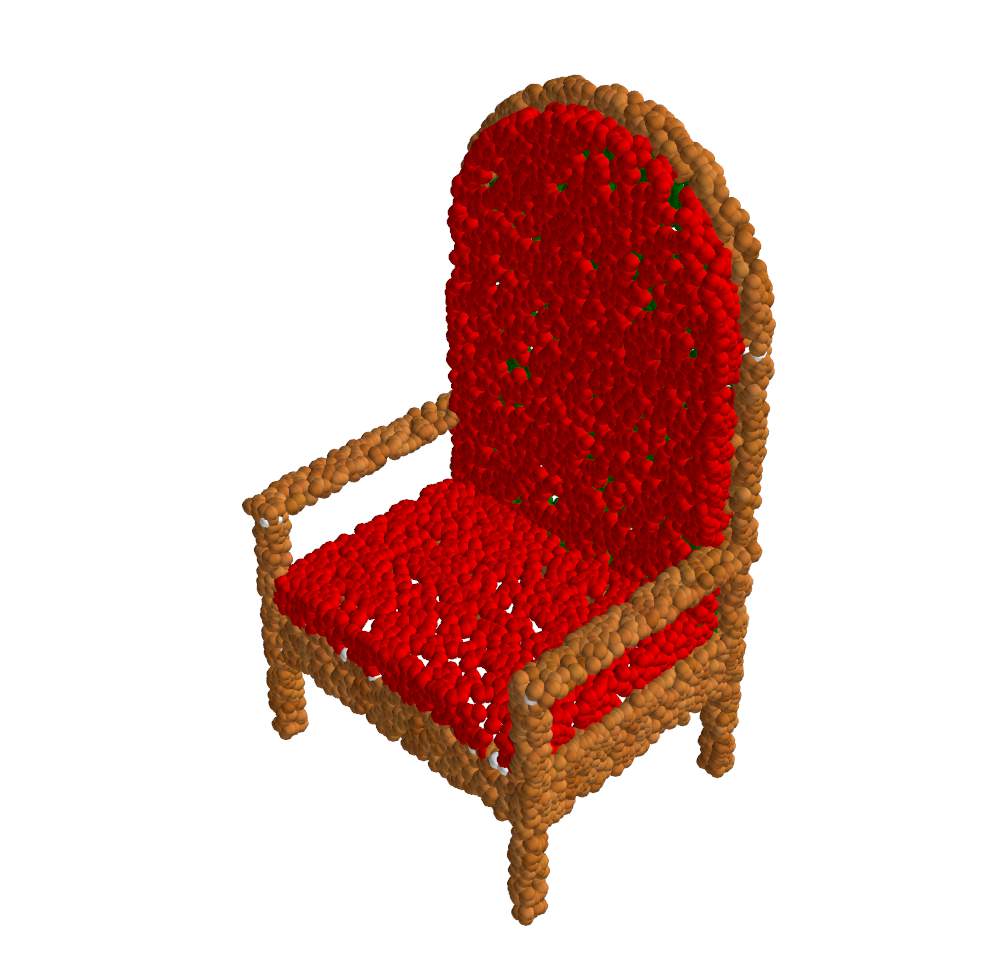} &
    \selectedshape{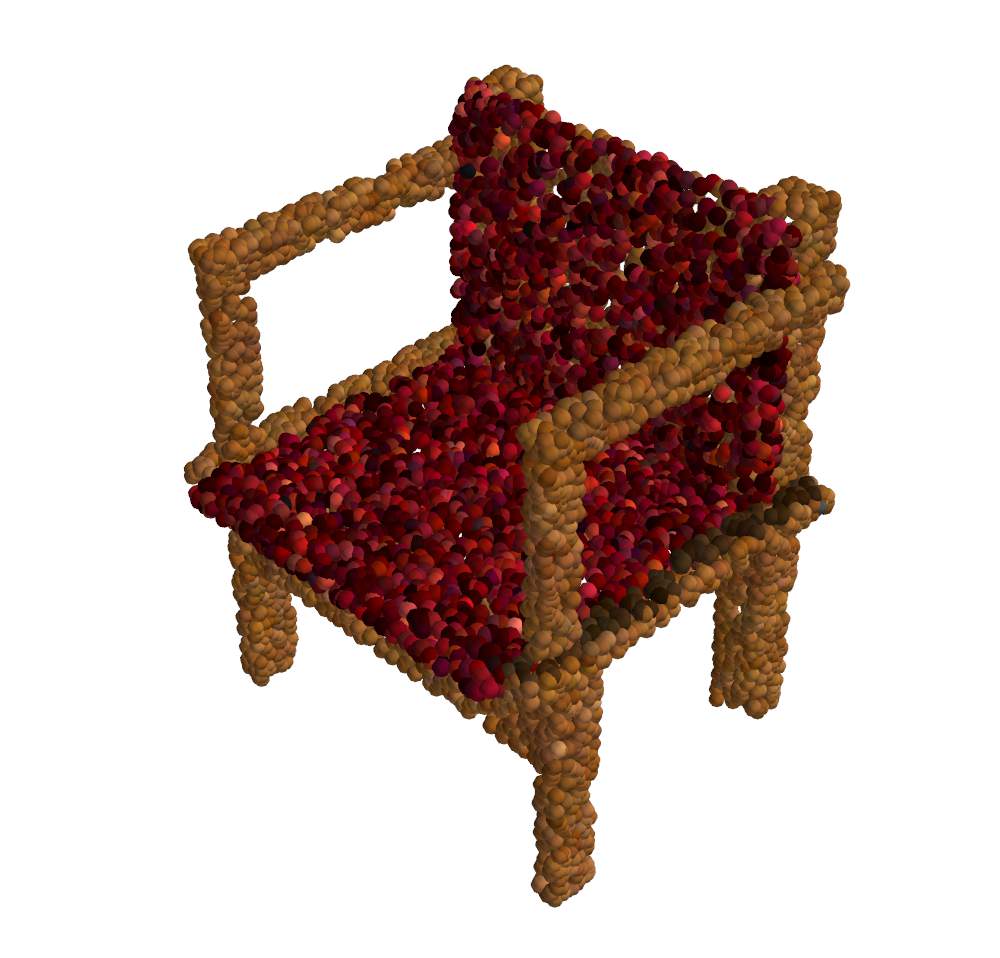} &
    \shape{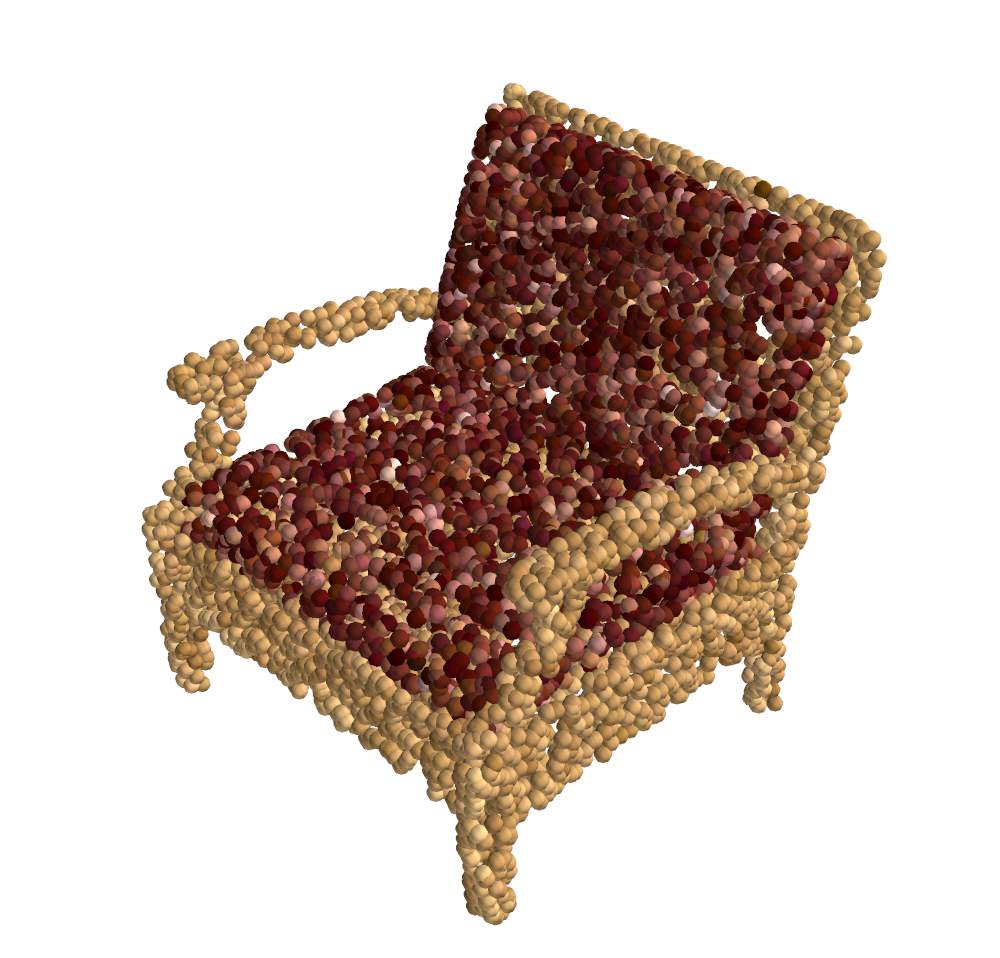} &
    \shape{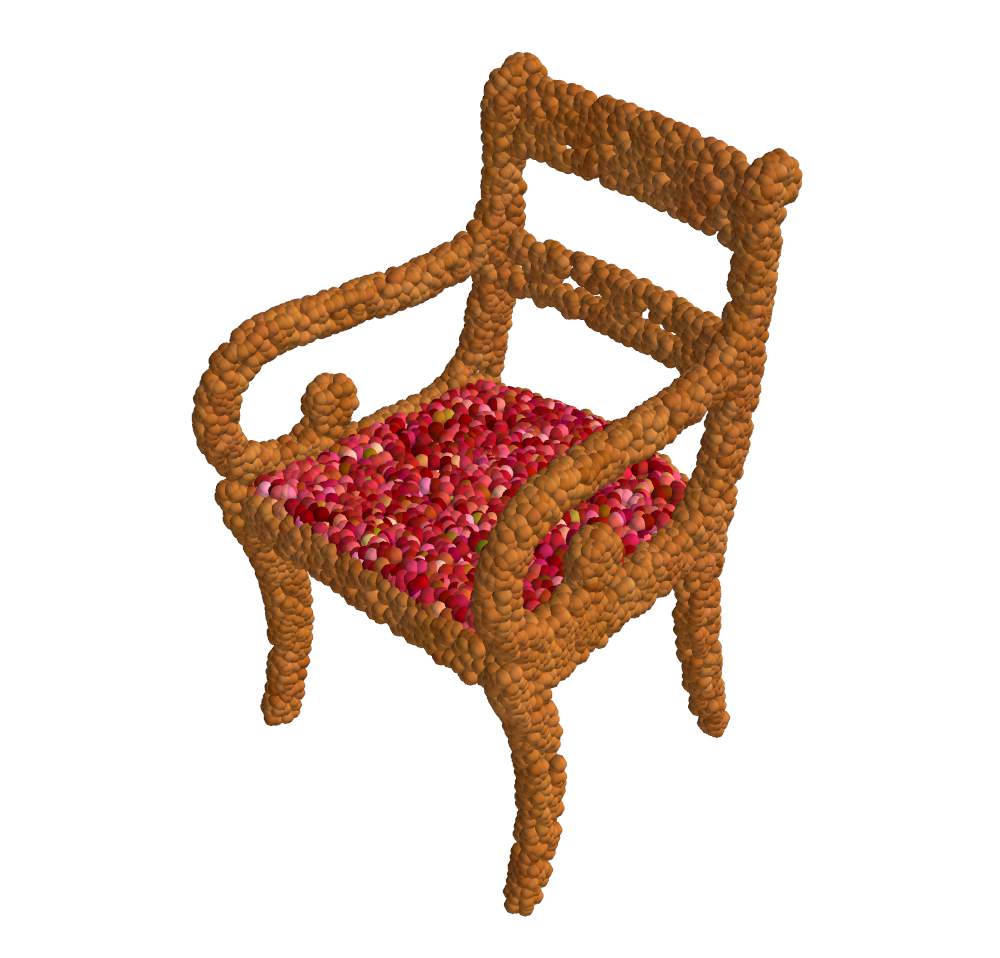} &
    \shape{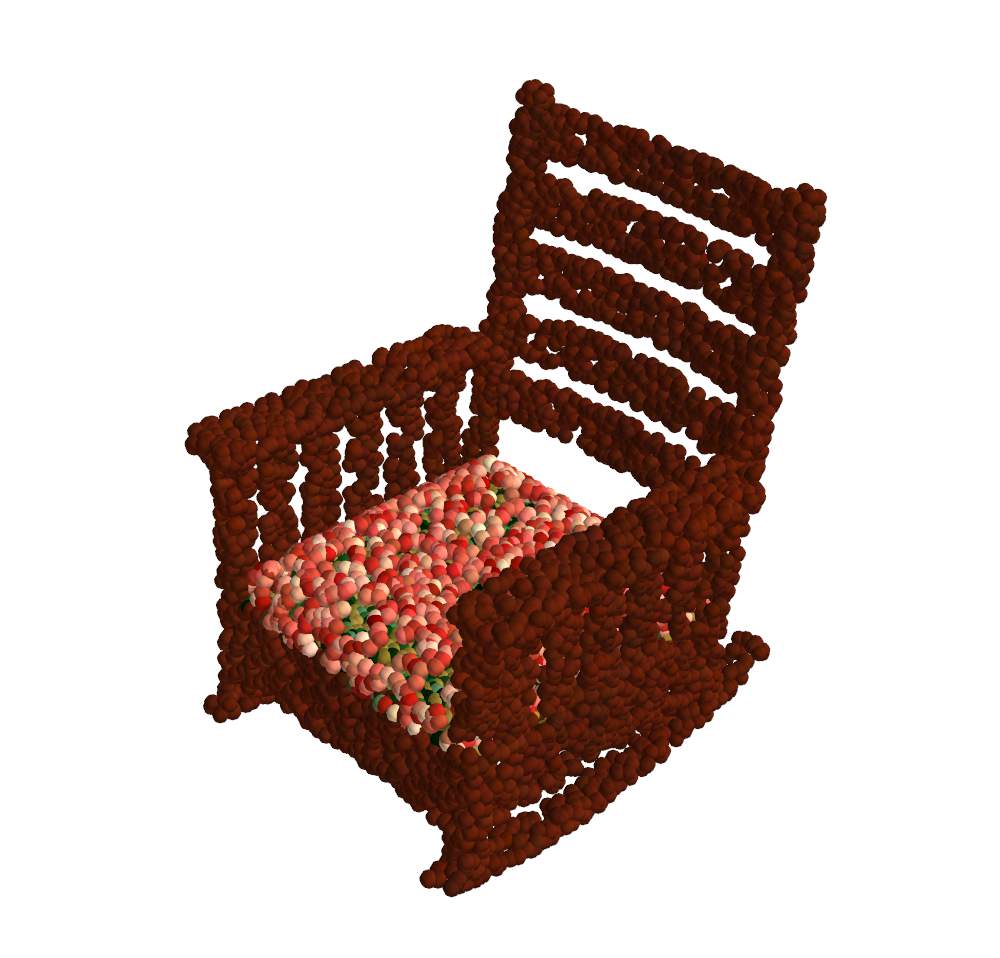} \\
    
    \midrule
    \textbox{\normalsize{A \dashuline{wooden} table with \dashuline{glass parts in the center}.There is also storage facility to keep any material below the table.It is \dashuline{rectangular} in shape with curves in the four ends .}} &
    \shape{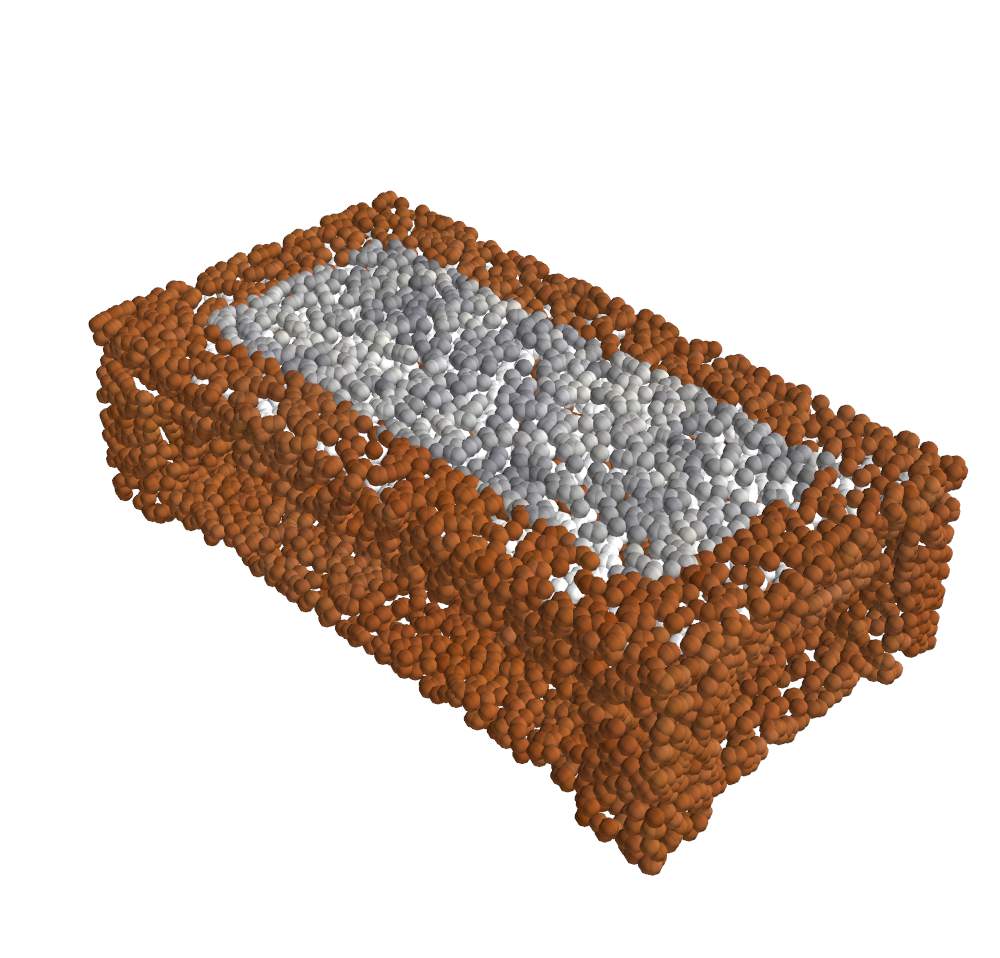} &
    \selectedshape{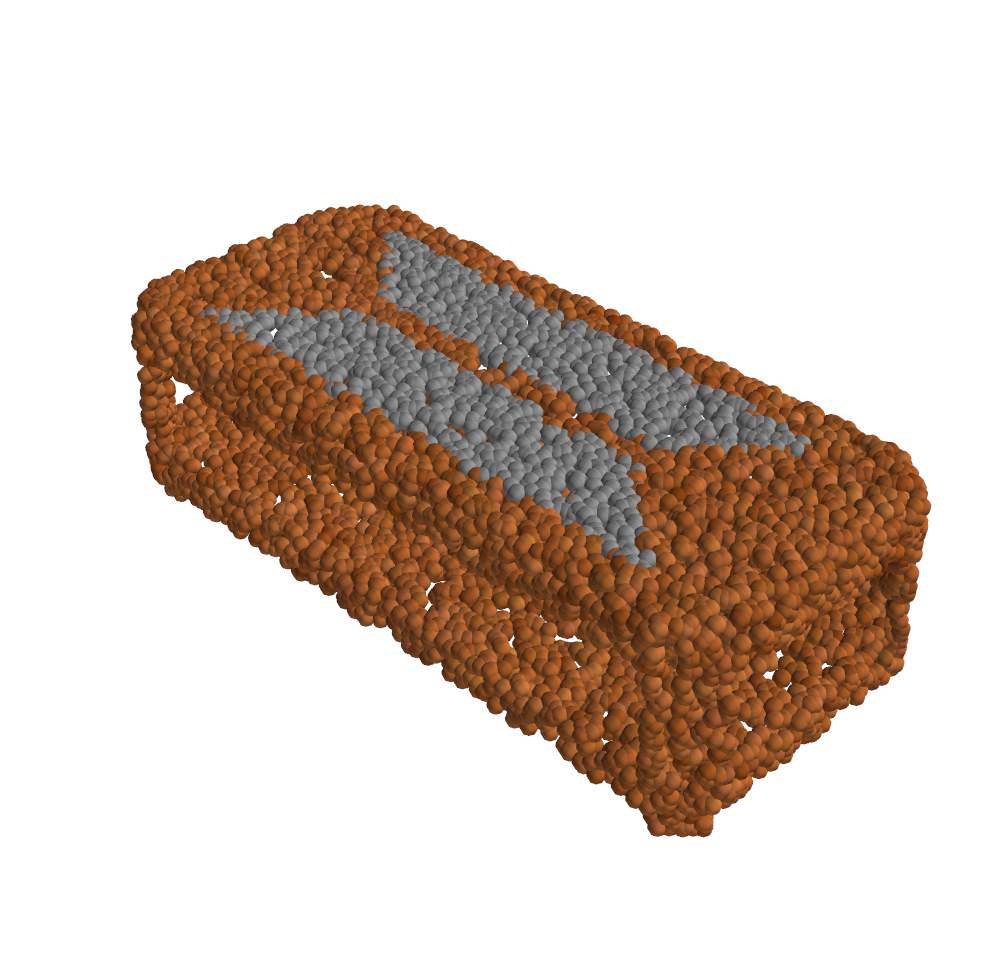} &
    \shape{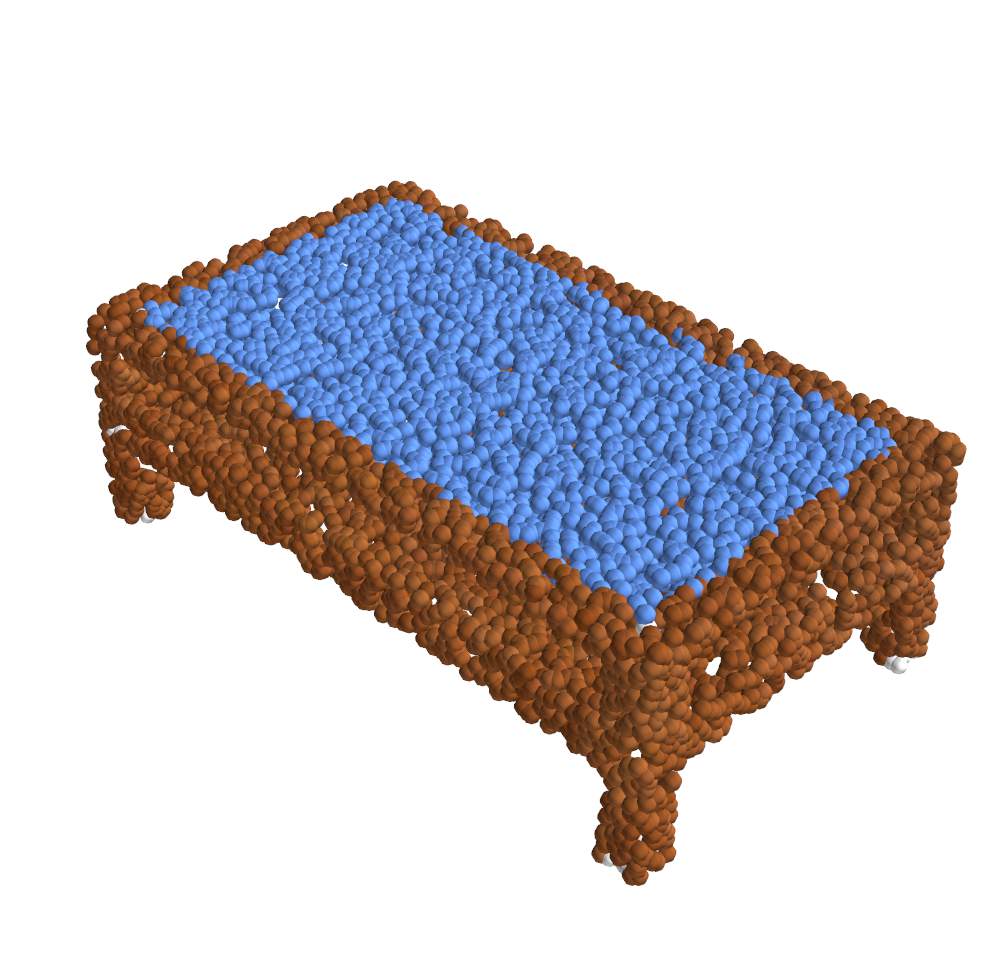} &
    \shape{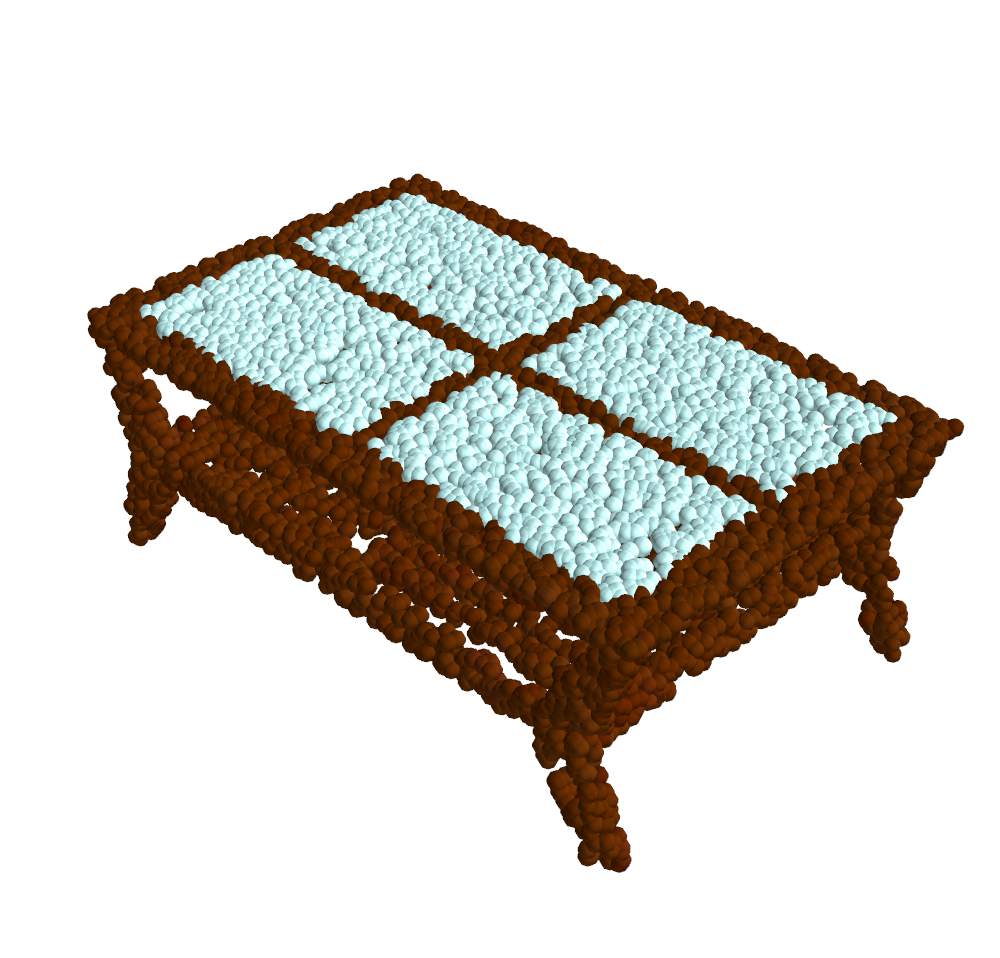} &
    \shape{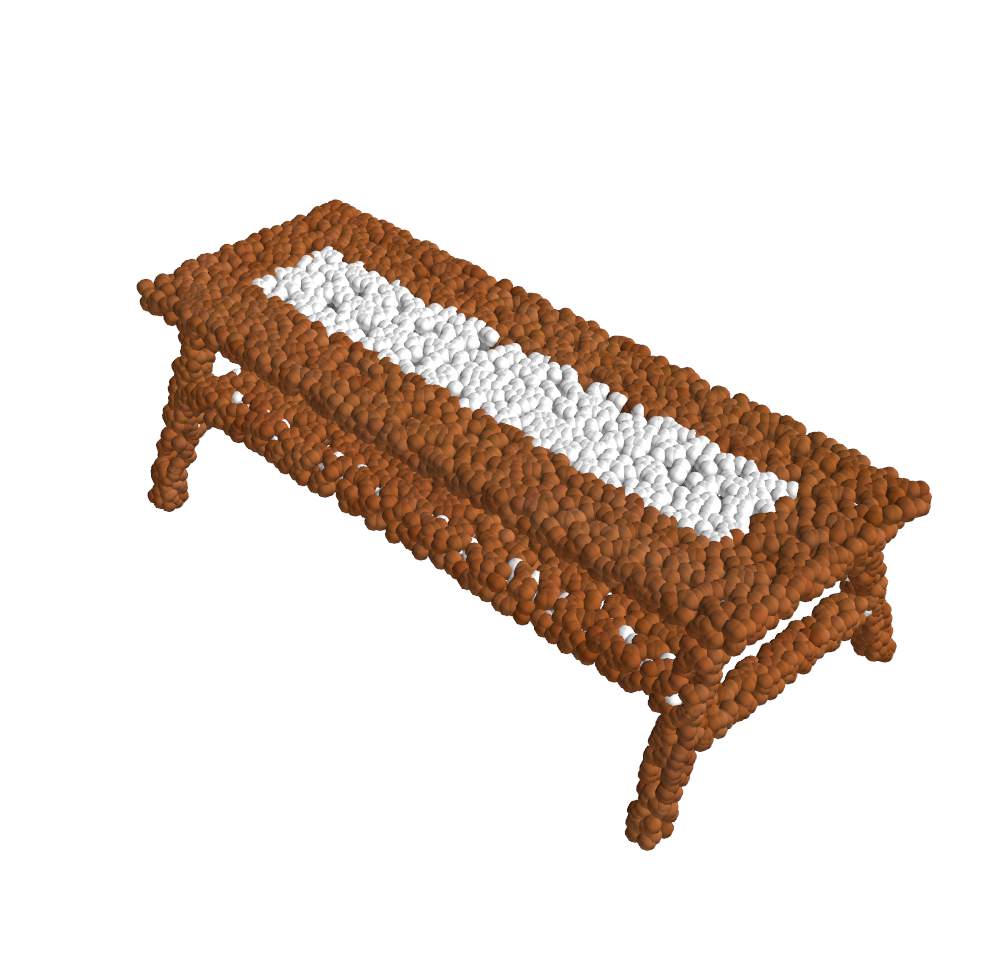} 
    \\
    
    \midrule
    \textbox{\normalsize{A modern \dashuline{blue} colored chair with \dashuline{no armrest}.}} &
    \selectedshape{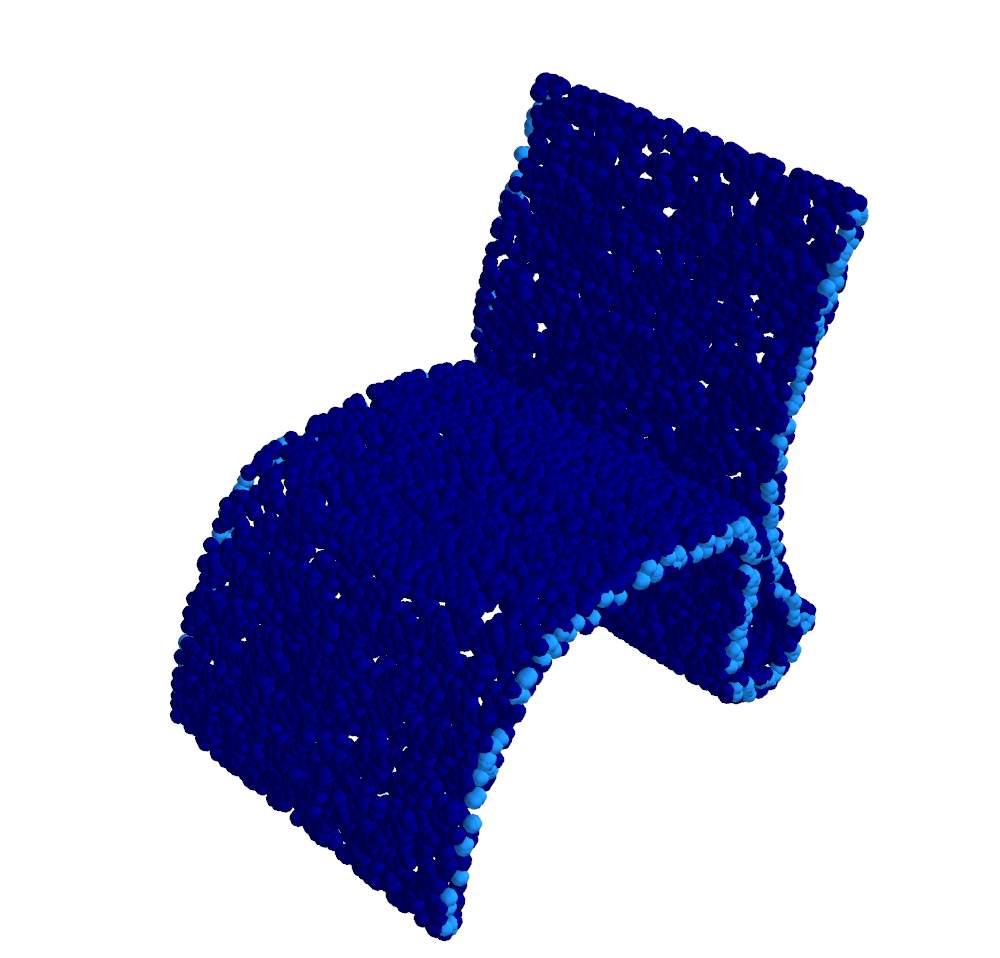} &
    \shape{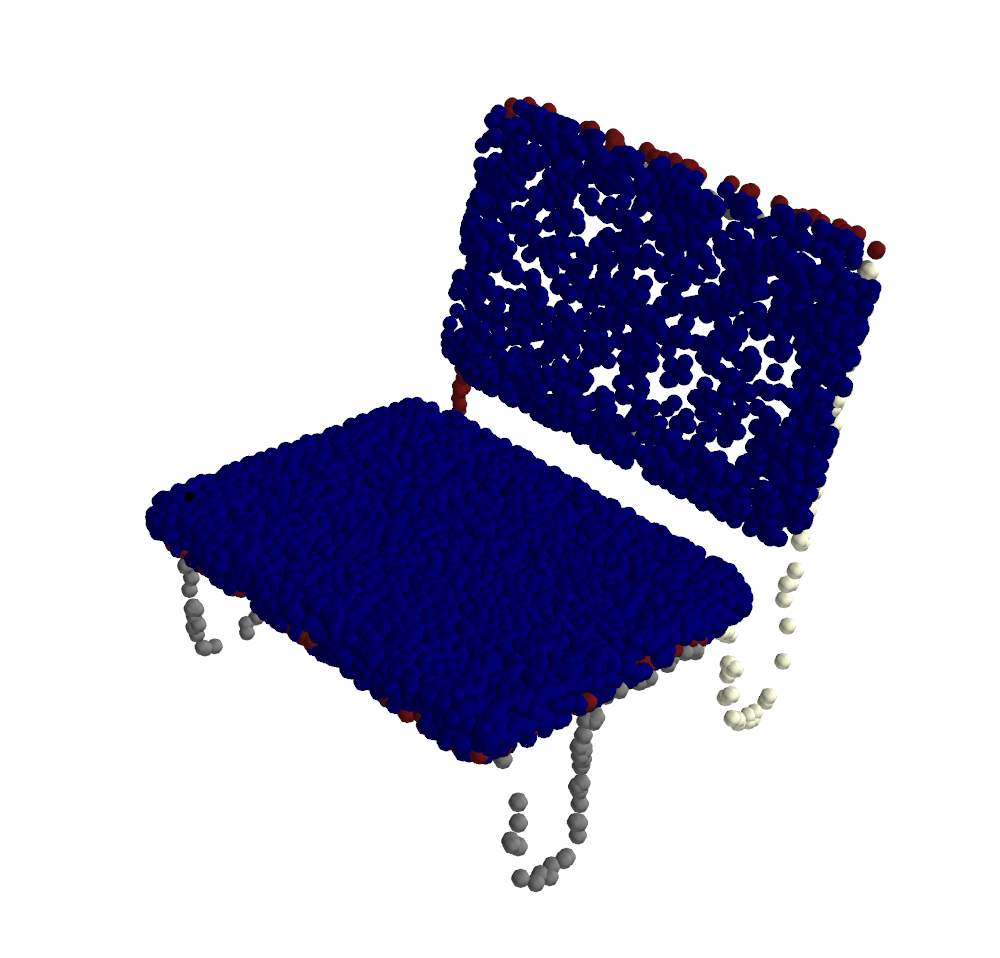} &
    \shape{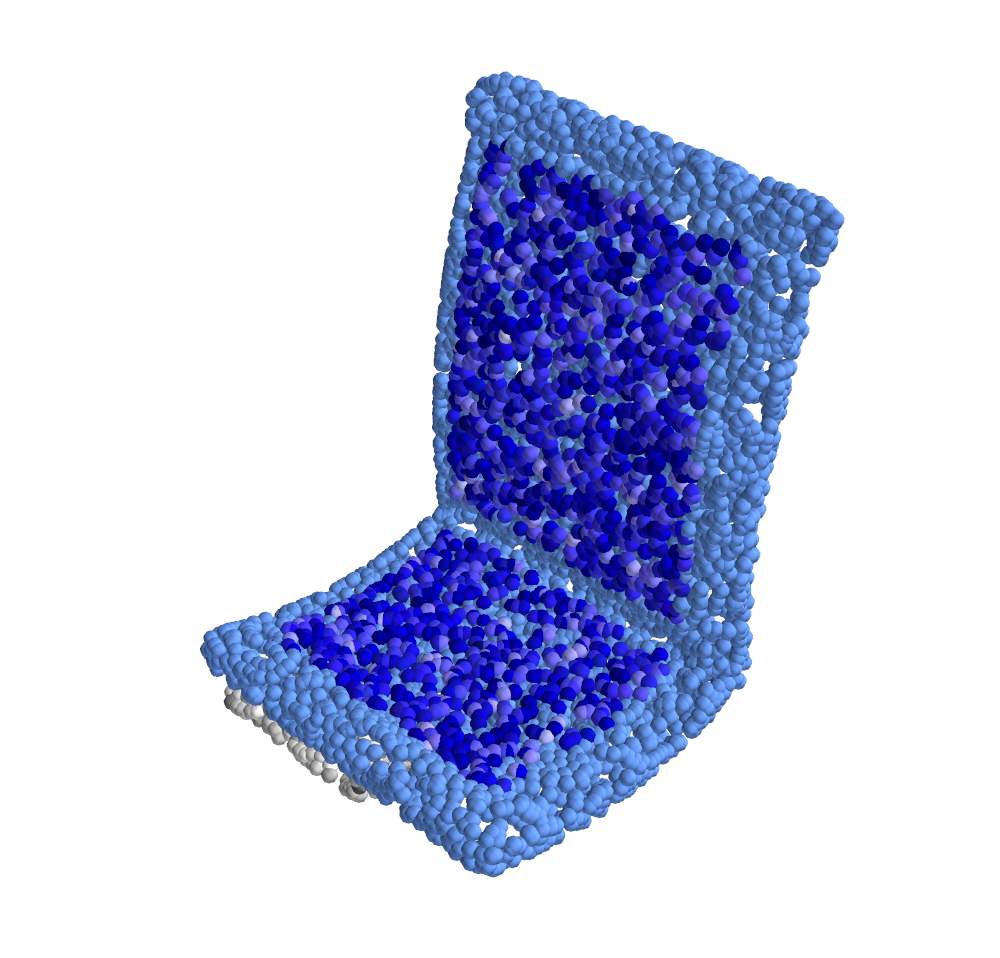} &
    \shape{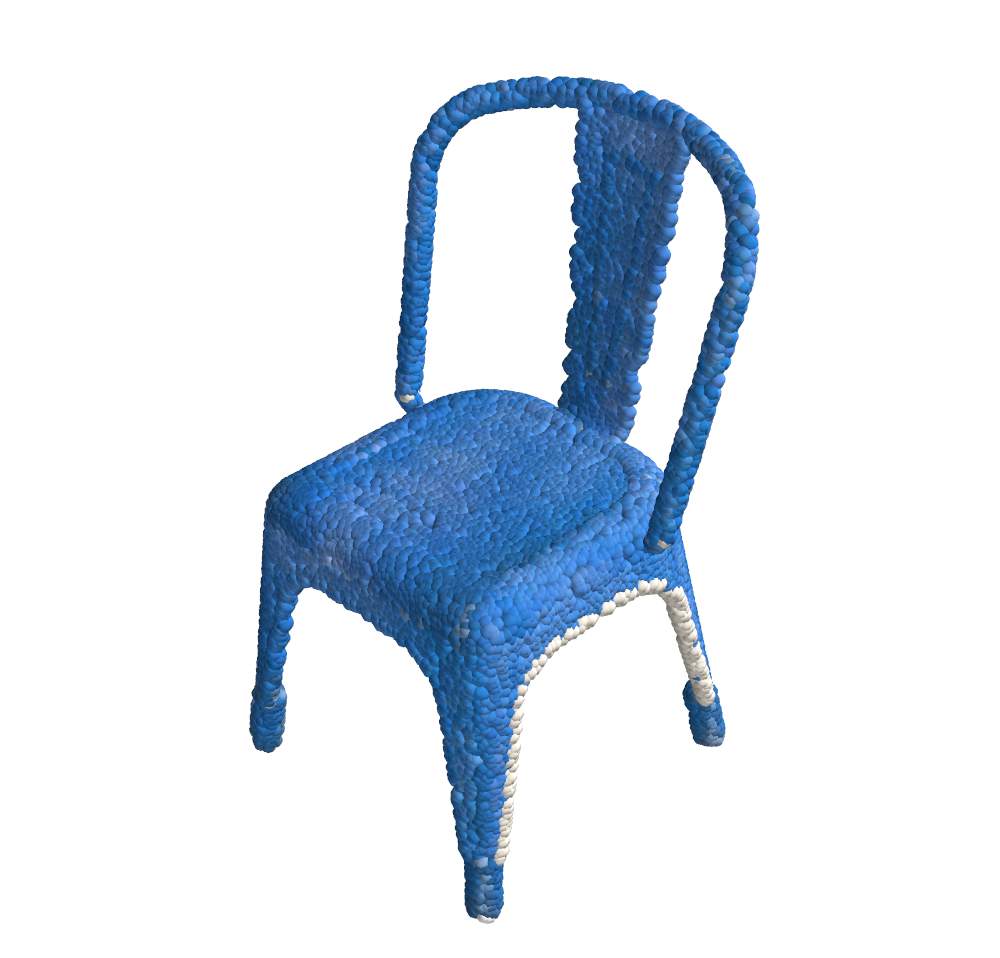} &
    \shape{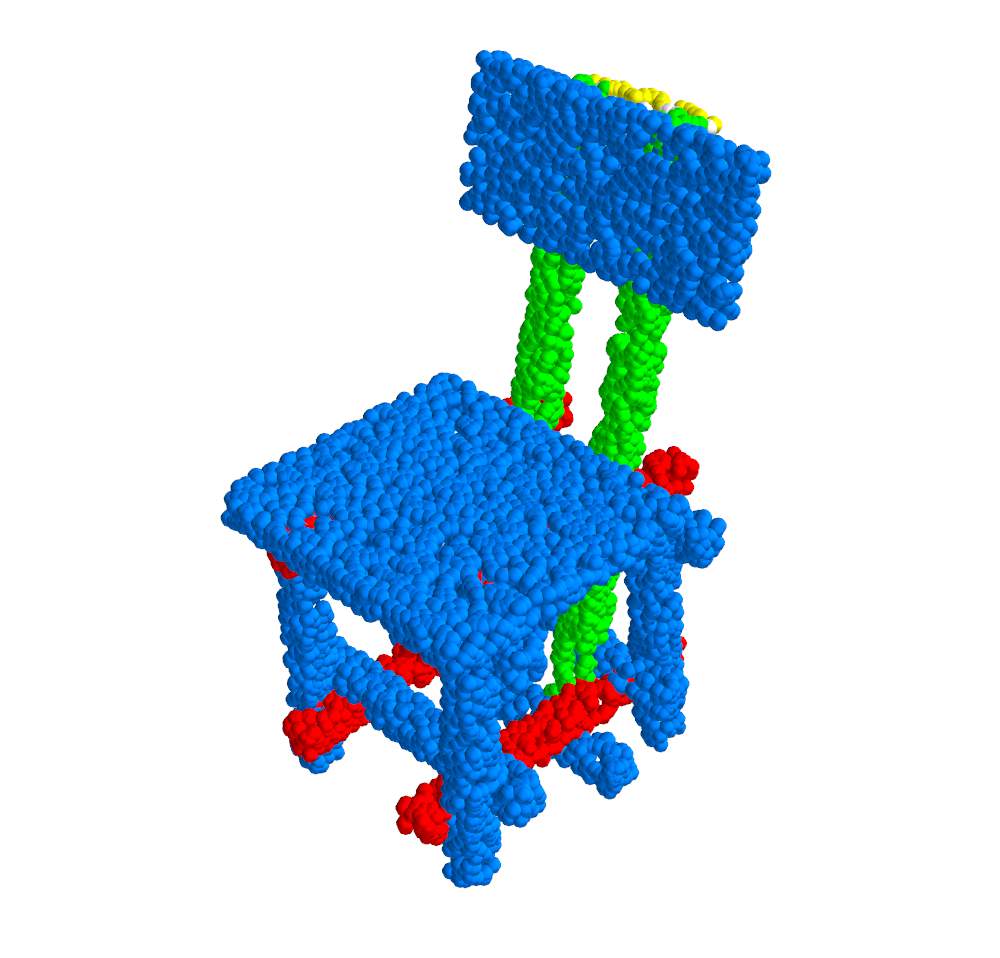} \\

    \midrule
    \textbox{\normalsize{This is a \dashuline{white} plastic chair, with a  \dashuline{square seat and curved back}.  The legs are tapered.}} &
    \shape{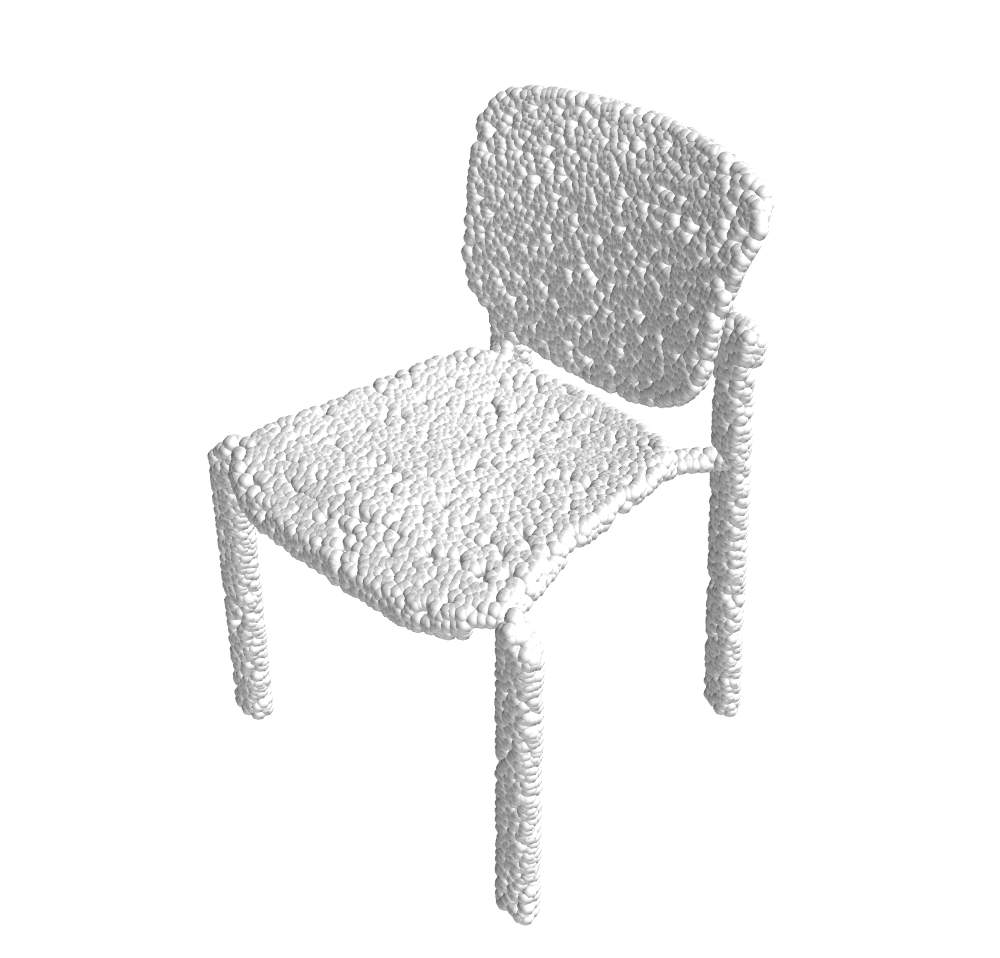} &
    \shape{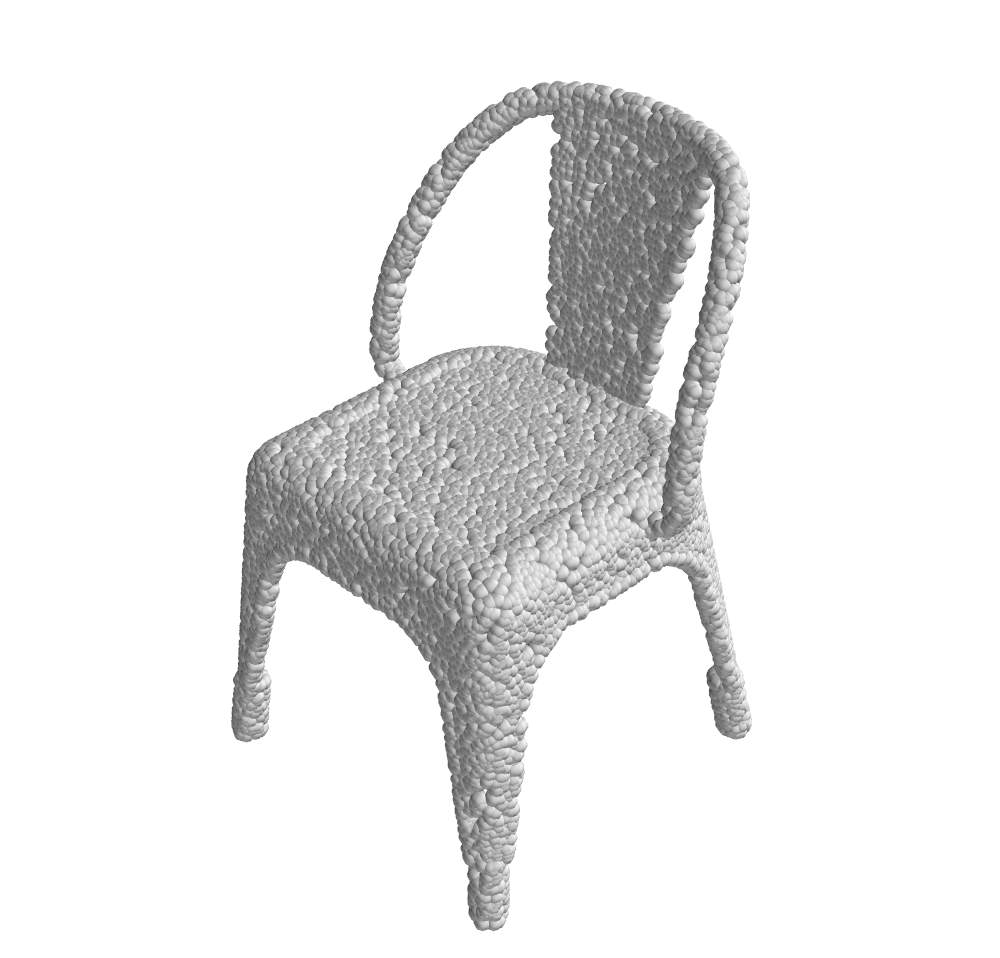} &
    \shape{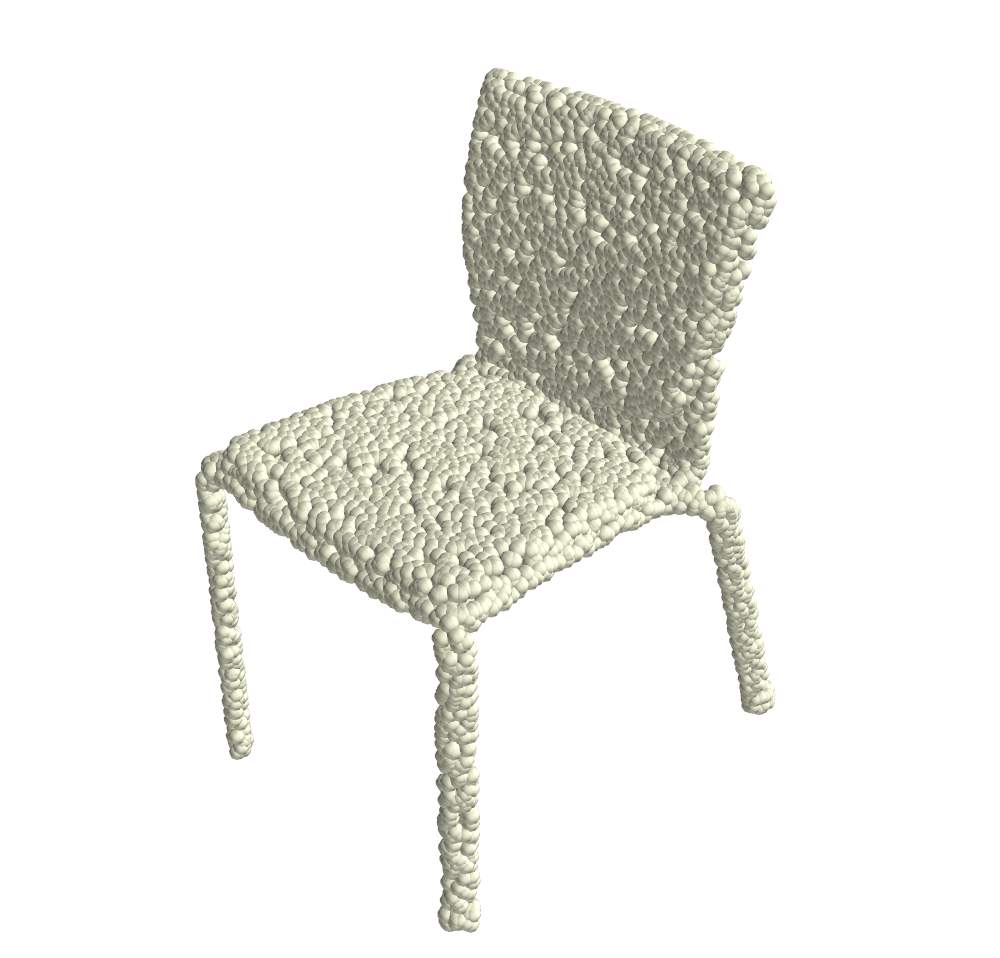} &
    \selectedshape{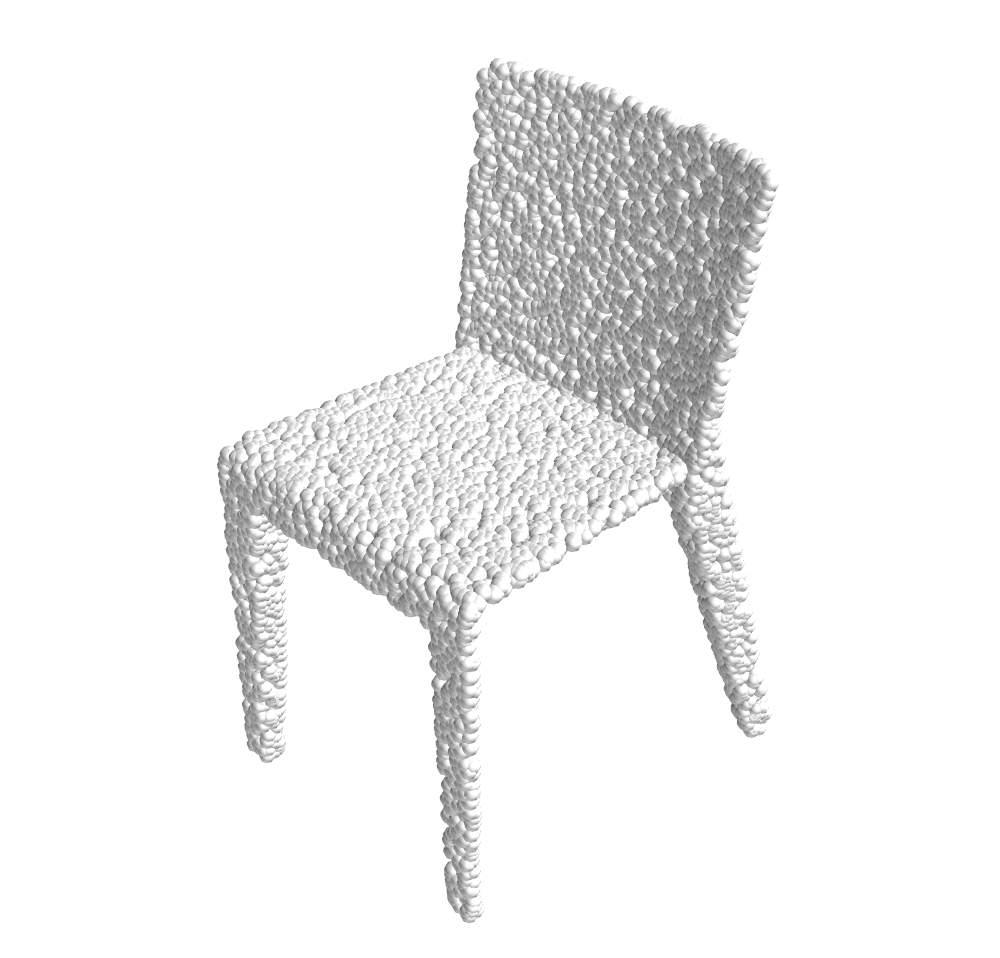} &
    \shape{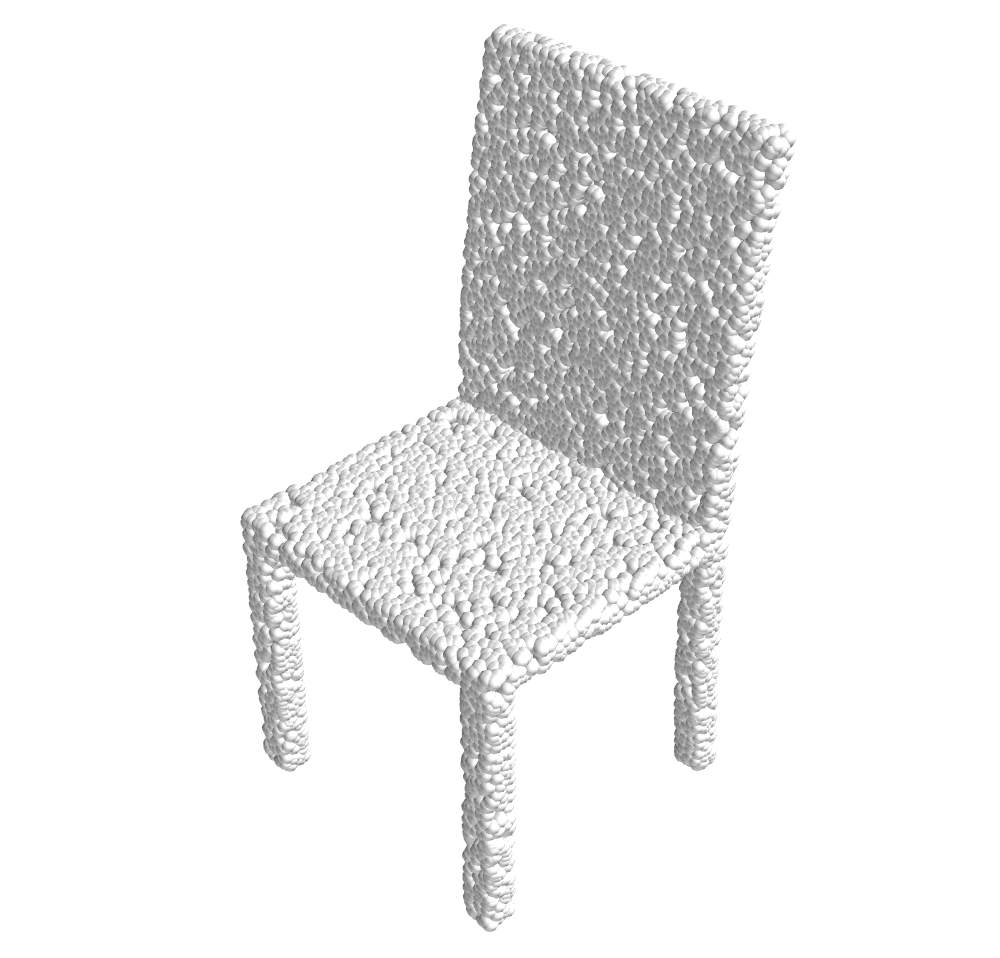} \\

    \bottomrule
\end{tabular}}
\caption{Retrieval result of the proposed Parts2Words (T2S). For each query sentence, we show the top-5 ranked shapes, the ground truth shapes are marked as \textcolor{red}{red boxes}. The words on the corresponding to the details of the retrieved shapes are marked in \dashuline{bold}.}
\label{fig:t2s}
\end{figure}

\begin{figure}
\normalsize
\setkeys{Gin}{width=\linewidth}
\centering
\resizebox{\columnwidth}{!}{
\begin{tabular}{p{2cm}p{0.8\textwidth}}
    \toprule
    \normalsize{Query Shape} & \normalsize{Retrieval Result} \\
    \midrule
    
    \multirow{5}{*}[-1.4em]{
    \begin{minipage}[c]{0.13\textwidth}
    \includegraphics[]{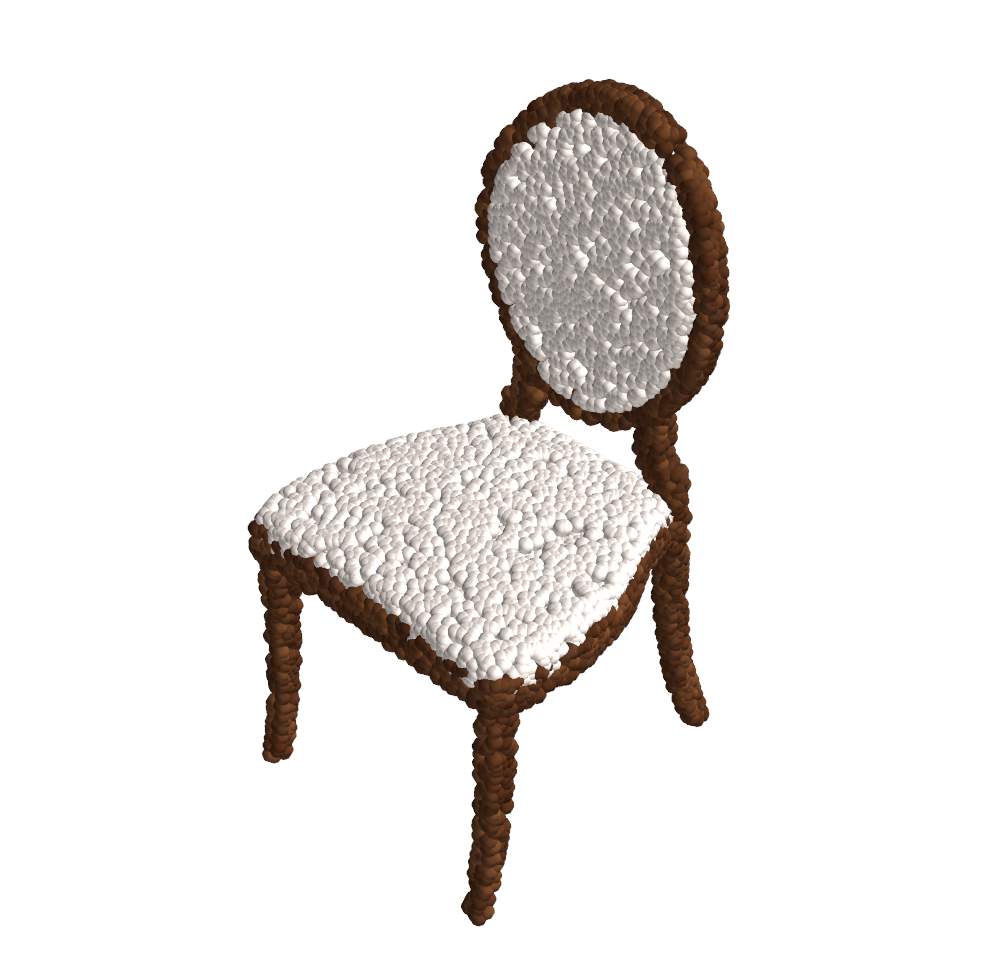} 
    \end{minipage}
    } & 
    \small{\textcolor{red}{1. wooden chair with white color sponge on seating area and back support. trapezium shape seating area and eclipse shape back support .}} \\
    & \small{\textcolor{red}{2. brown and white with round back . appear to be wood and a soft material on the seat and back .}} \\
    & \small{\textcolor{red}{3. a fancy dining chair with white padding and brown , wooden frame . the backing be round with wooden framing and a white pad .}} \\
    & \small{4. wooden chair with four leg , gray padded seat and back rest , the back rest be tall and straight . on the top of the back rest there be a semicircular protuberance . the wooden frame have classical style .} \\
    & \small{5. a wooden chair with white upholstered seat} \\
    \midrule
    
    \multirow{5}{*}[-1.4em]{
    \begin{minipage}{0.13\textwidth} \includegraphics[]{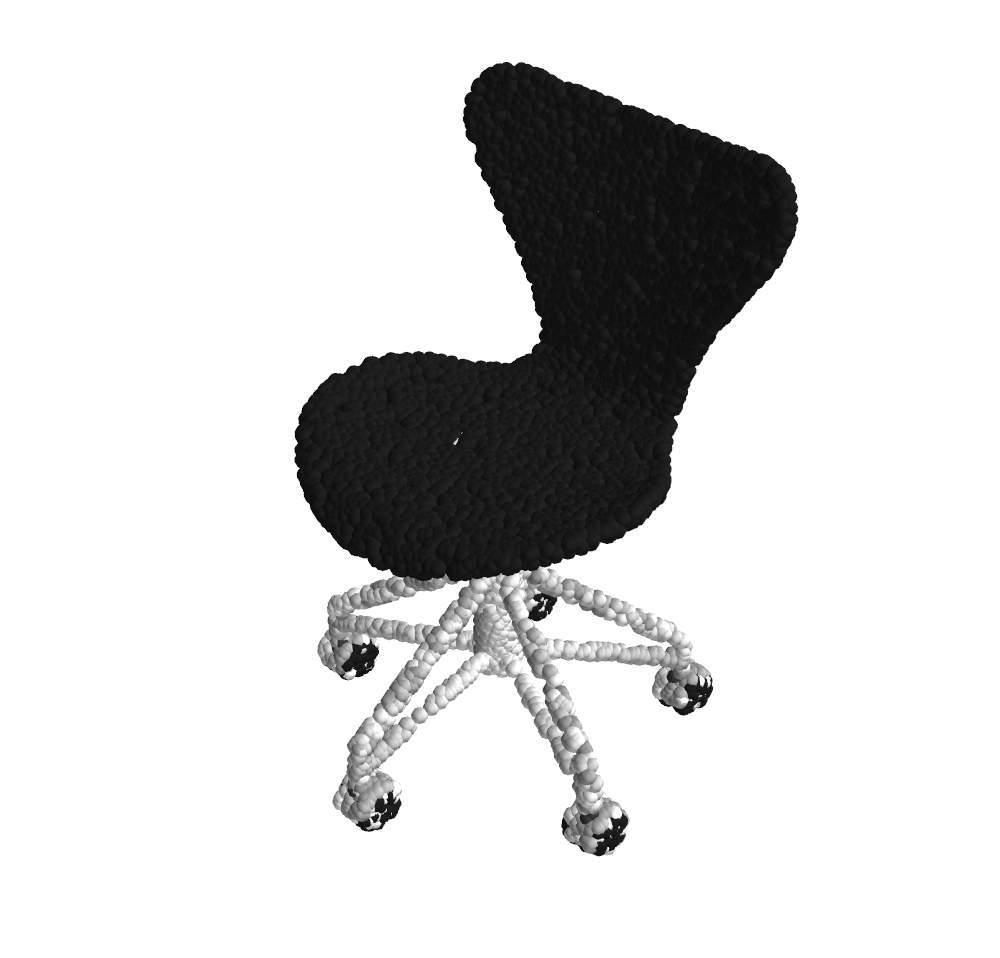} \end{minipage}
    } & 
    \small{\textcolor{red}{1. it be a five - wheeled , black , height - adjustable office chair on coaster wheel .}} \\
    & \small{2. modern irregular shape black chair make of plastic , with an iron structure and with 5 wheel on the bottom} \\
    & \small{\textcolor{red}{3. this be a black , height - adjustable , office chair without arm .   its 5 wheel and single base be make of silver color metal and the seat be make out of a hard , black material .}} \\
    & \small{\textcolor{red}{4. black chair with back support . have single leg with 5 wheel branch from the main support .}} \\
    & \small{5. a black armless computer stool with leg on roller .} \\
    \midrule
    
    \multirow{5}{*}[-1.4em]{
    \begin{minipage}{0.13\textwidth} \includegraphics[]{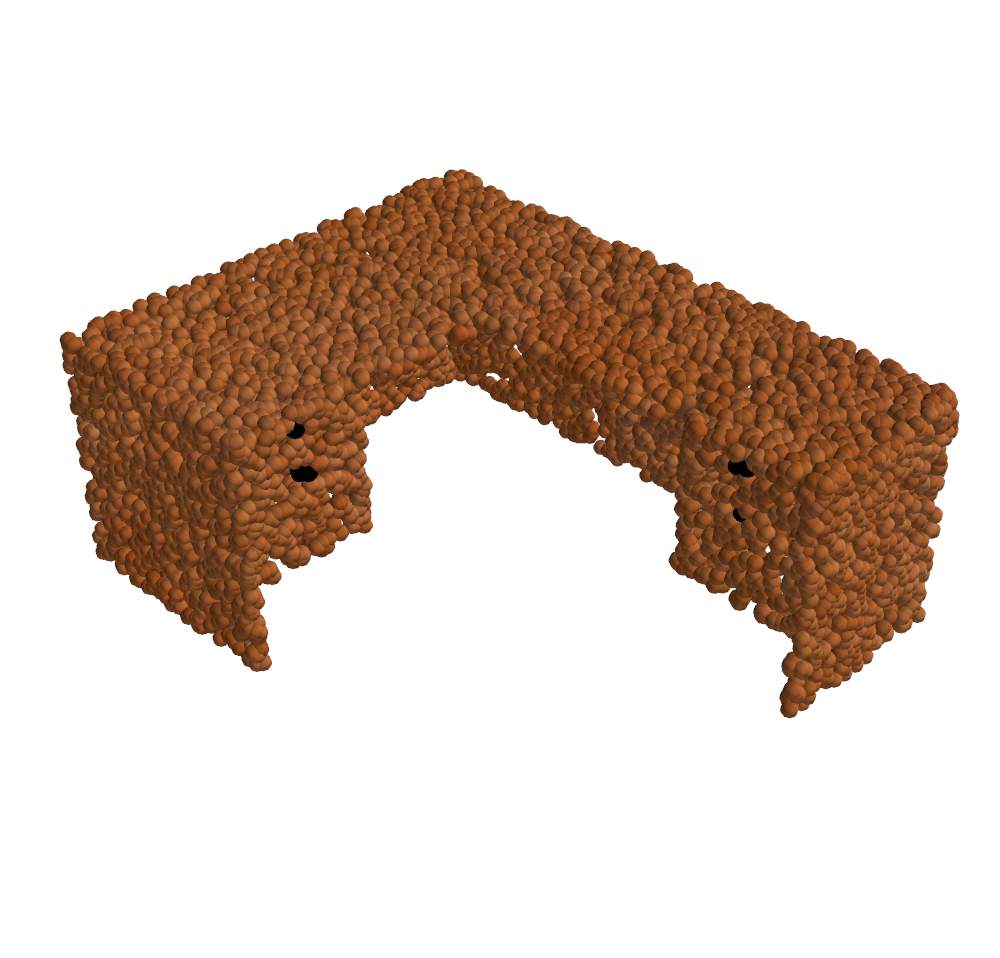} \end{minipage}
    } & 
    \small{1. brown color , l shape , wooden table . box drawer at both leg side . with rectangular l shape plain top .} \\
    & \small{\textcolor{red}{2. brown wooden corner desk unit with drawer two set of two}} \\
    & \small{3. brown business desk .   have an l shape appearance} \\
    & \small{\textcolor{red}{4. a brown wooden desk at a 90 degree angle and drawer on both end of the l shape}} \\
    & \small{\textcolor{red}{5. an l shape dark brown colored wooden table}} \\
    \midrule
    
    \multirow{5}{*}[-1.4em]{
    \begin{minipage}{0.13\textwidth} \includegraphics[]{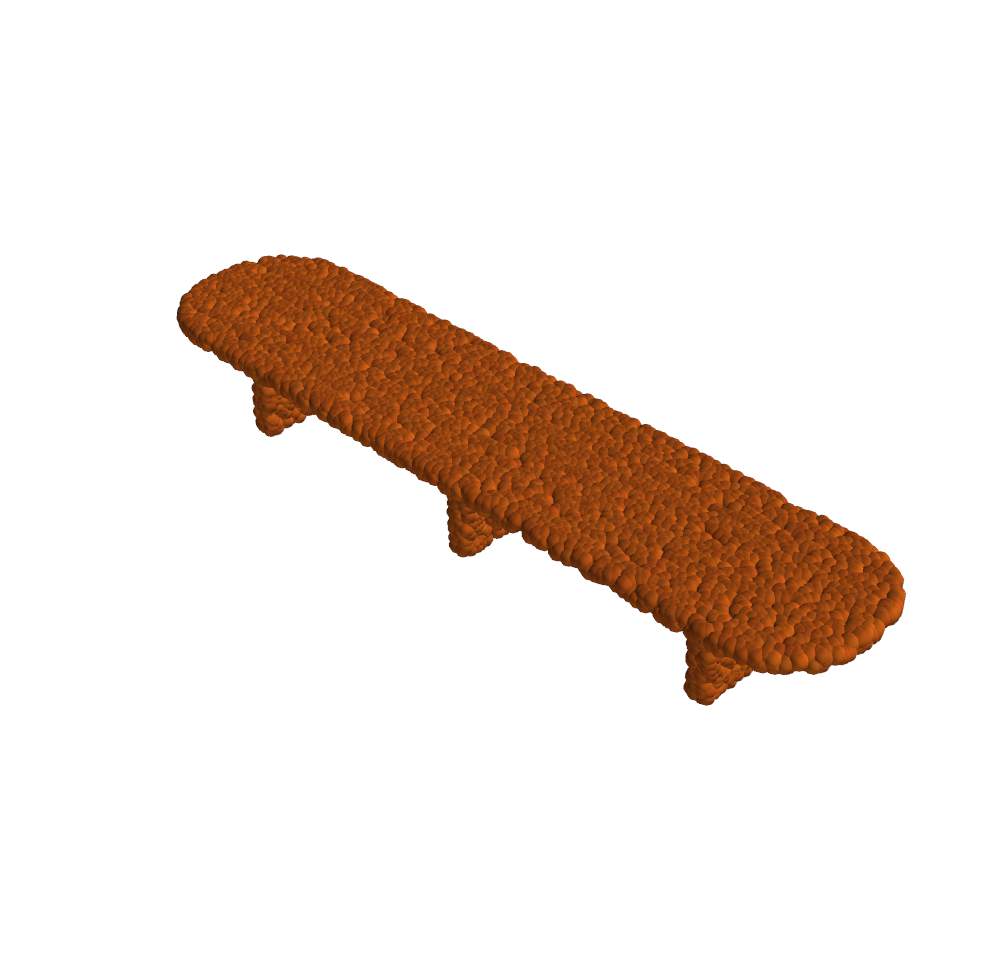} \end{minipage}
    } & 
    \small{\textcolor{red}{1. a long brown table that be oblong in shape}} \\
    & \small{2. a modern oval shape wooden table with six design short leg} \\
    & \small{\textcolor{red}{3. a long brown wooden rectangular table with three full cover leg}} \\
    & \small{4. a brown , rounded wooden table with three leg} \\
    & \small{5. this be a short brown elongated round table . this table would be use as a coffee table with a small table in the middle of a room .} \\
    \midrule
    
    \multirow{5}{*}[-1.4em]{
    \begin{minipage}{0.13\textwidth} \includegraphics[]{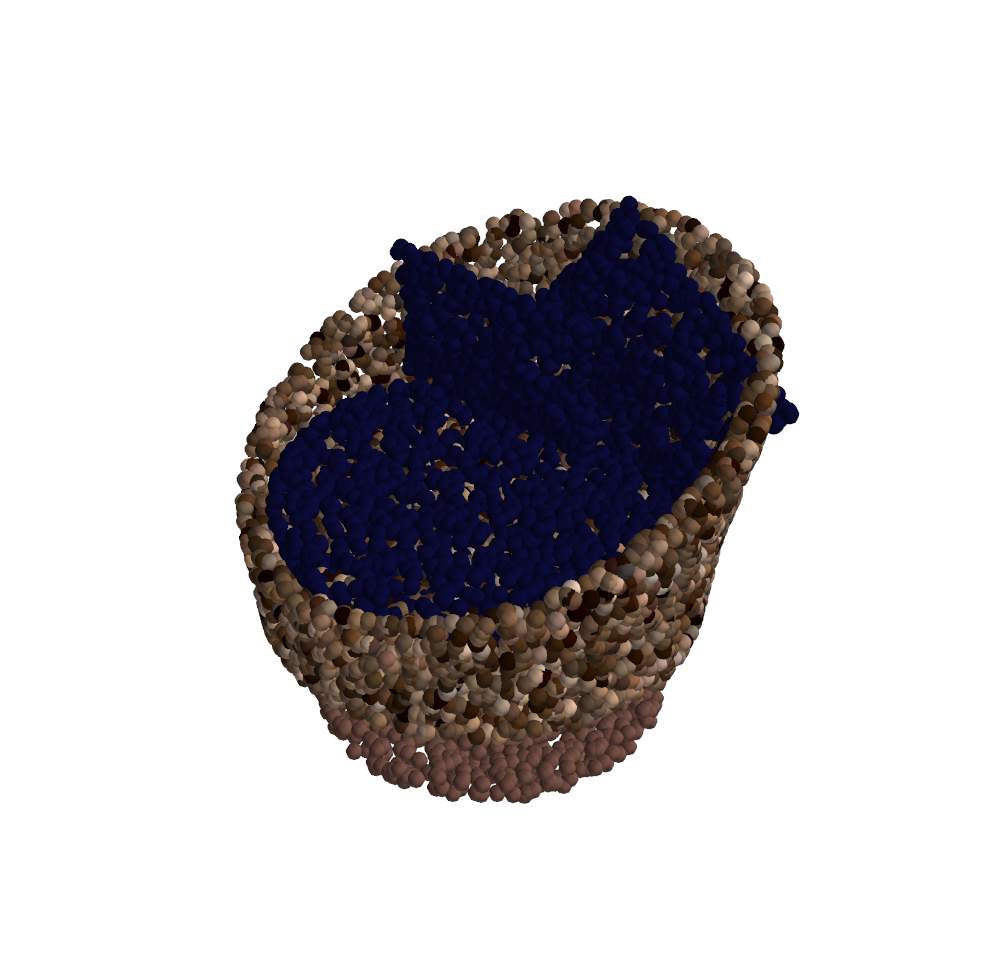} \end{minipage}
    } & 
    \small{\textcolor{red}{1. a wicker round chair with blue plush seat and pillow . the chair be on a circular pedestal}} \\
    & \small{\textcolor{red}{2. round shape brown chair with blue cushion in it}} \\
    & \small{3. a multi color round shape fashion chair with cushion} \\
    & \small{4. round c shape back chair in a checkered pattern all through the chair . circle cushion seat with a cylinder checkered single pole in the middle to support the cushion . the bottom be circle foundation in same checkered pattern .} \\
    & \small{\textcolor{red}{5. one short round chair with three royal blue colored cushion and have a round backrest}} \\
    \midrule
    
    \multirow{5}{*}[-1.4em]{
    \begin{minipage}{0.13\textwidth} \includegraphics[]{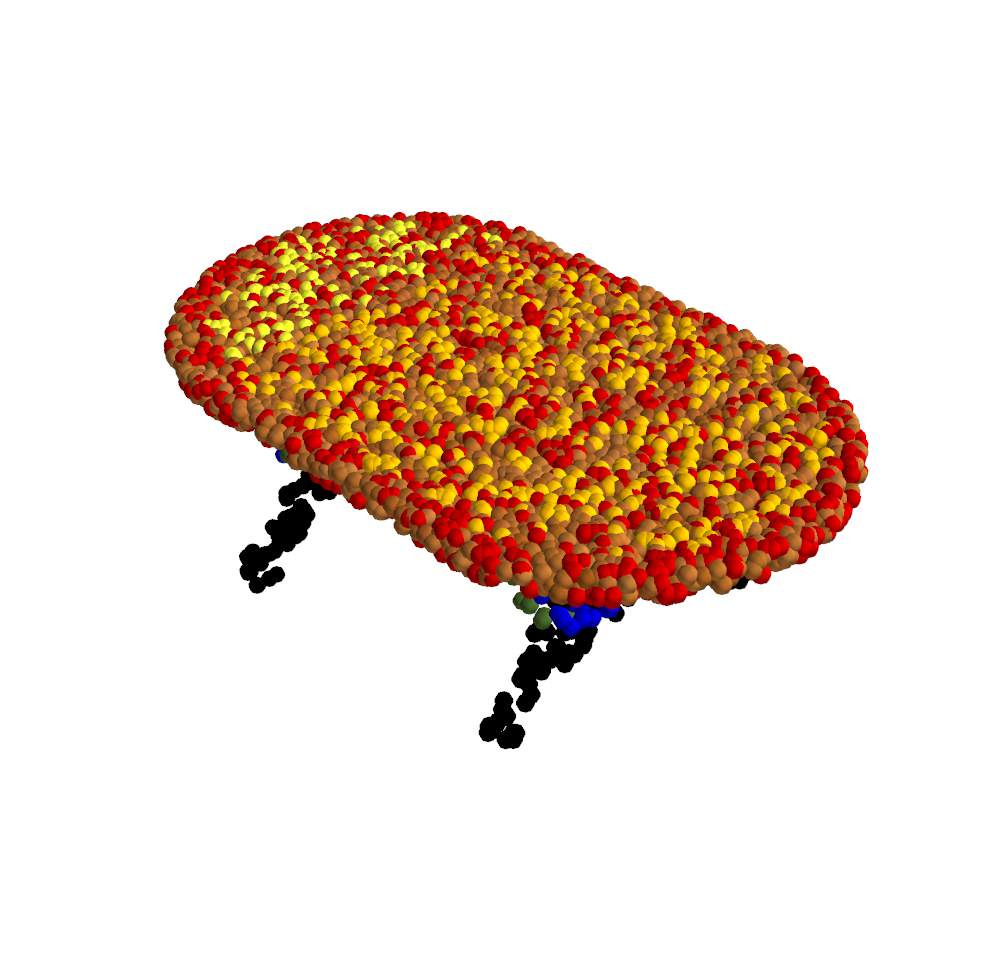} \end{minipage}
    } & 
    \small{\textcolor{red}{1. an oval shape table which be yellow at top with red lining . below that blue color be see .}} \\
    & \small{\textcolor{red}{2. an oval table with a blue shelf under it .   the tabletop be yellow with a narrow red trim around it .}} \\
    & \small{\textcolor{red}{3. the object be a large oval table with a yellow top and red rim around the yellow . it also have a blue shelf underneath and black leg .}} \\
    & \small{4. an oval shape pool table with a green felt top , wooden out layer and two moon shape wood leg} \\
    & \small{5. oval table with 4 leg and material from wood , plastic , brown wood and blue plastic help the table very luxury} \\
    
    \bottomrule
\end{tabular}}
\caption{Retrieval result of the proposed Parts2Words (S2T). For each query shape, we show the top-5 ranked sentences, the ground truth sentences are marked in \textcolor{red}{red}.}
\label{fig:s2t}
\end{figure}

\paragraph{Data preparation}
We evaluate our proposed network on a 3D-Text cross-modal dataset~\cite{chen2018text2shape}. However, this dataset does not include 3D point clouds and the segmentation prior. To resolve this issue, we establish our training samples using two additional datasets, ShapeNet~\cite{2015ShapeNet} and PartNet~\cite{mo2019partnet}, which share the same 3D models. 
Moreover, in contrast to the segmentation annotations offered by ShapeNet~\cite{2015ShapeNet}, the segmentation annotations present in PartNet~\cite{mo2019partnet} demonstrate a level of inherent object classification supervisory capability.
We sample point clouds with color from meshes in ShapeNet~\cite{2015ShapeNet}, and assign each point a label using the fine-grained, instance-level, and hierarchical 3D segmentation ground truth of the same 3D shapes in PartNet~\cite{mo2019partnet}.
We notice that the original point clouds in ShapeNet dataset~\cite{2015ShapeNet} have the wrong color which is inherited from its inner meshes (The origin point cloud with incorrect color is shown in Figure~\ref{fig:data}). The incorrect color input will cause problems for our model to understand color information. To mitigate this problem, we first remove the inner mesh from the origin mesh data, then we sample points with the correct color.
At the same time, we leverage ICP~\cite{DBLP:journals/pami/BeslM92} 
to map the 3D segmentation ground truth on shapes in PartNet to the sampled point clouds. Finally, we obtained point clouds with color and segmentation ground truth with different granularities.
In the 3D-Text dataset that contains chairs and tables, we use 11498 3D shapes as training samples, and the remaining 1434 ones are considered as test data.
And each 3D shape has an average of 5 text descriptions.

\paragraph{Evaluation metrics} 
We employ recall rate (RR@$k$, $k=1, 5$) and NDCG~\cite{DBLP:journals/tois/JarvelinK02} to conduct quantitative evaluation. RR@$k$ is defined as the percentage of correct text/shape in the top $k$ retrieval results. 

\paragraph{Parameter Setting}
We set the number of each point cloud $l$ to $2500$, and output the segmentation result and the matching scores between shapes and text. 
We use the coarse granularity of 17 classes as segmentation ground truth.
For the shape encoder module, we tried to use PointNet and PointNet++ respectively as the backbone. 
In the group pooling module, we use average pooling to aggregate part embedding. Additionally, we ignore the part with less than $1\%$ of total points. 
In the matching module, we set the dimension of embedding to $1024$, which is consistent with \cite{han2019y2seq2seq,han2020shapecaptioner}. 
We also use the vocabulary of $3587$ unique words and a single-layer Bi-directional GRU as the text encoder. 
Due to the limited vocabulary of the Text2Shape dataset used in the experiment, the impact of text pre-training methods like BERT on the results is minimal (within $1\%$). Therefore, we only use a text encoder trained from scratch in the comparative and ablation experiments.
For the loss function, we adopt a semi-hard negative mining strategy, and the margin $\alpha$ of the triplet ranking loss is set to $0.2$. 
We divide the training process into two stages. First, we pre-train the model only by semantic segmentation loss with 50 epochs and then train multi-task loss with $20$ epochs and we set the balance weight of loss $\beta$ to $40$. Our model uses the Adam \cite{kingma2014adam} algorithm as the optimizer and set the initial learning rate to $0.001$. We used RTX 6000 with 24GB for training and set the batch size to 128.

\begin{figure}[t]
    \begin{minipage}{0.5\linewidth}
    \includegraphics[width=\linewidth]{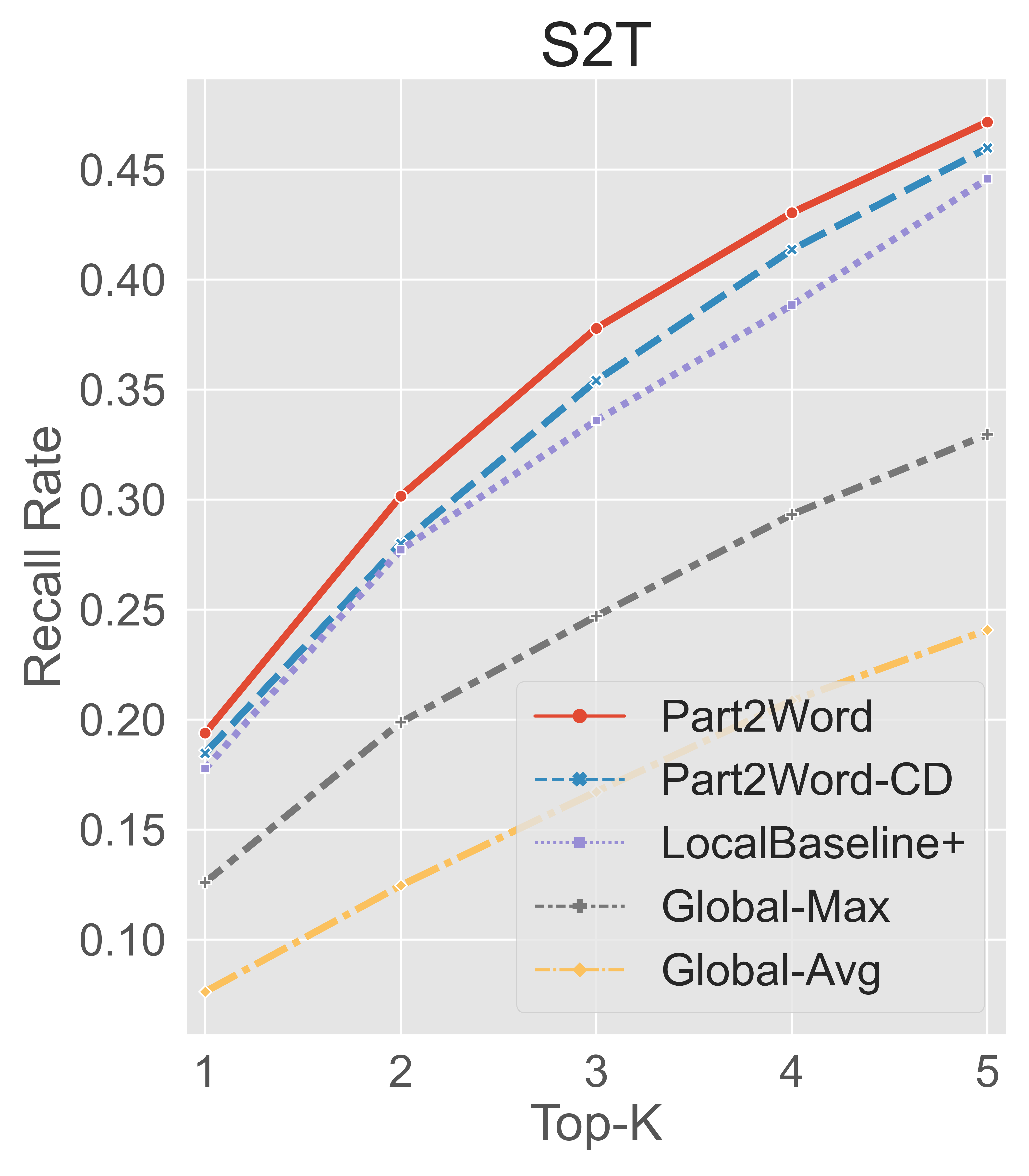}
    \end{minipage}
    \hspace{-0.5cm}
    \begin{minipage}{0.5\linewidth}
    \includegraphics[width=\linewidth]{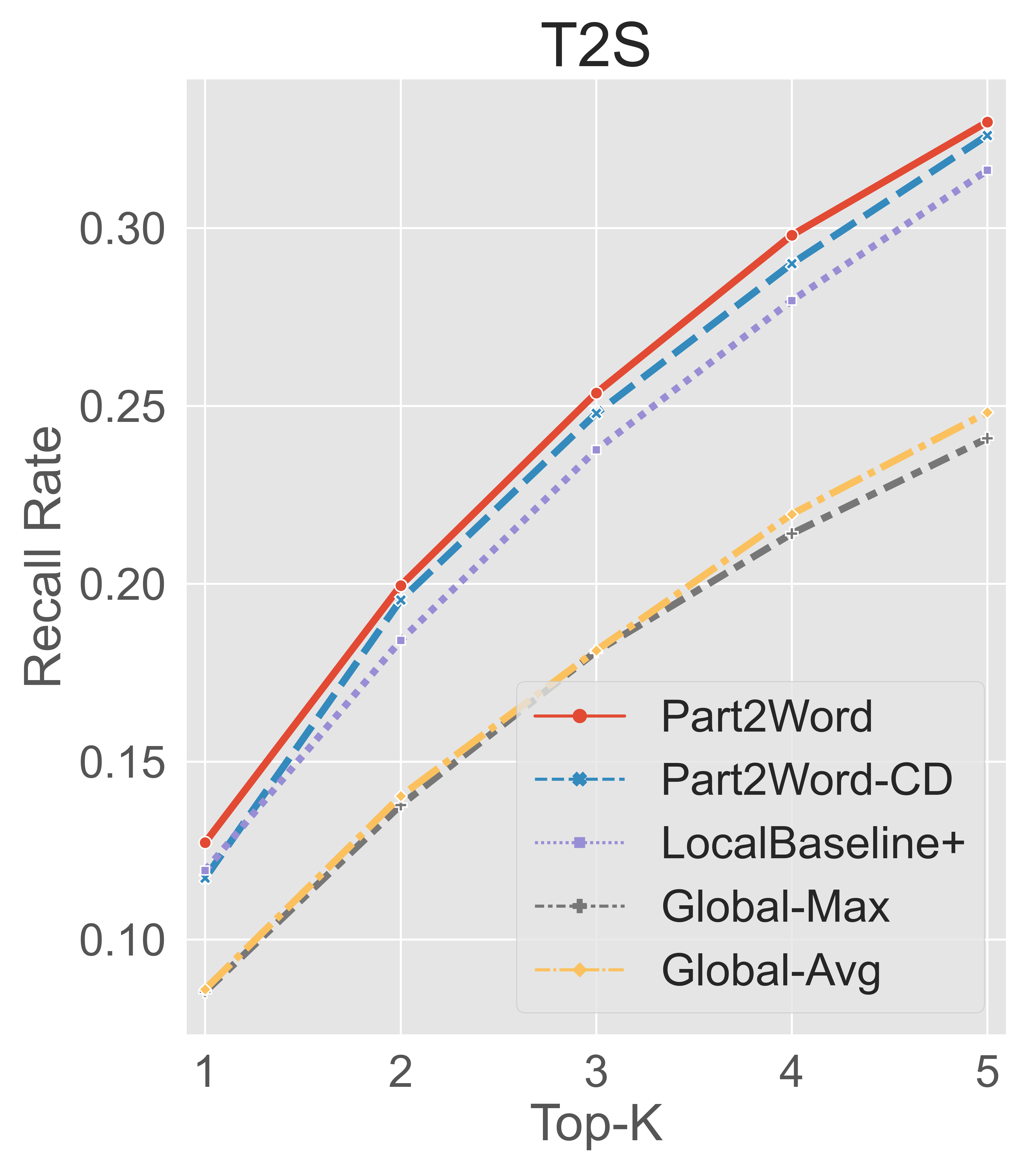}
    \end{minipage}
    \vspace{-0.3cm}
    \caption{Comparison between global/local-based methods. The result shows the local-based methods (Parts2Words, Parts2Words-CD, LocalBaseline+) outperform the global-based methods (Global-Max, Global-Avg).}
    \label{fig:topk}
\end{figure}

\begin{figure}
\centering
\setkeys{Gin}{width=\linewidth}
\large
\renewcommand{\multirowsetup}{\centering}
\newcommand{\shape}[1]{\begin{minipage}{0.08\textwidth} \includegraphics[]{#1} \end{minipage}}
\newcommand{\selectedshape}[1]{\begin{minipage}{0.08\textwidth} \fbox{ \includegraphics[]{#1}} \end{minipage}}
\newcommand{\textbox}[1]{\multirow{2}{6cm}{\begin{minipage}{6cm} \large{#1} \end{minipage}}}
\newcommand{\rotbox}[1]{\begin{minipage}{1.5em} \rotatebox{90}{\normalsize{#1}} \end{minipage}}
\resizebox{0.98\columnwidth}{!}{%
\begin{tabular}{p{6cm}cccccc}
    \toprule
    Query Text &  & top1 & top2 & top3 & top4 & top5 \\

    \midrule
    \textbox{\large{A simple 
\dashuline{circular} table with 
\dashuline{no legs} and only one \dashuline{circular base}.}} &
    \rotbox{TriCoLo} &
    \shape{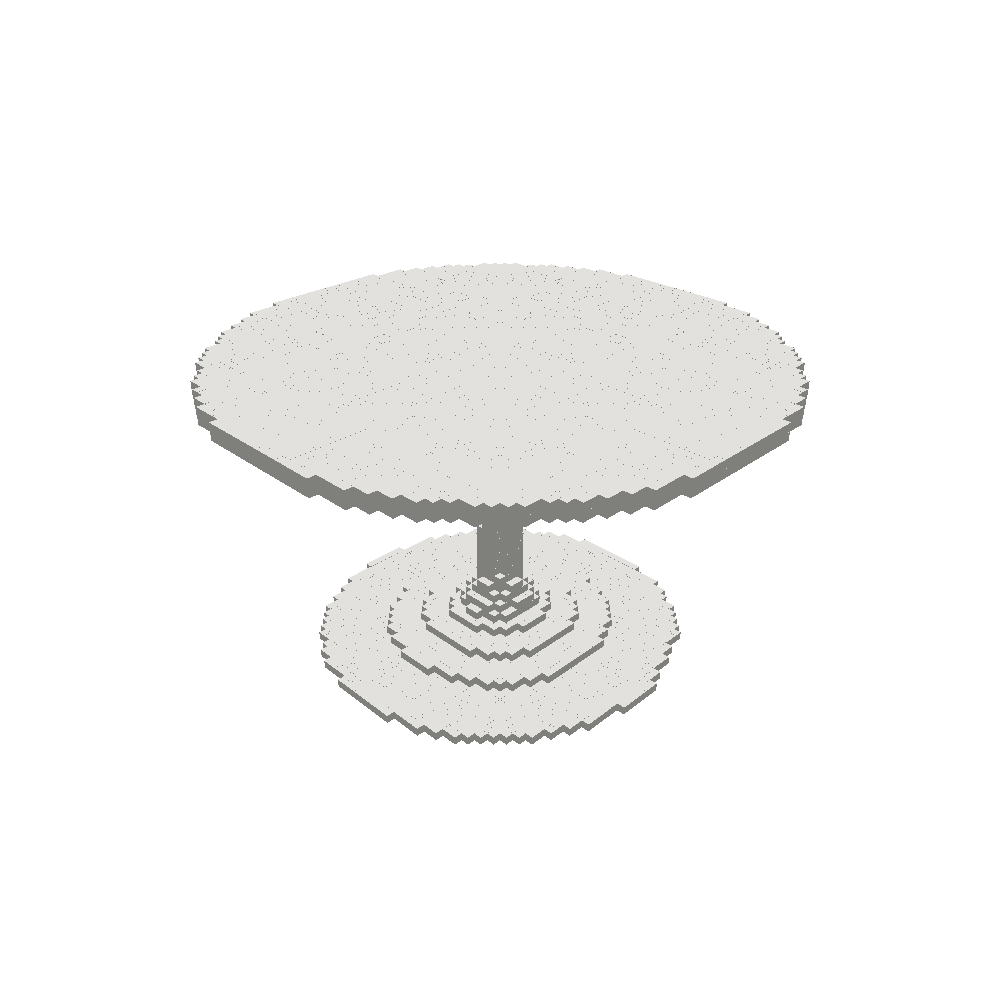} &
    \shape{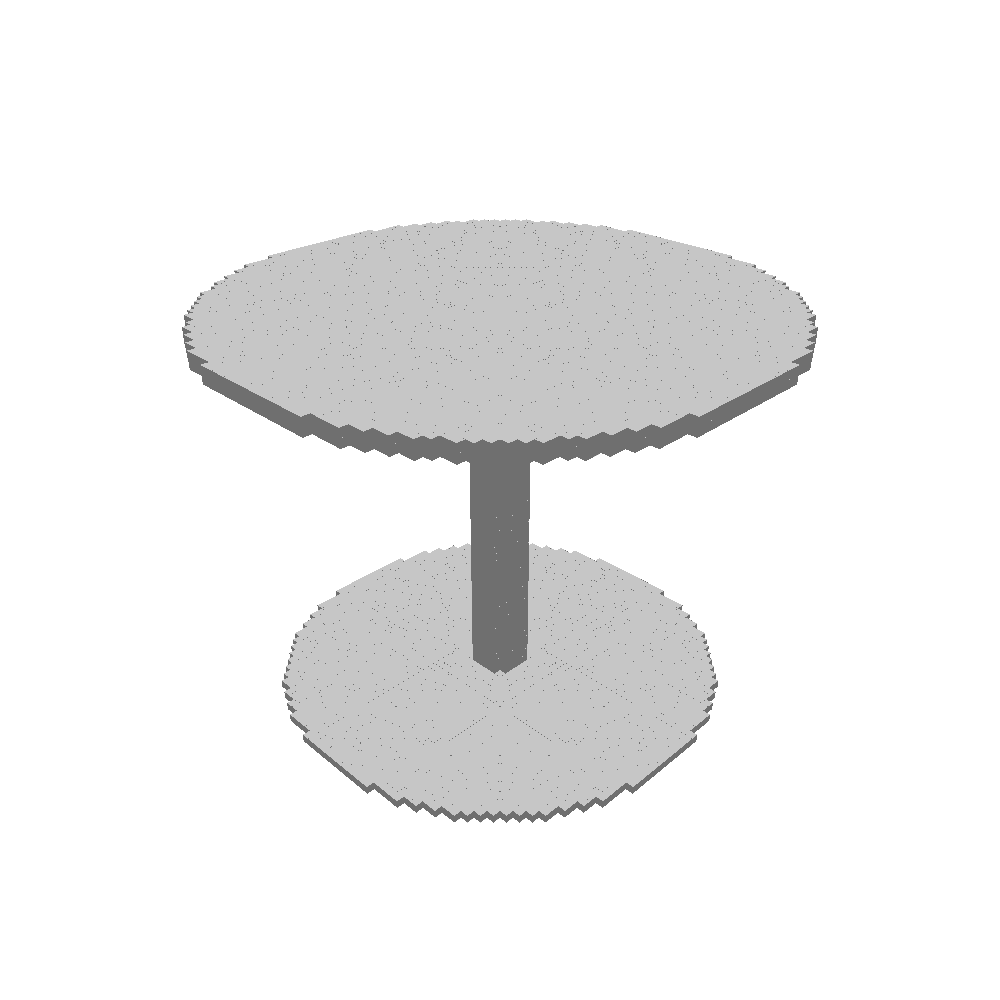} &
    \shape{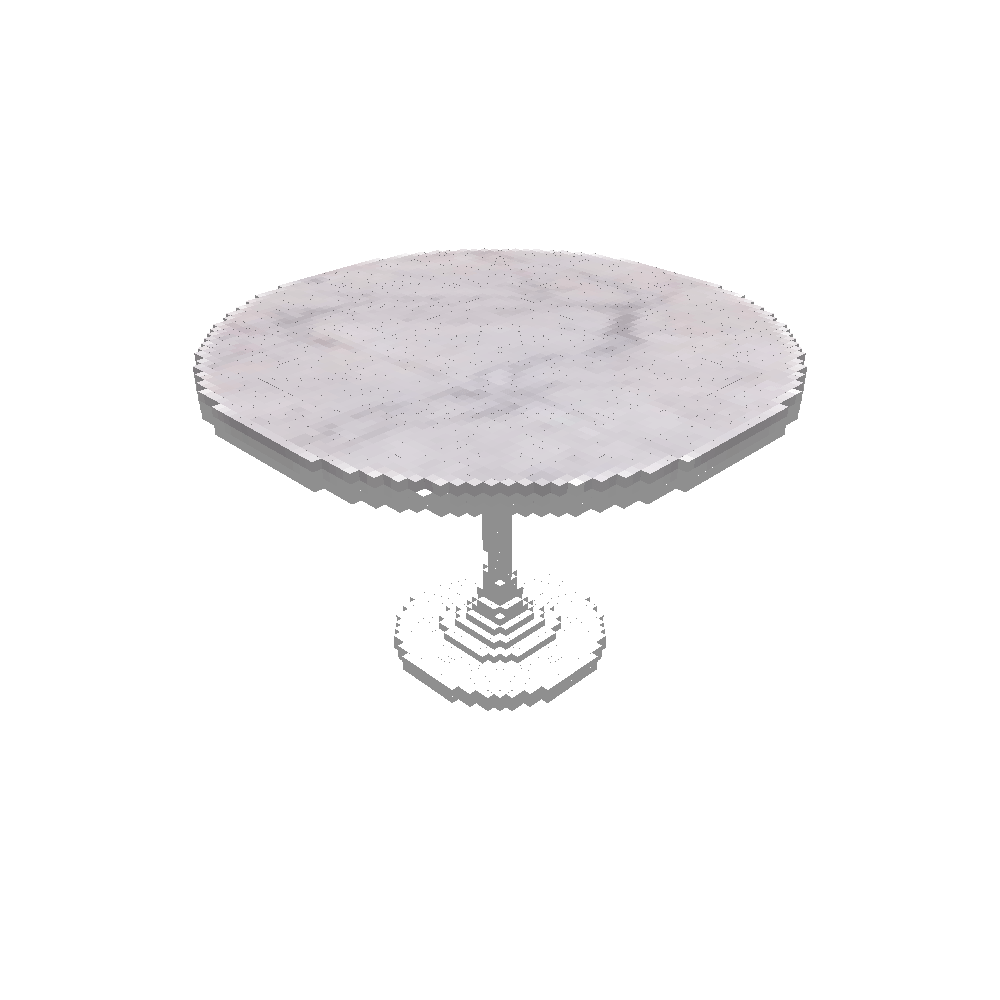} &
    \selectedshape{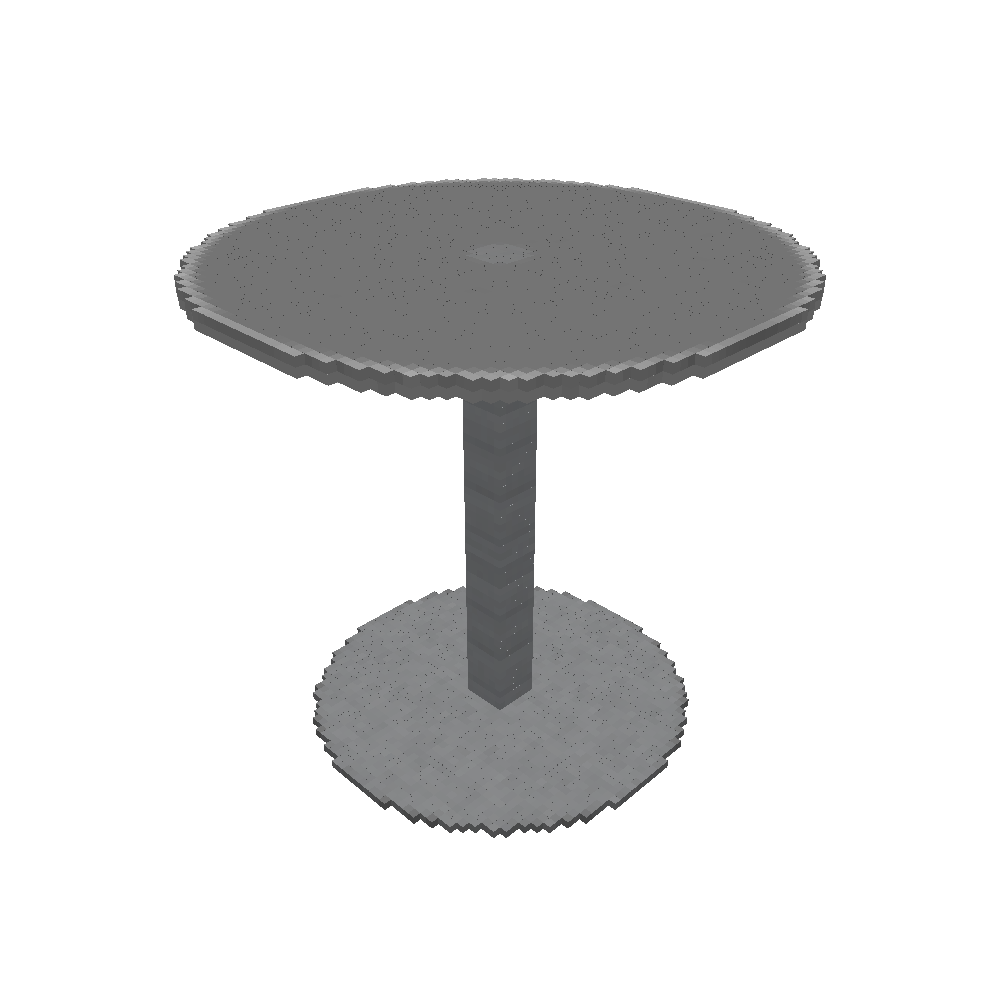} &
    \shape{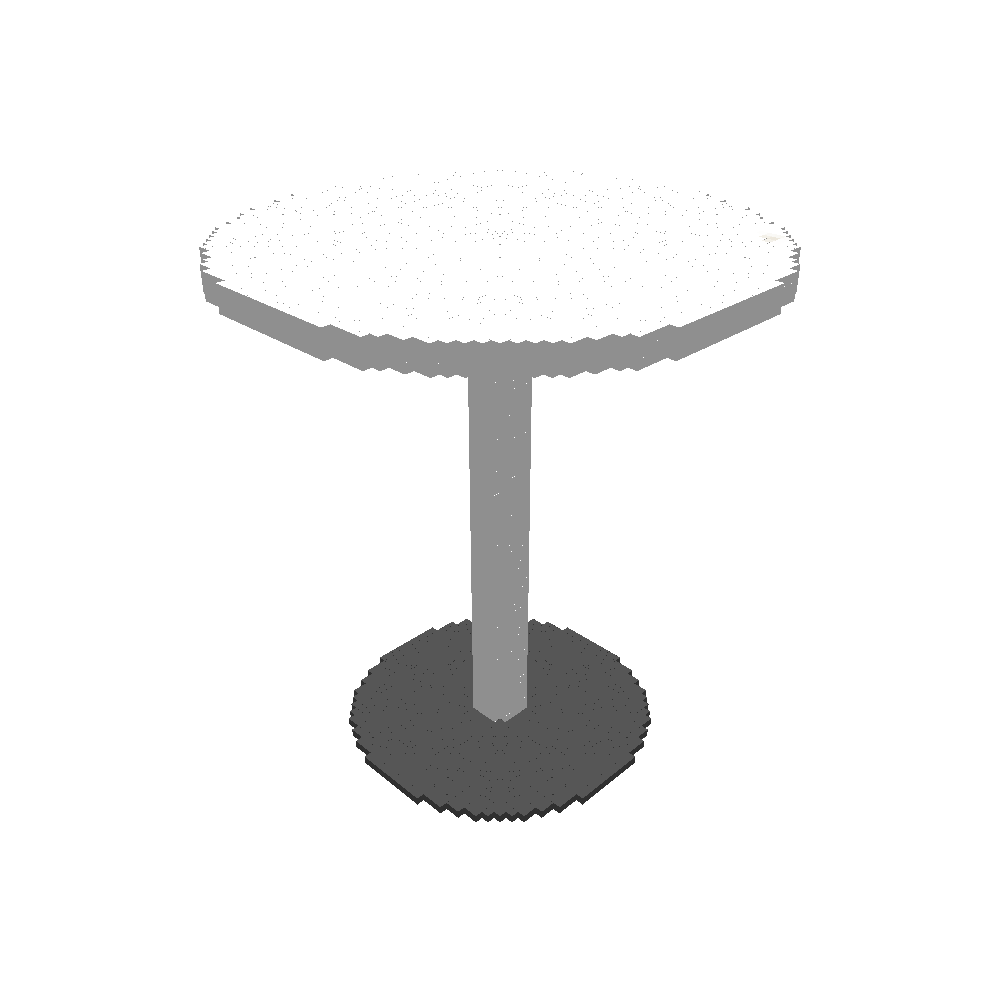} \vspace{0.5em} \\
    &
    \rotbox{Ours} &
    \selectedshape{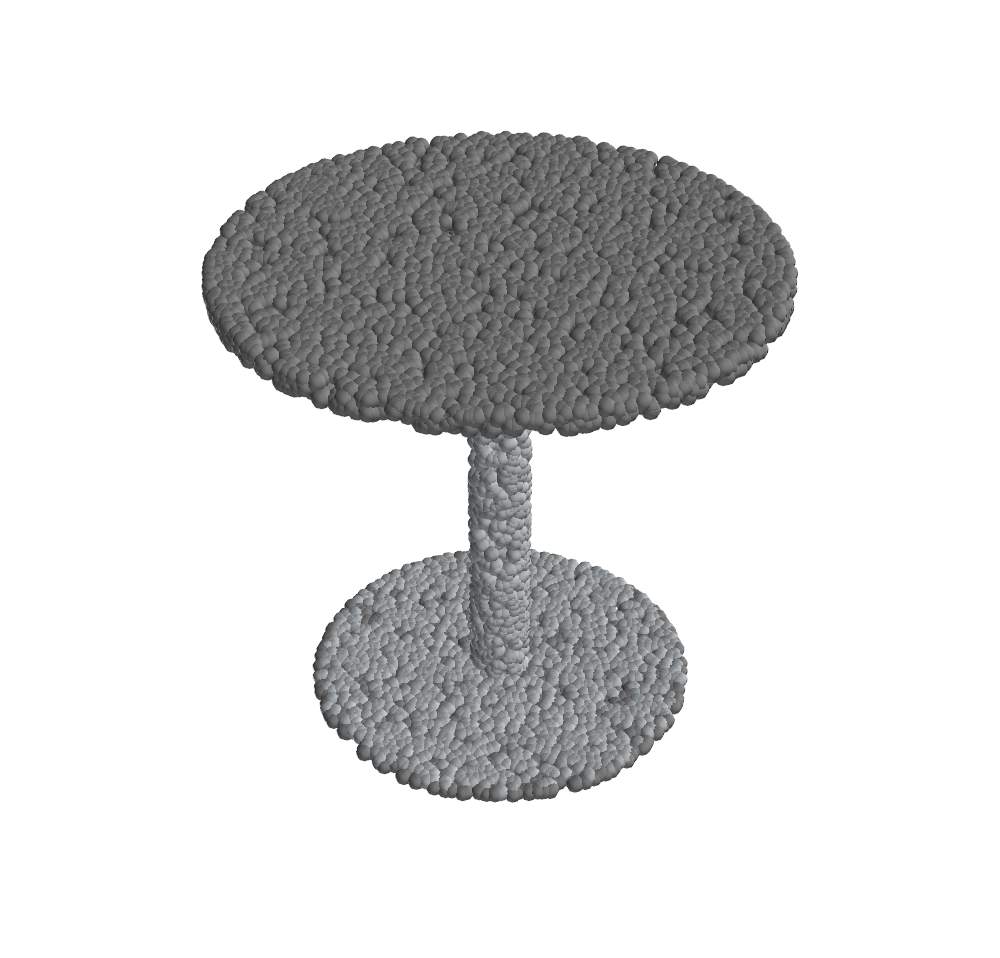} &
    \shape{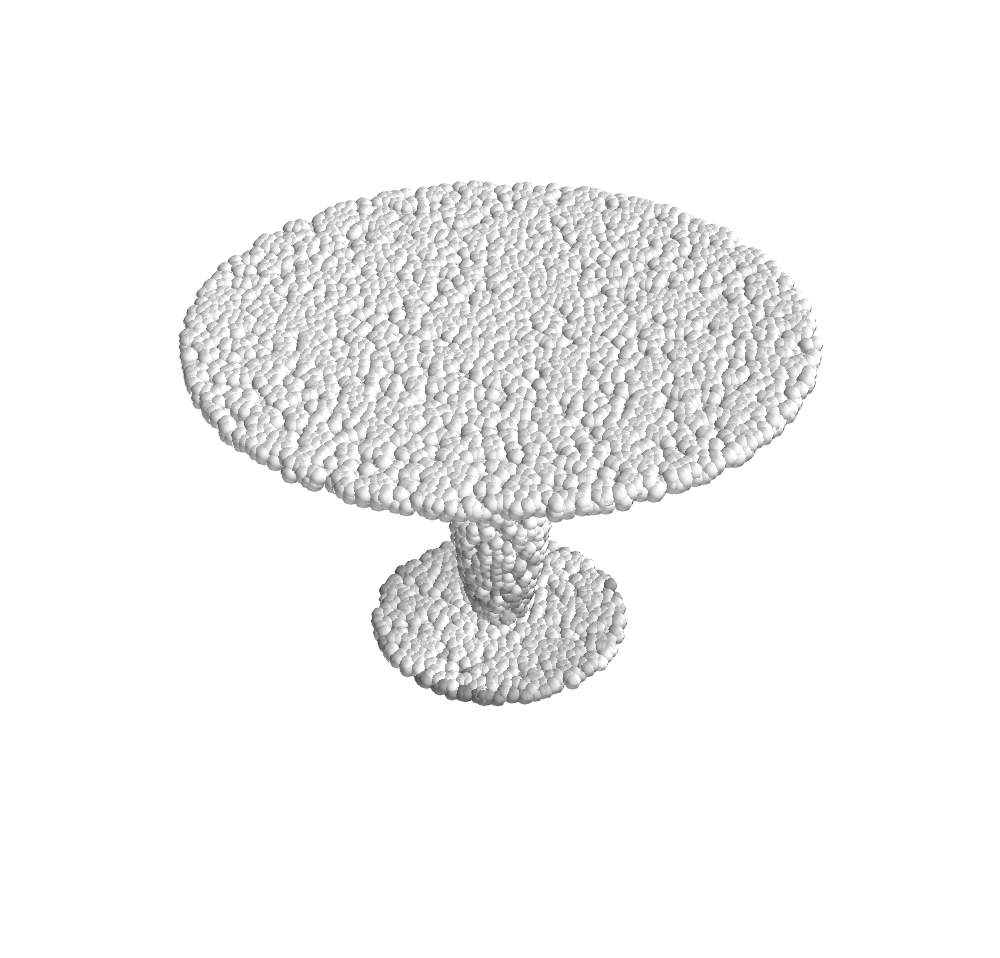} &
    \shape{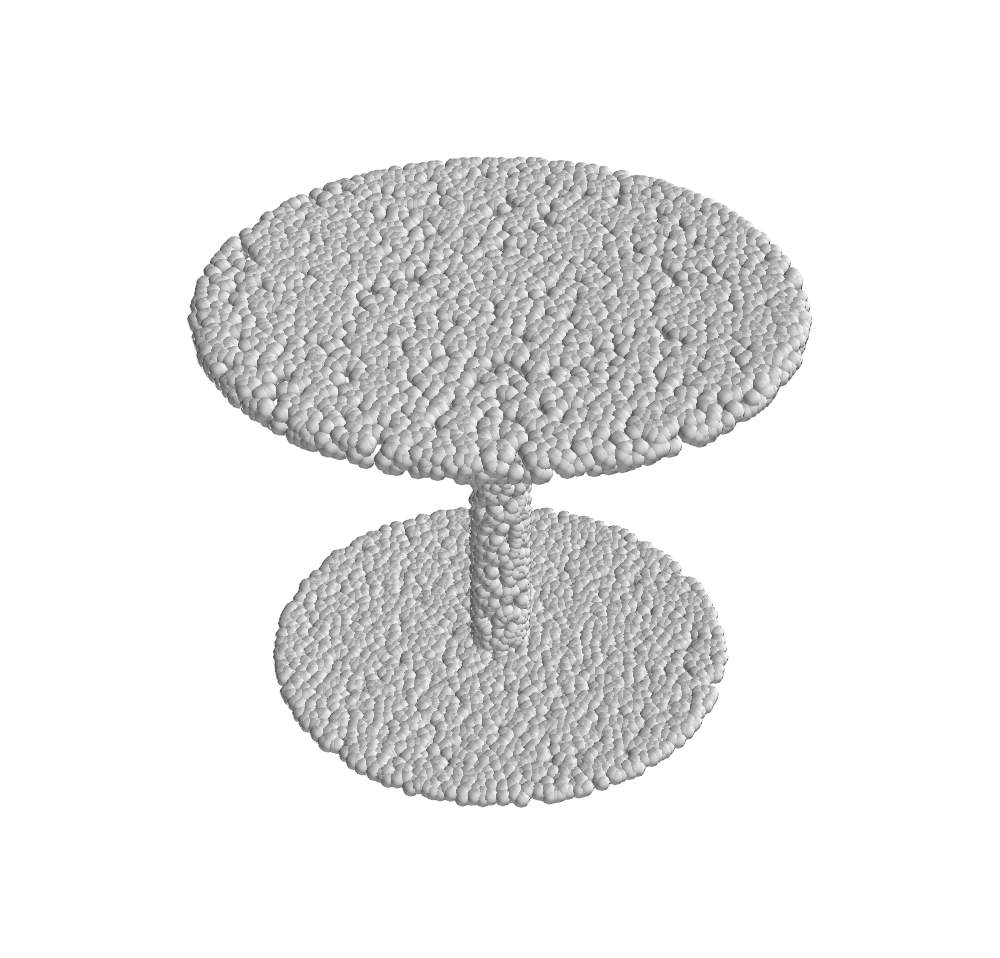} &
    \shape{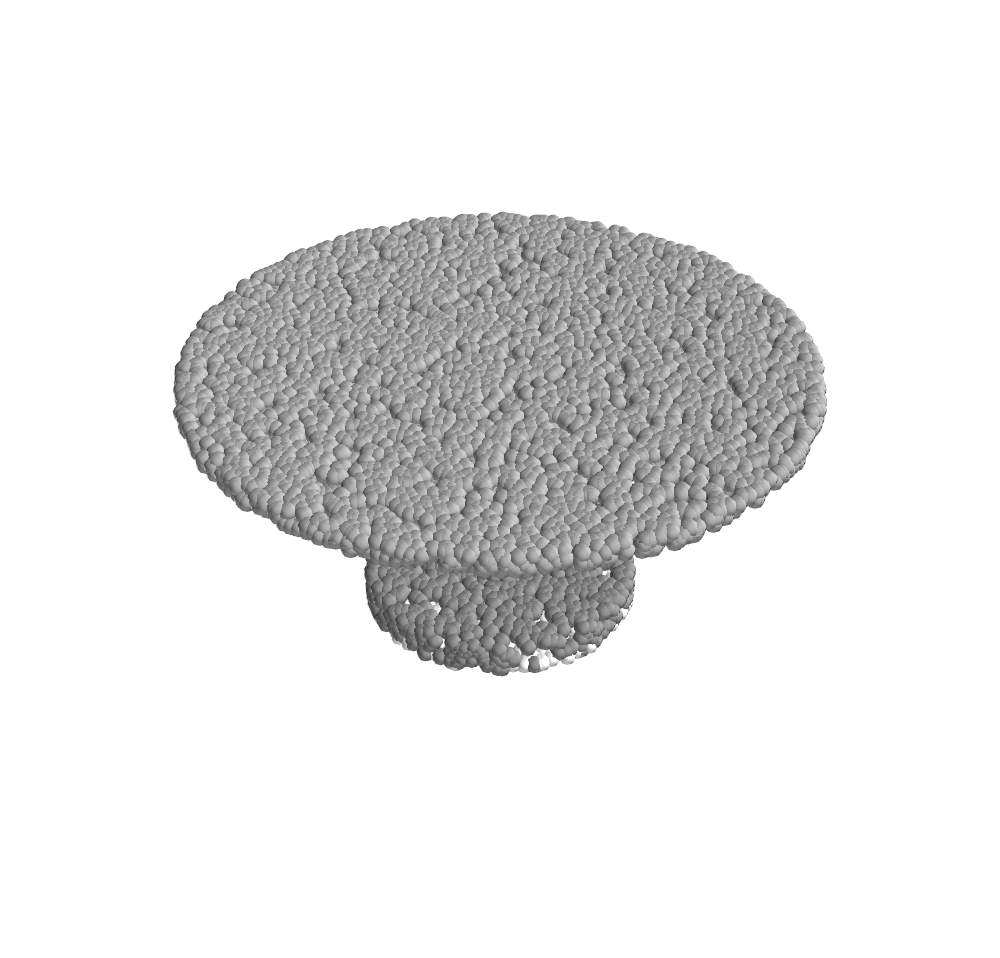} &
    \shape{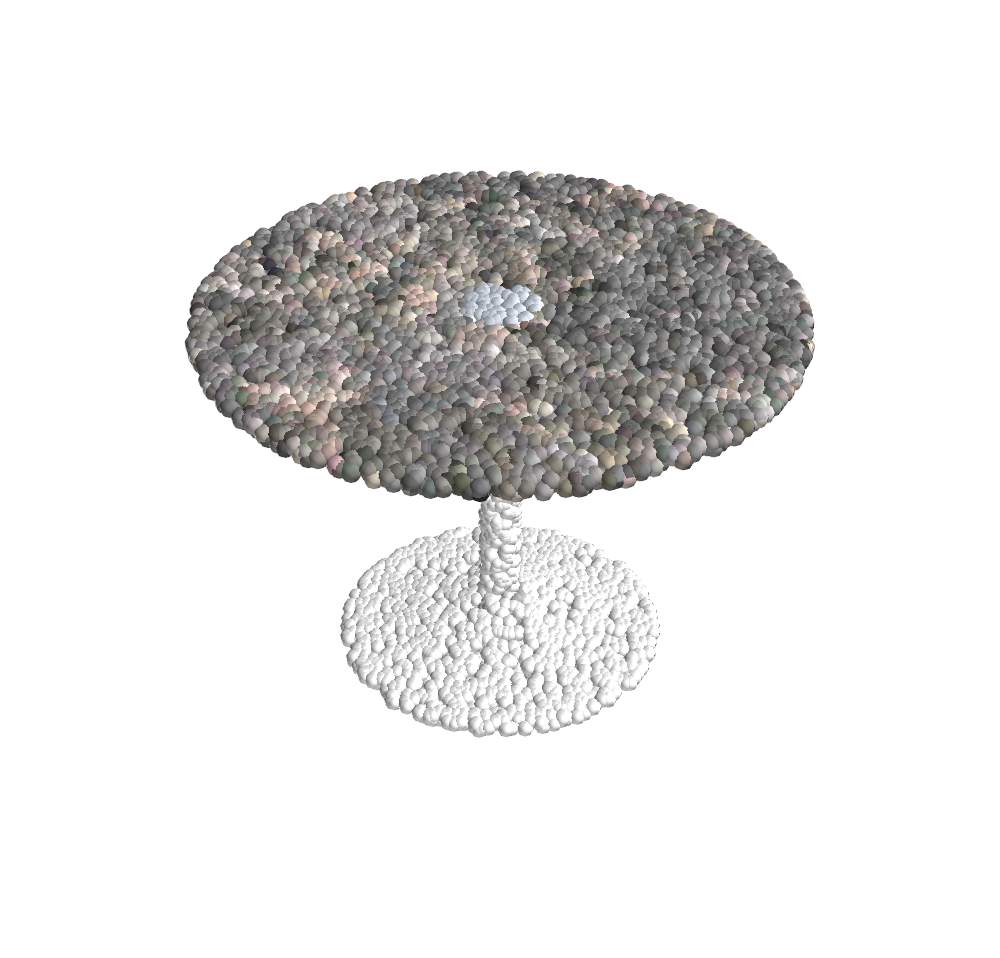} \\

    \midrule
    \textbox{\large{This is 
\dashuline{greenish top} wooden Billiards table}} &
    \rotbox{TriCoLo} &
    \shape{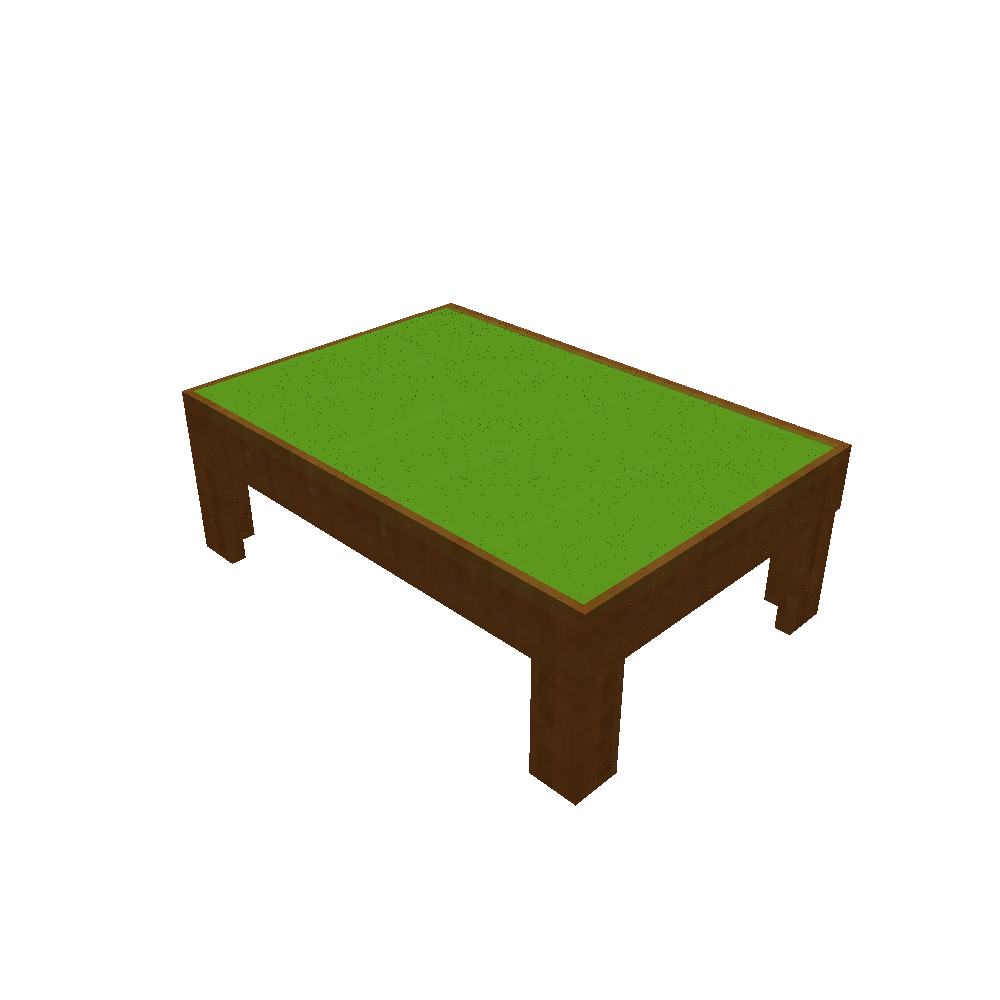} &
    \selectedshape{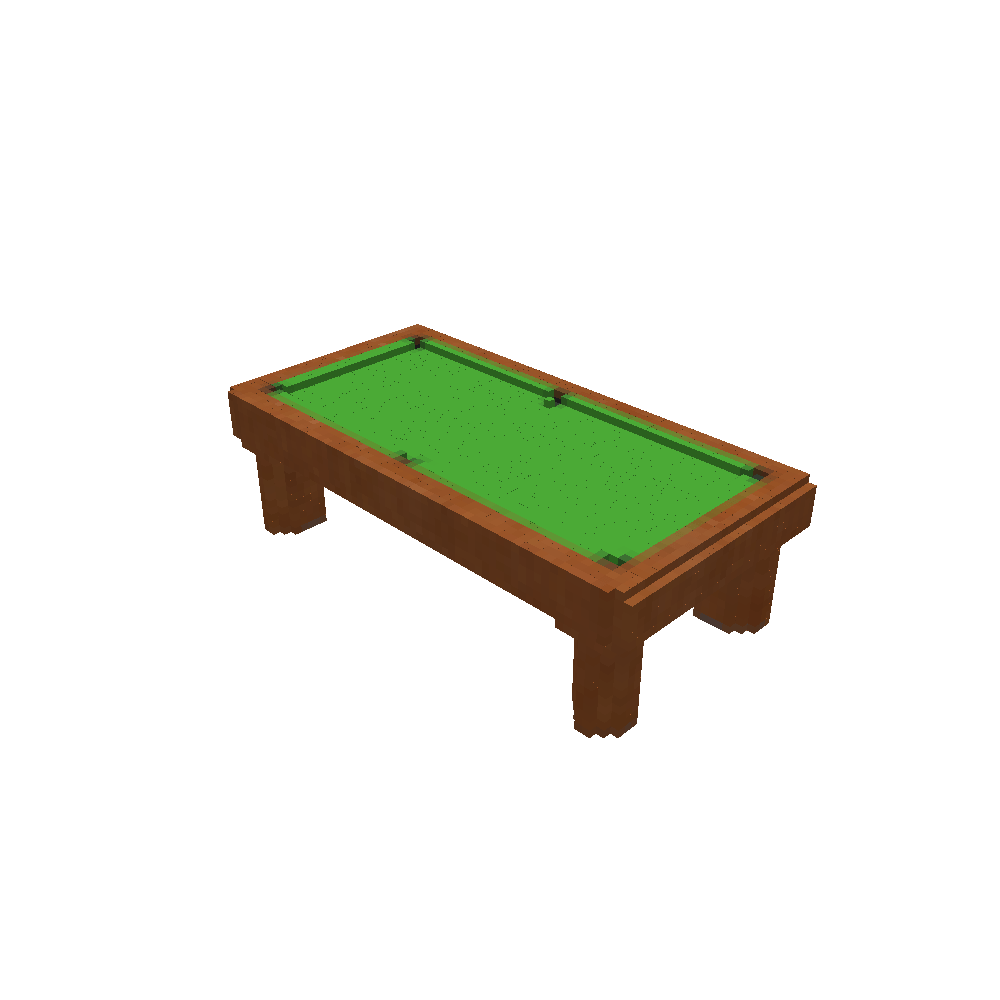} &
    \shape{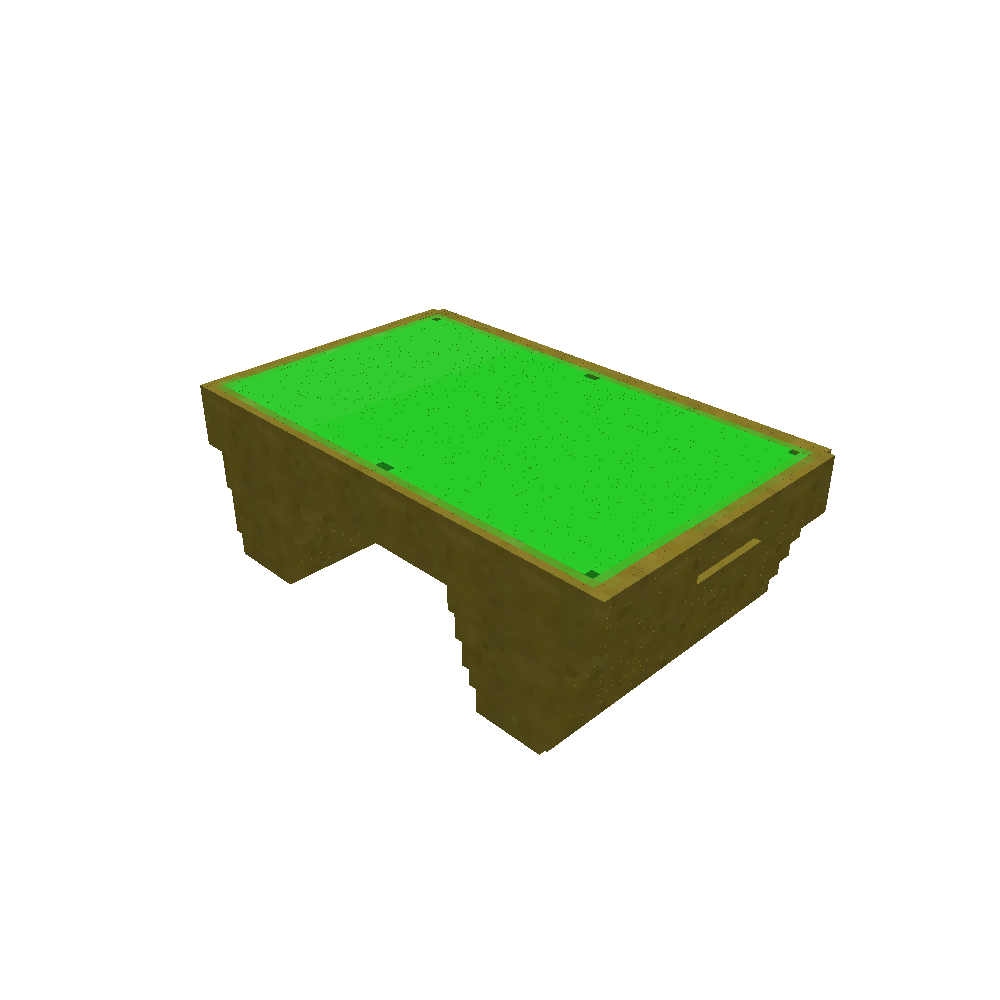} &
    \shape{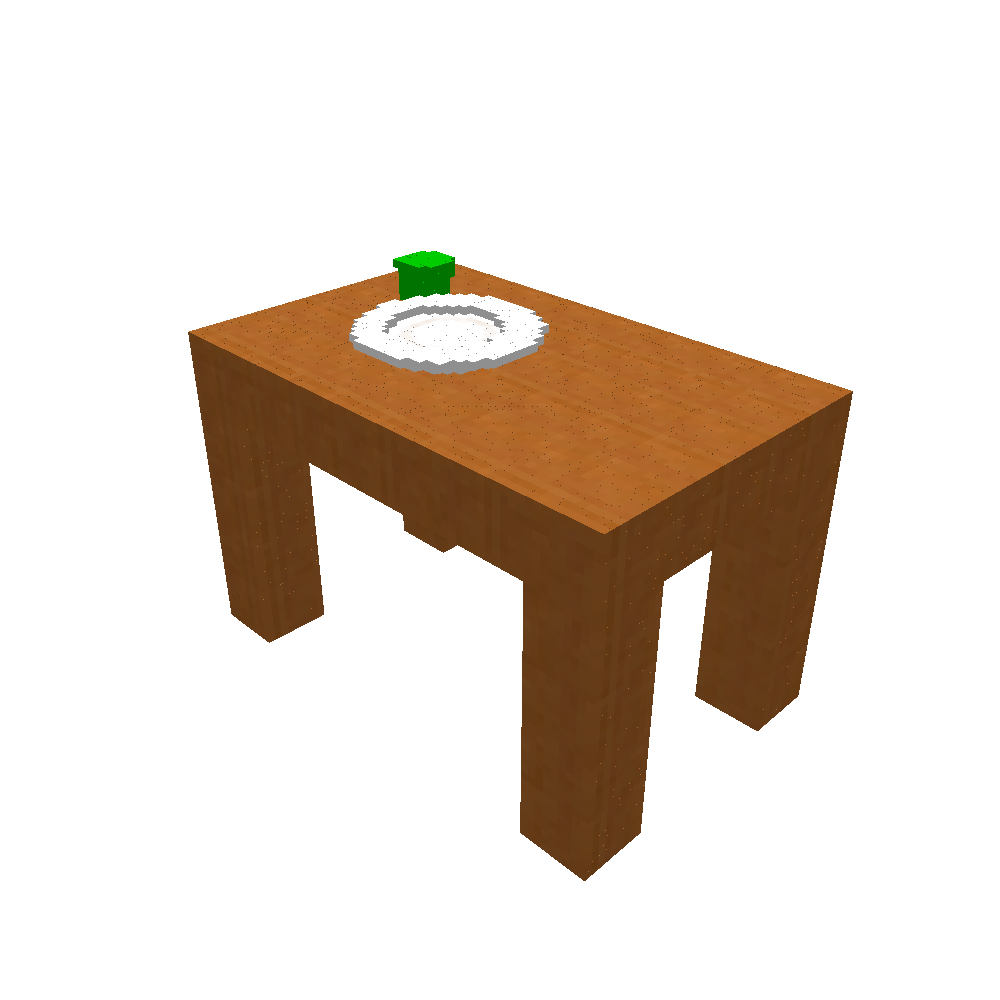} &
    \shape{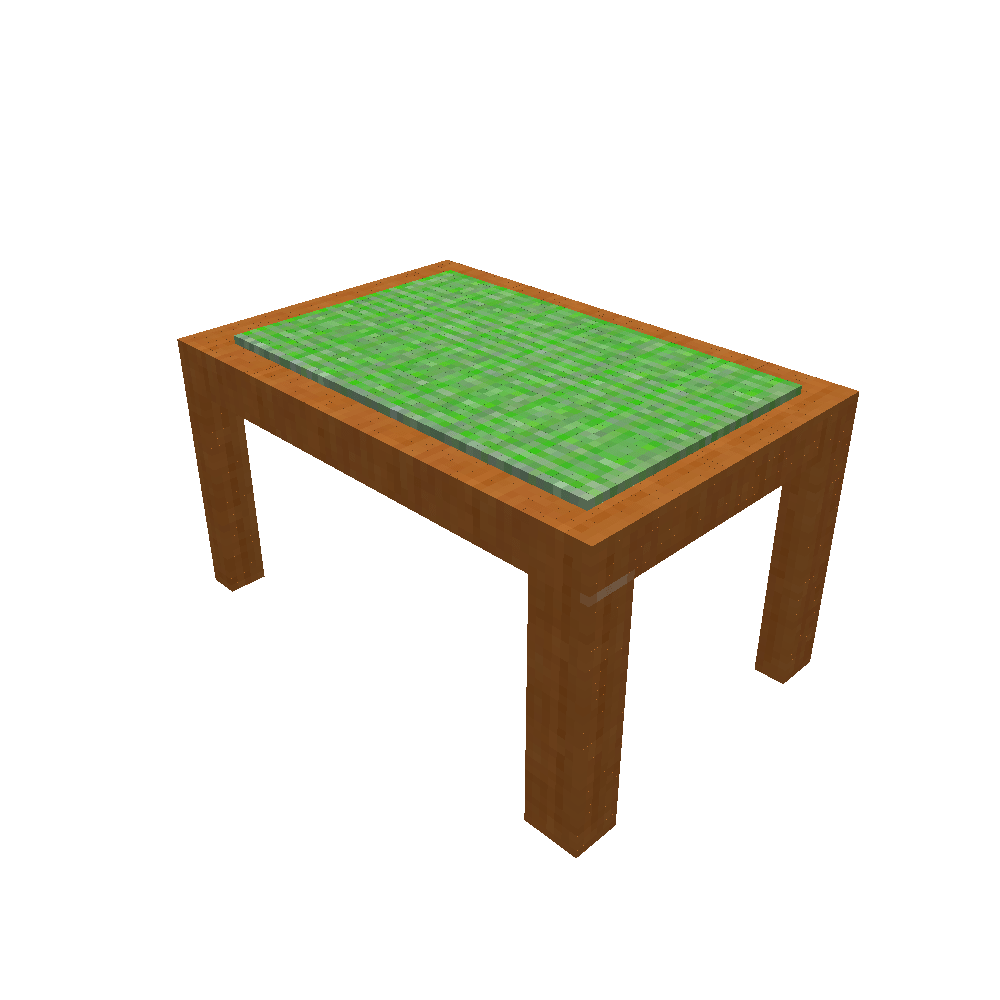} \vspace{0.5em} \\
    &
    \rotbox{Ours} &
    \selectedshape{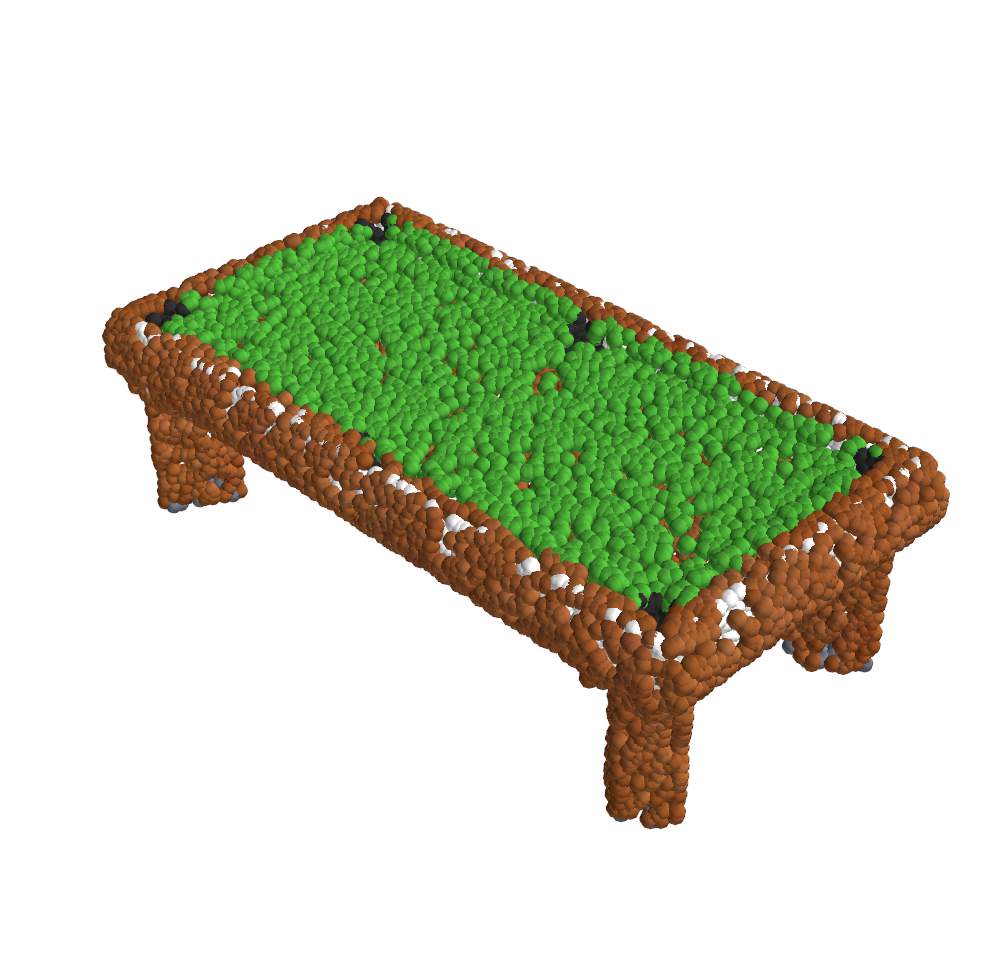} &
    \shape{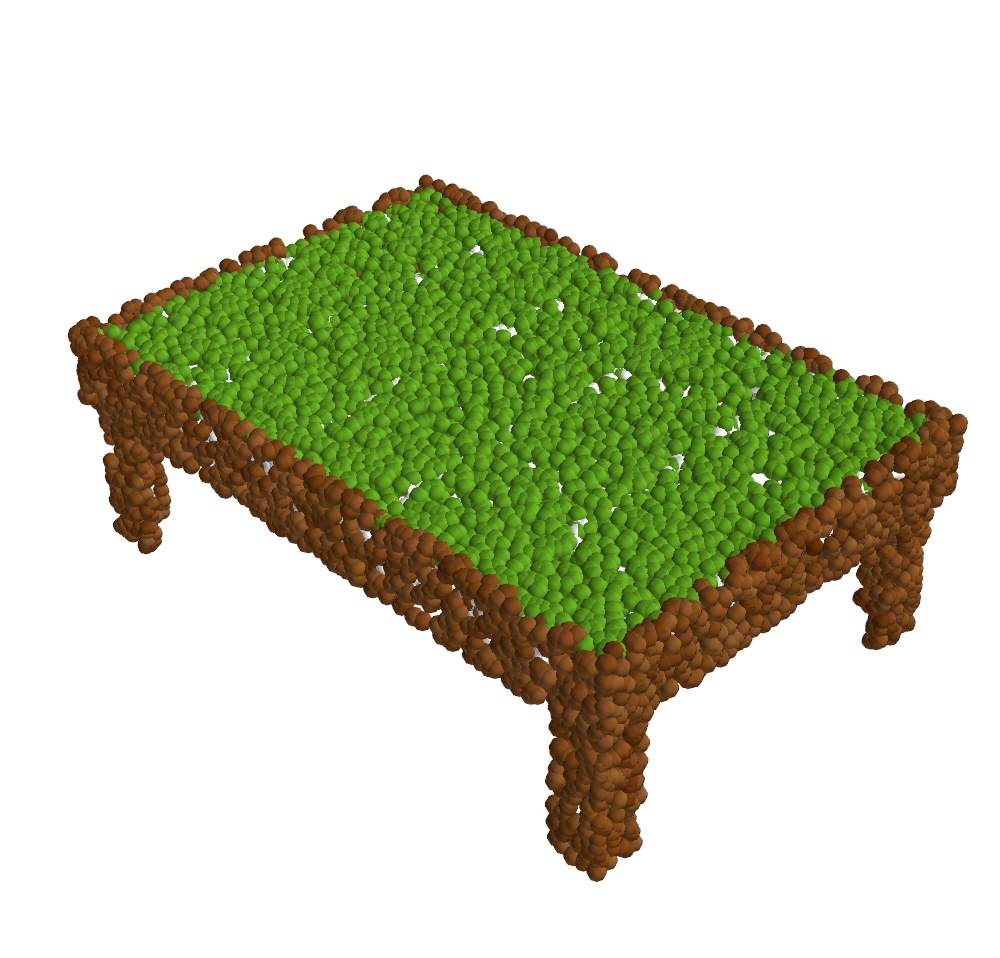} &
    \shape{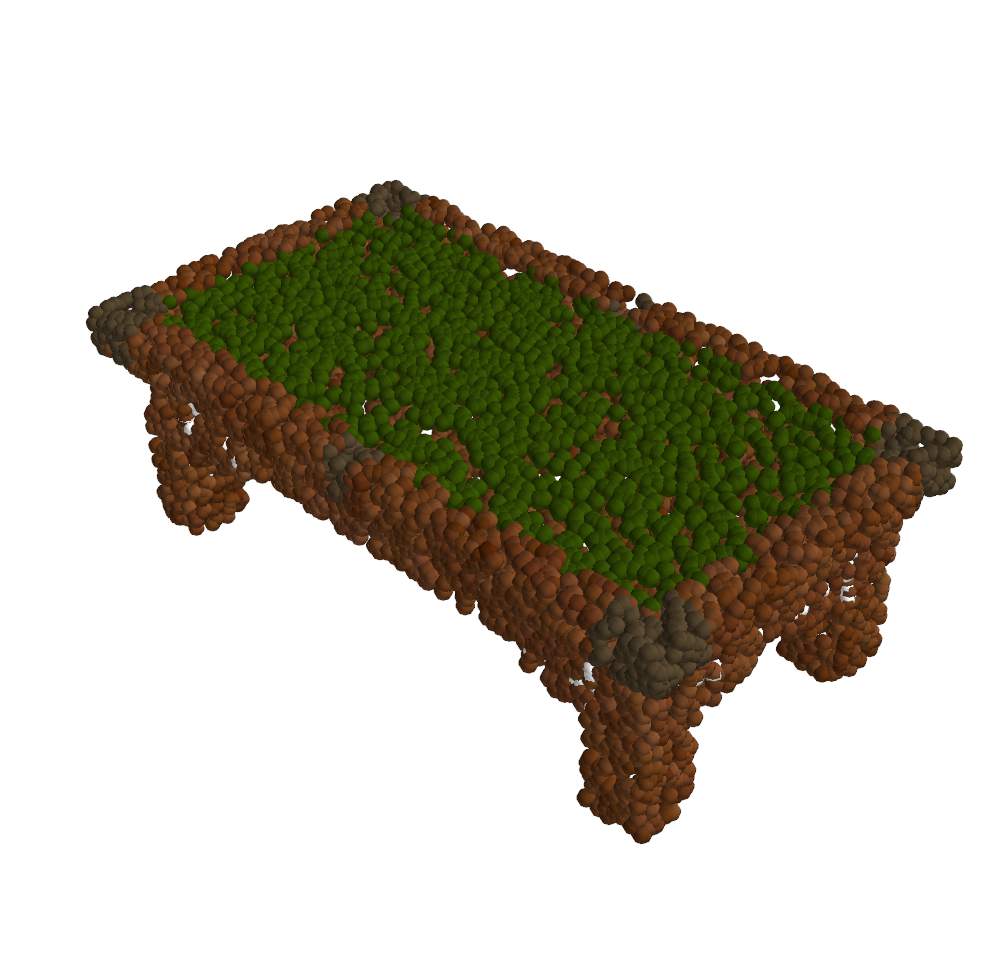} &
    \shape{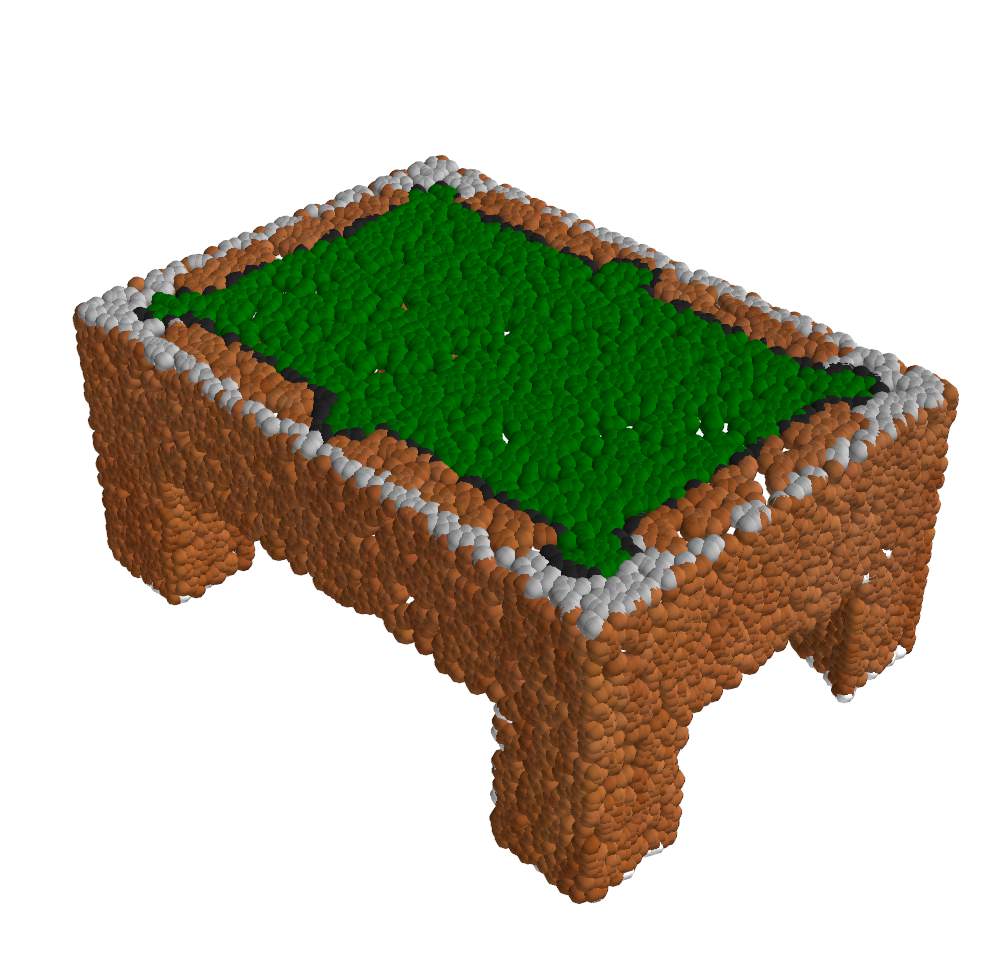} &
    \shape{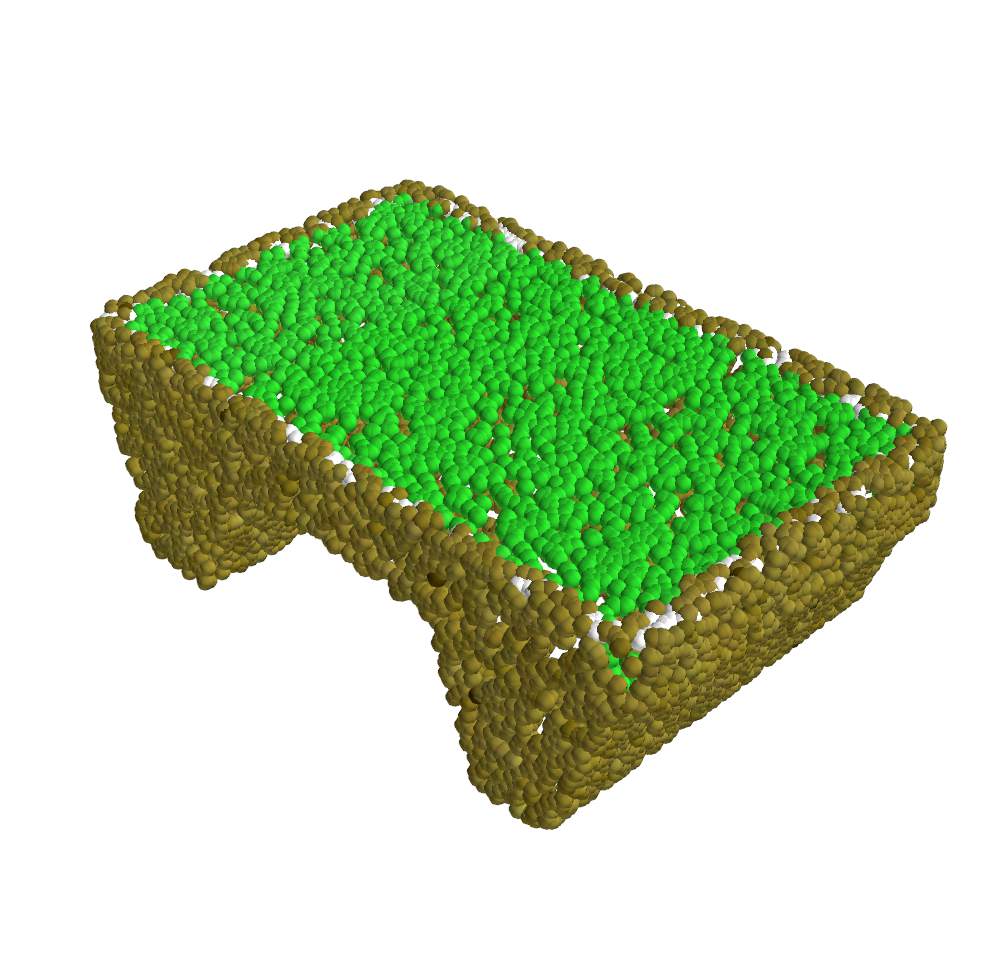} \\
    
    \midrule
    \textbox{\large{this is a boxy look \dashuline{gray} chair. It appears to be made out of granite and is gray with \dashuline{4 short legs} and a \dashuline{high, arched back}.}} &
    \rotbox{TriCoLo} &
    \selectedshape{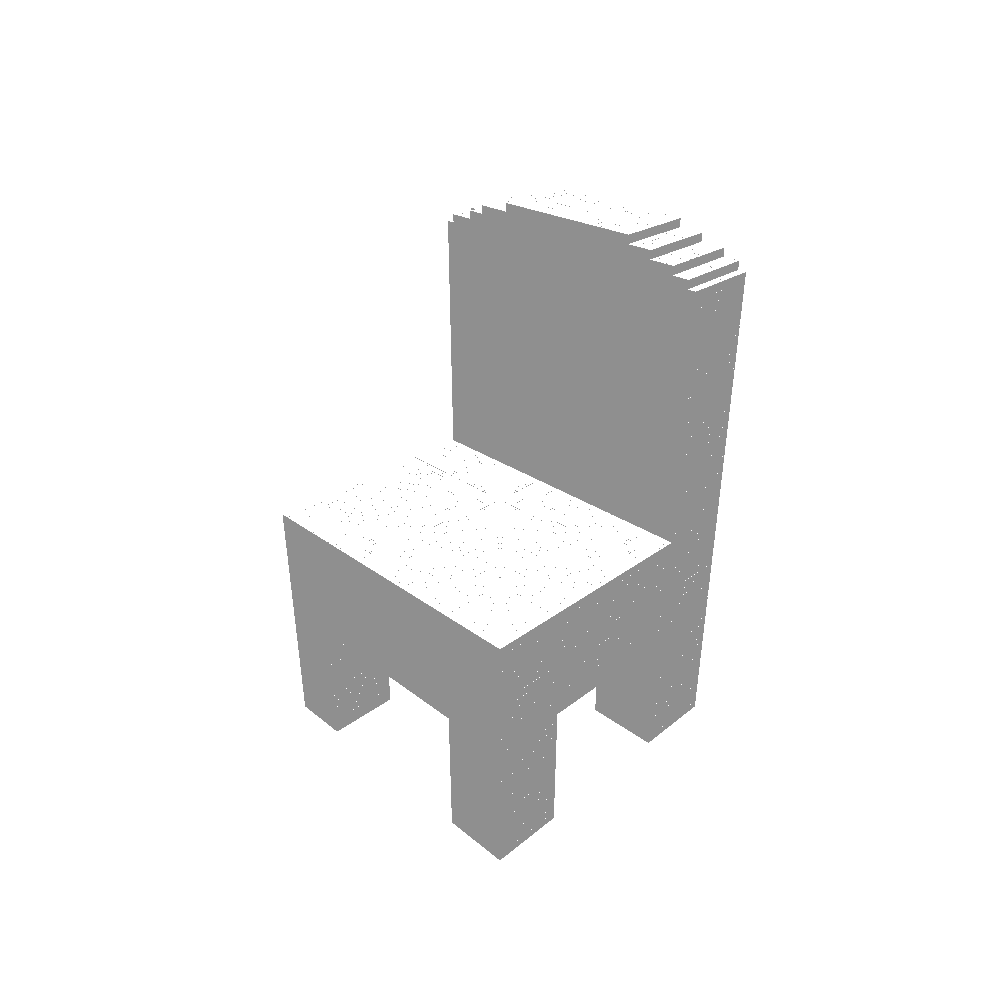} &
    \shape{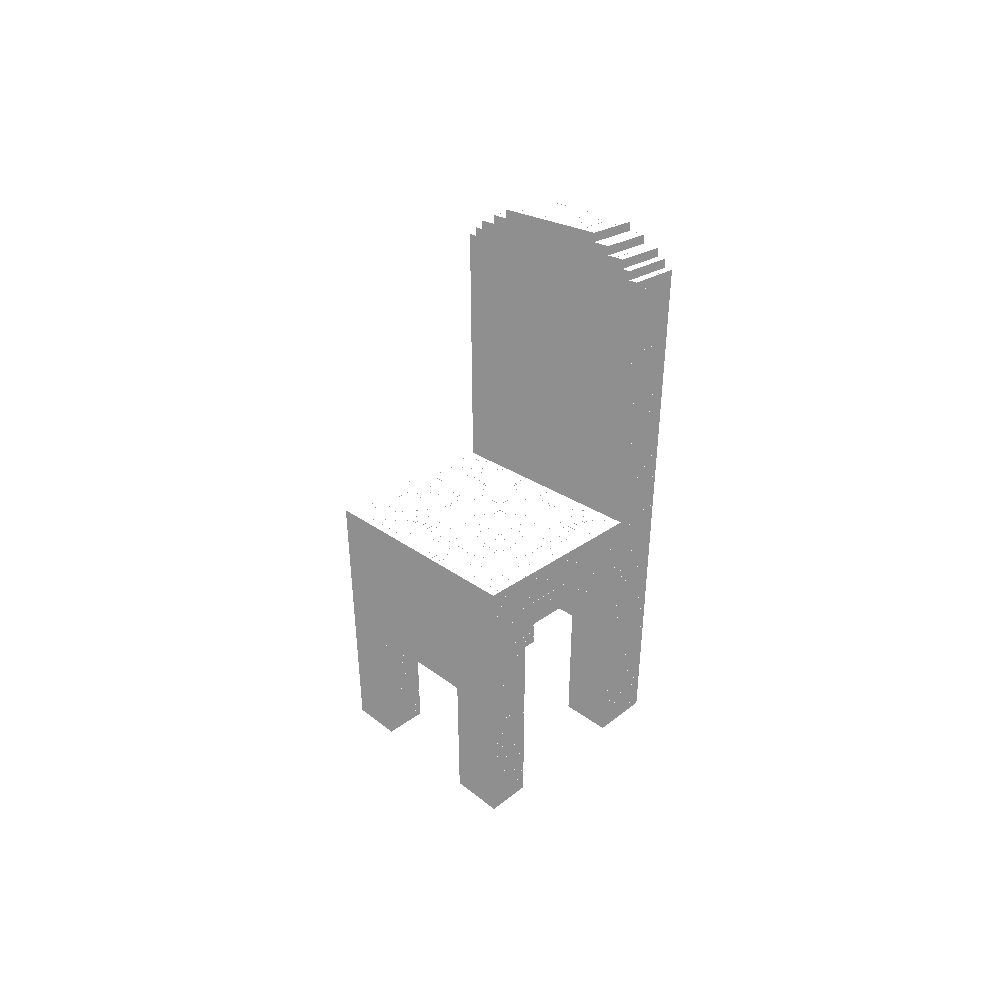} &
    \shape{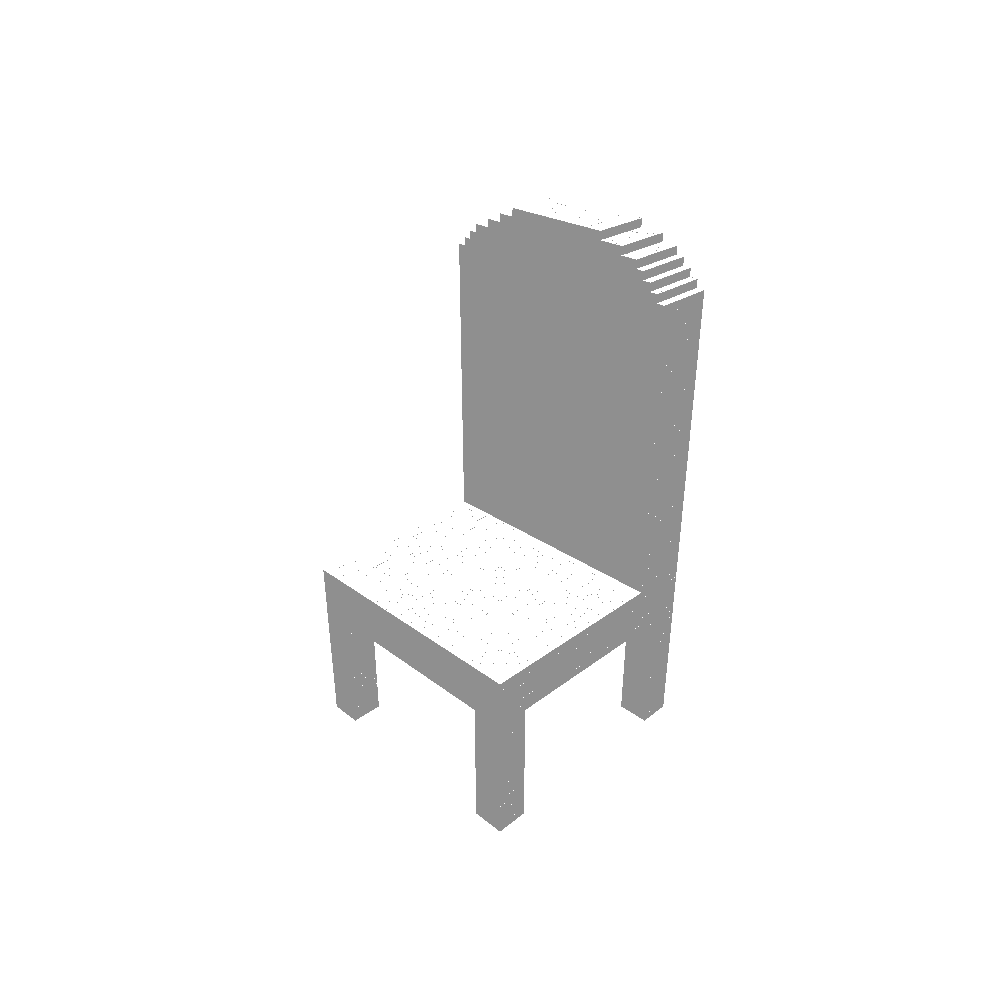} &
    \shape{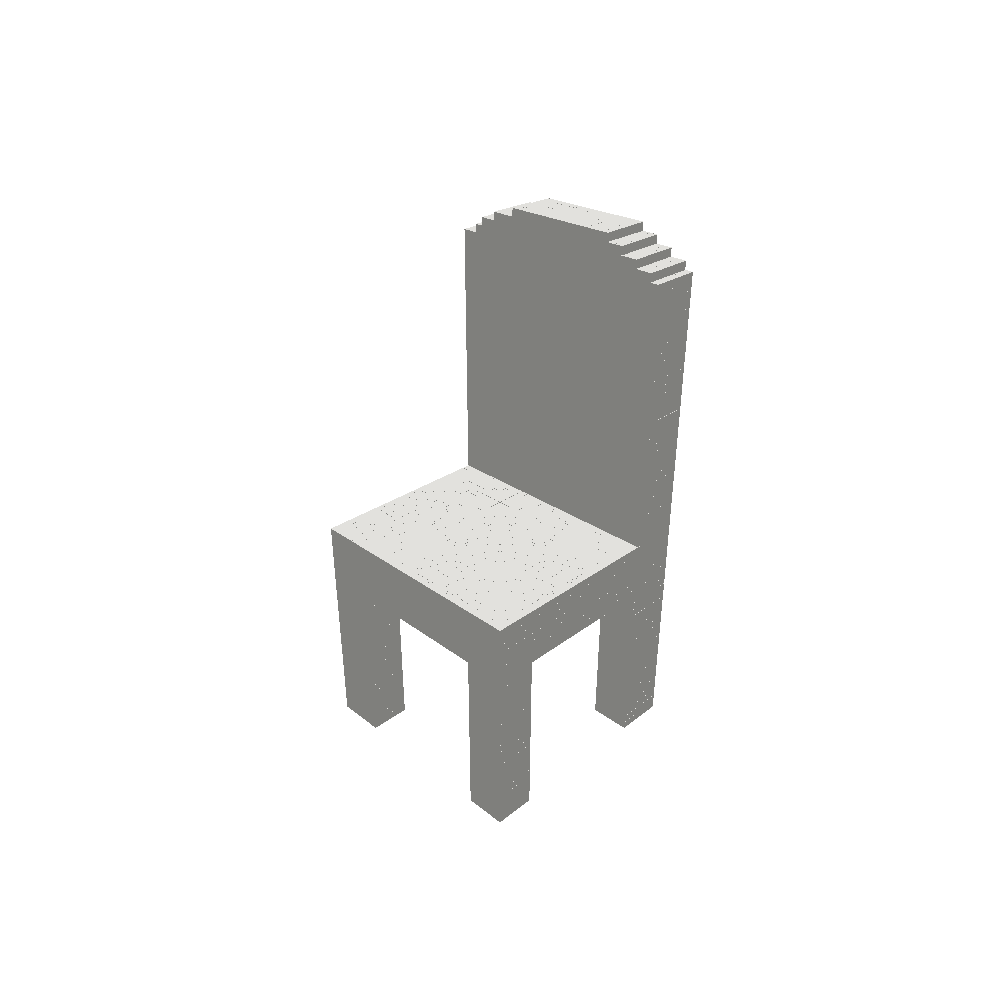} &
    \shape{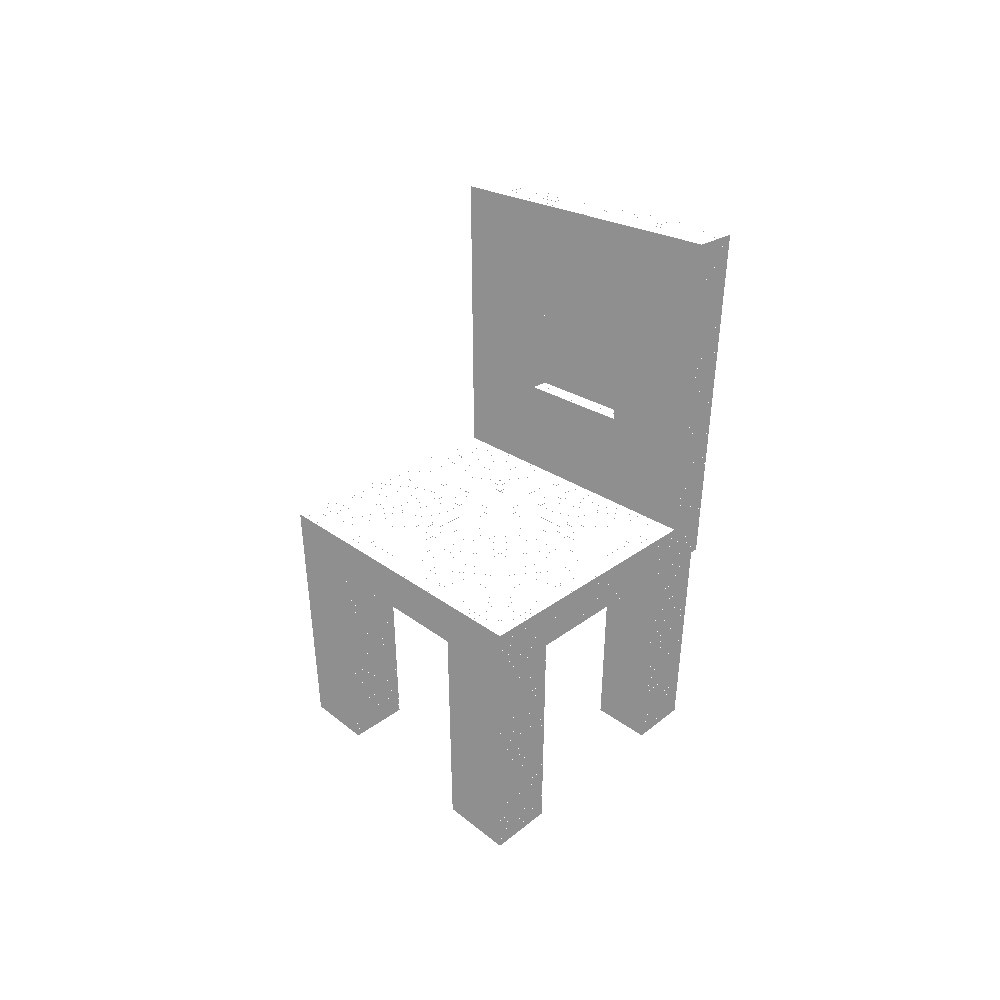} \vspace{0.5em} \\
    &
    \rotbox{Ours} &
    \shape{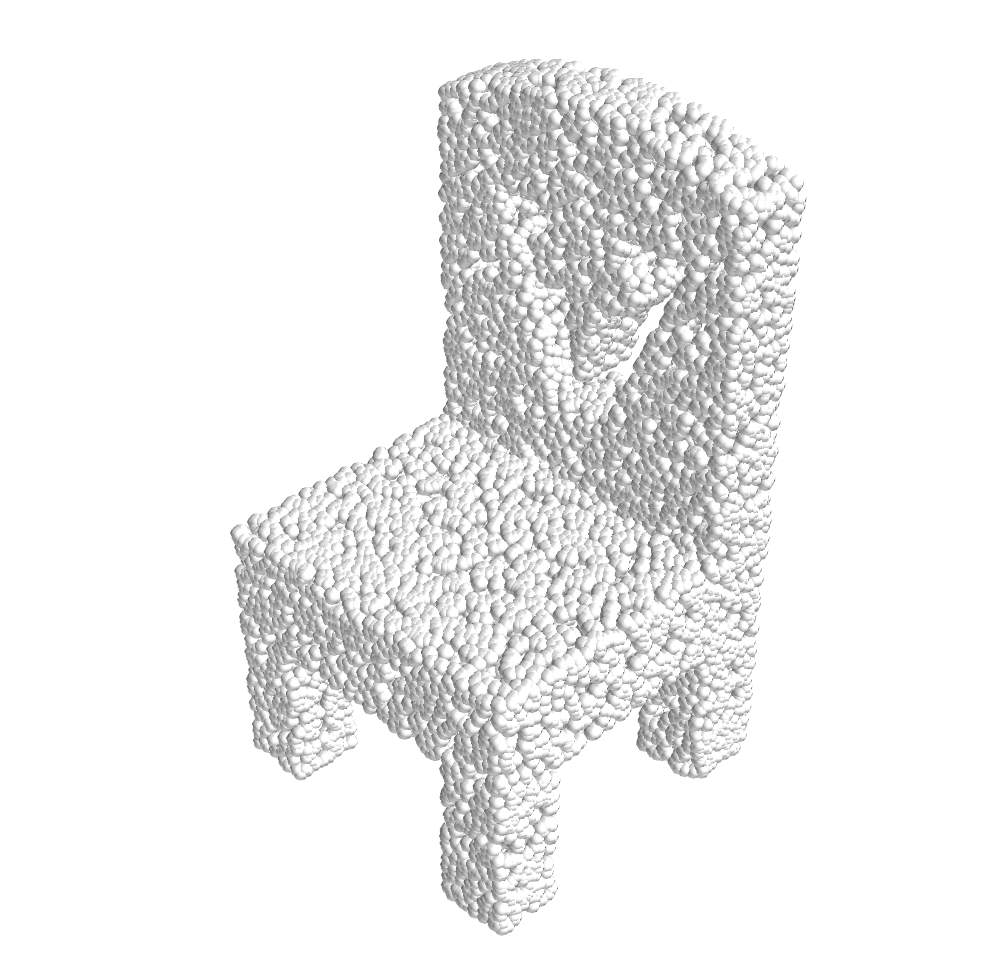} &
    \selectedshape{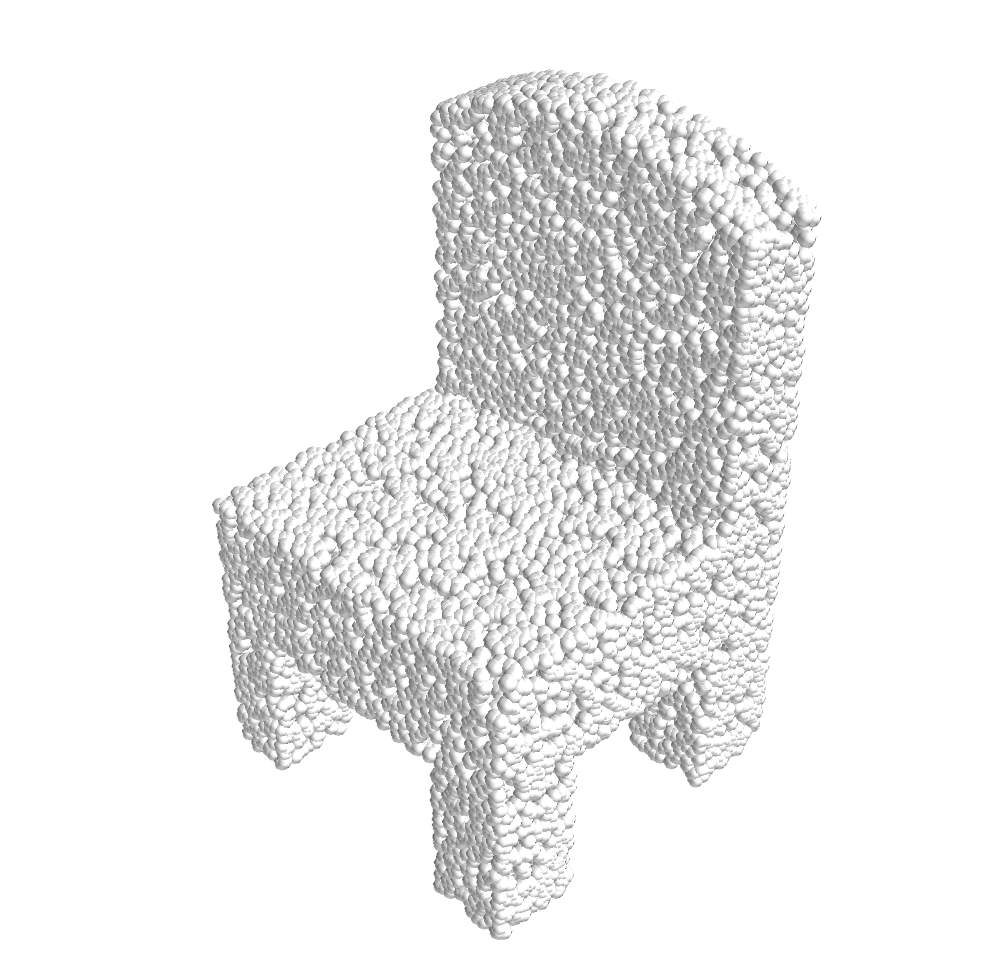} &
    \shape{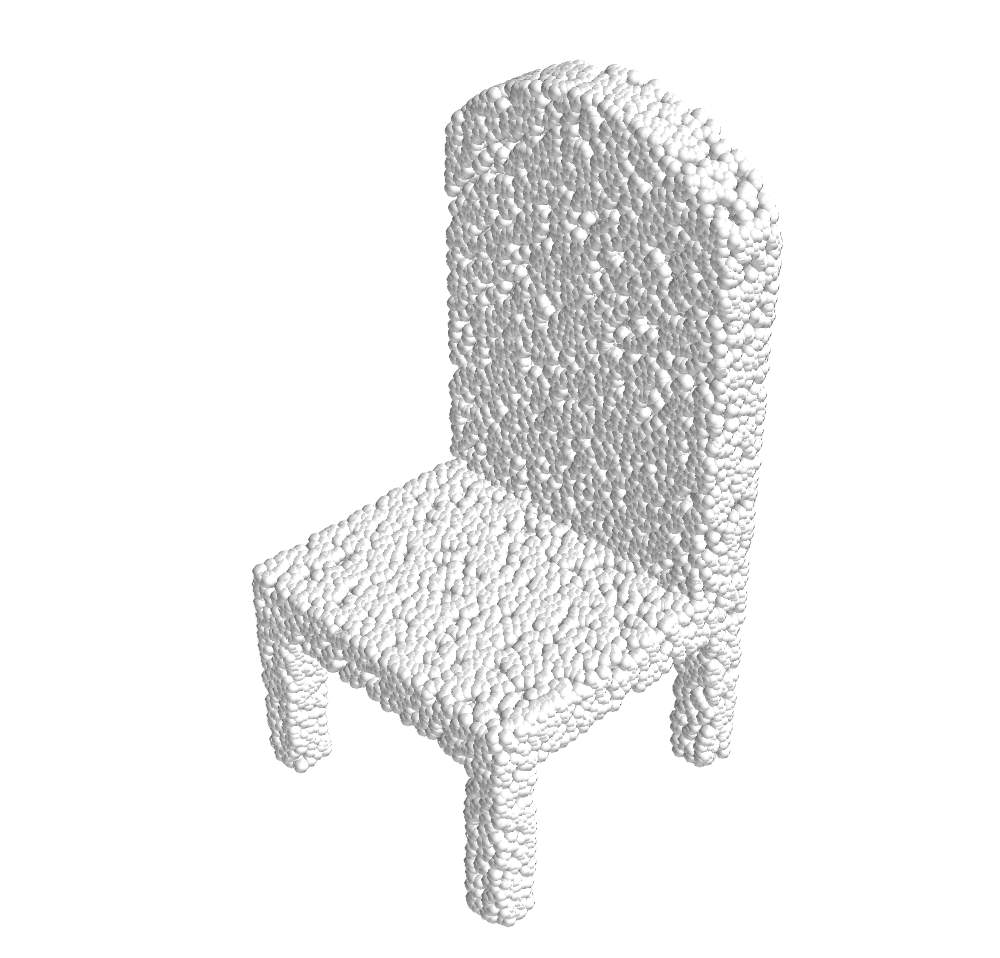} &
    \shape{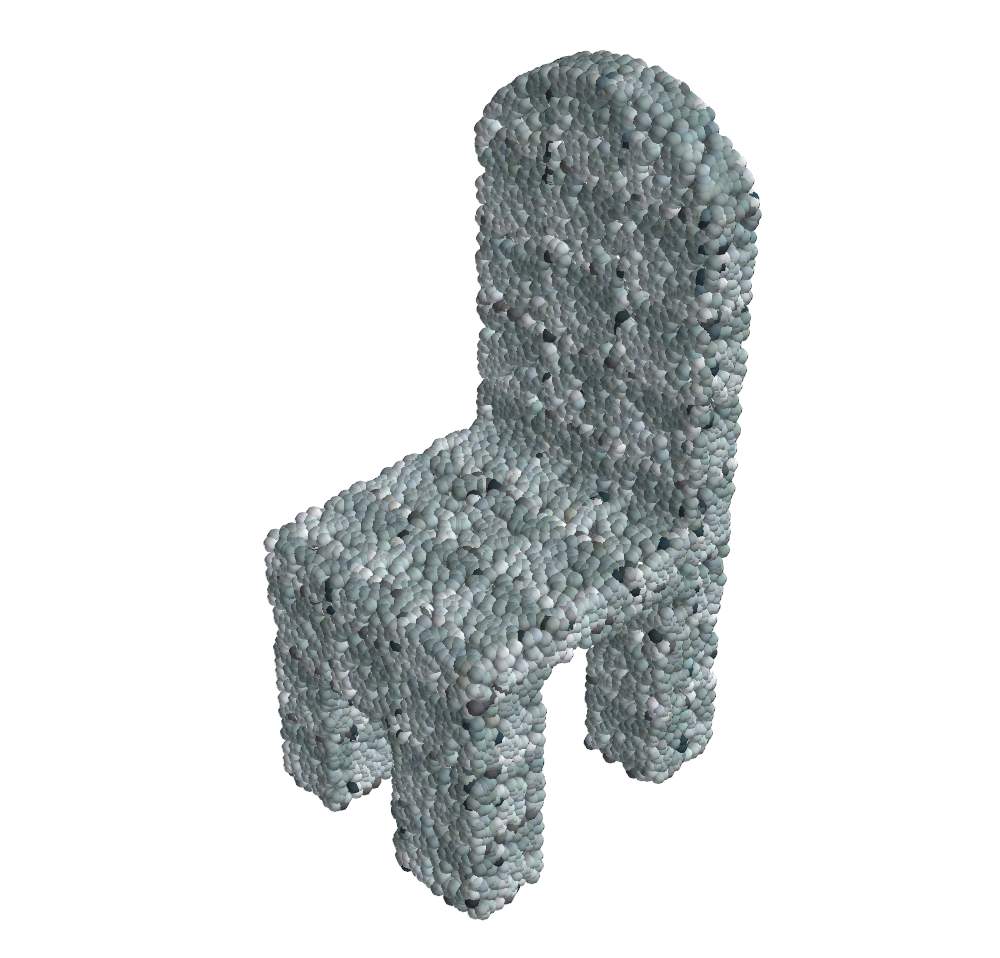} &
    \shape{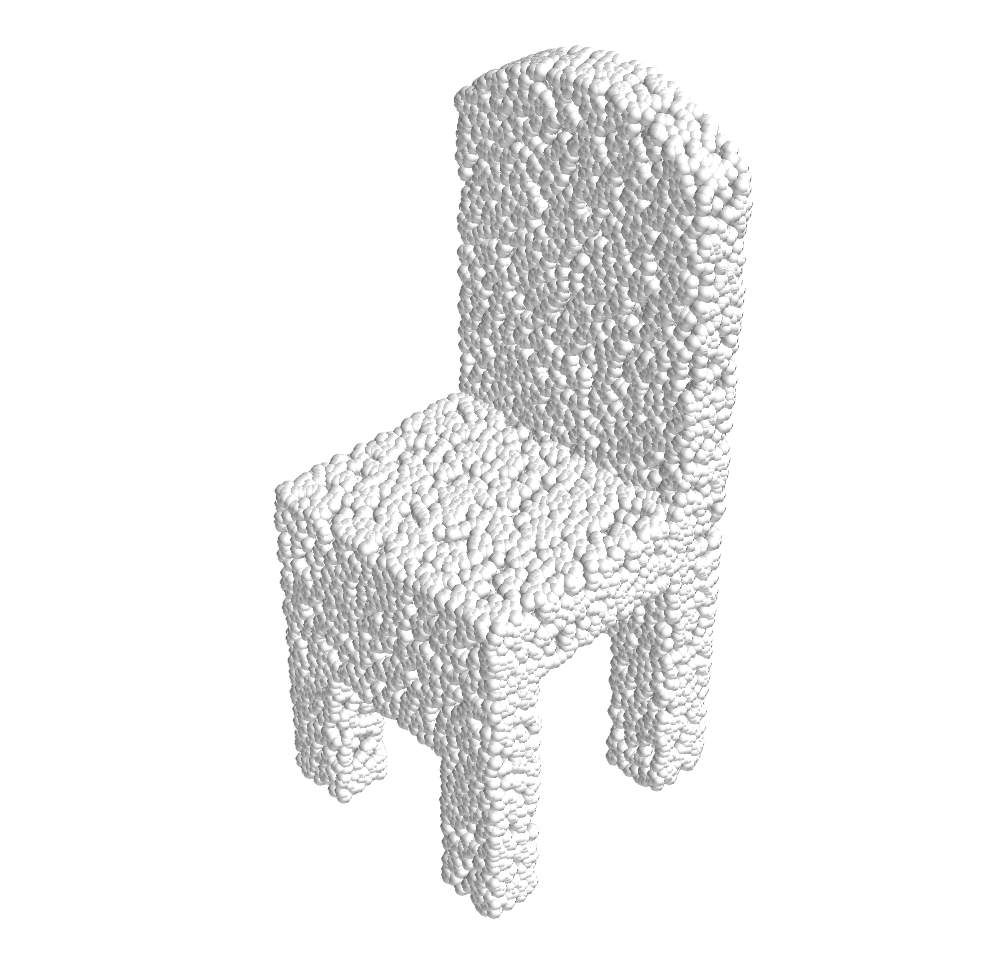} \\
    
    \bottomrule
\end{tabular}}
\caption{Retrieval result of the proposed Parts2Words and TriCoLo~\cite{TriCoLo} (T2S). the ground truth shape are marked as \textcolor{red}{red boxes}. Words corresponding with the parts in retrieved shapes are marked in \dashuline{bold}.}
\label{fig:cmp_t2s}
\end{figure}


\begin{figure*}\scriptsize
\centering
\setkeys{Gin}{width=\linewidth}
\begin{tabular}{p{0.11\linewidth}p{0.4\linewidth}p{0.4\linewidth}}
\toprule
\normalsize{Query Shape} & \normalsize{Parts2Words} & \normalsize{TriCoLo} \\
\midrule
\makecell[{c}{p{\linewidth}}]{\includegraphics[trim=50 120 50 150,clip]{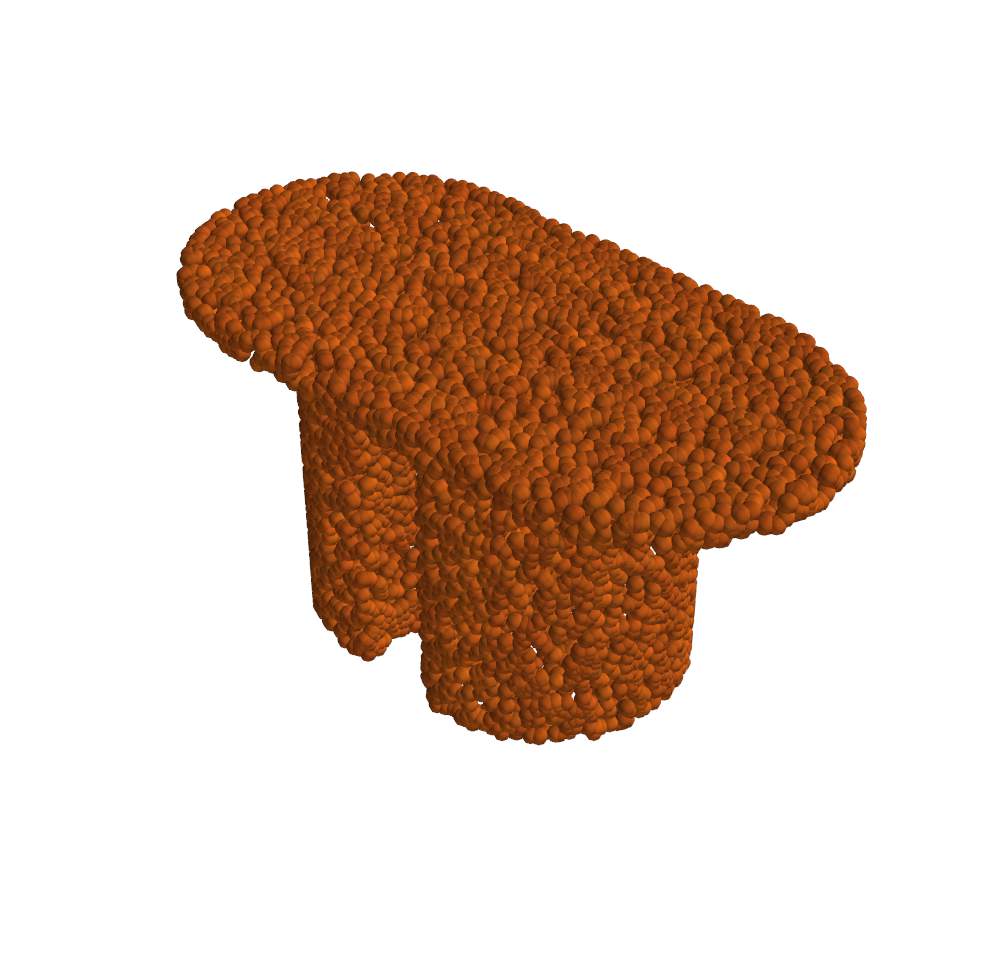}}
& 
\makecell[{c}{p{\linewidth}}]{
\scriptsize{\textcolor{red}{1. a wooden oval brown small table . it have a rectangular hole at the middle below the table - top , seem like it have two leg .}} \\
\scriptsize{\textcolor{red}{2. a capsule shape wooden table with two half cylinder shape leg .}} \\
\scriptsize{3. brown color , oval shape , wood material , and physical appearance table} \\
\scriptsize{\textcolor{red}{4. this be a light brown wooden table , with a flat capsule shape surface at the top , and two thick leg that be vertically tall , and be semicircle shape with the flat side face the middle create an empty space .}} \\
\scriptsize{5. brown colored whole wood oval shape coffee table .}
}
& 
\makecell[{c}{p{\linewidth}}]{
\scriptsize{1. this oval light wood topped table is on a dark wood base .} \\
\scriptsize{\textcolor{red}{2. a wooden oval brown small table . it has a rectangular hole at the middle below the table top seems like it has two legs .}}\\
\scriptsize{3. an oval shaped table with two legs . it is also wooden and brown .} \\
\scriptsize{4. an brown oval table with three section base} \\
\scriptsize{5. brown color rectangle shape wood material and physical appearance table}
}

\\
\midrule
\makecell[{c}{p{\linewidth}}]{ \includegraphics[]{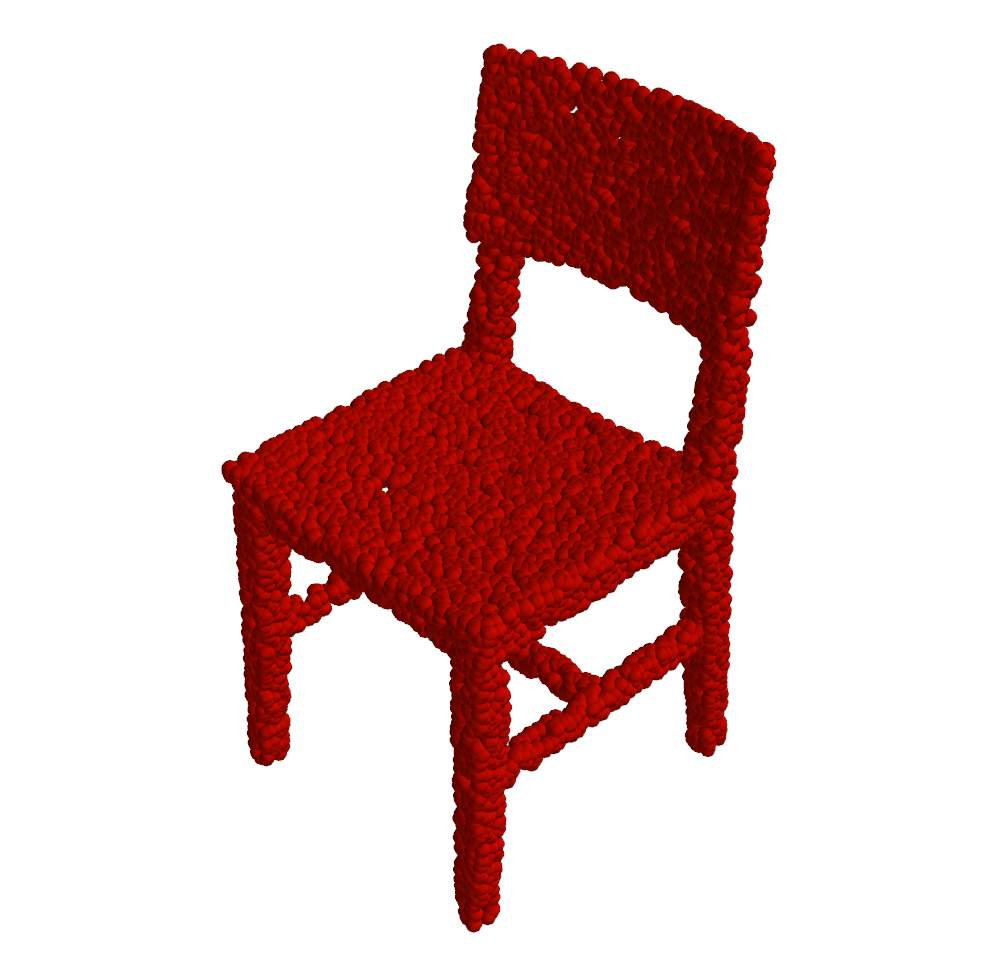}}
&
\makecell[{c}{p{\linewidth}}]{
\scriptsize{1. a red chair with a back .} \\
\scriptsize{2. a red chair .   the seat be curved downward and the back have a gap .} \\
\scriptsize{3. red side chair} \\
\scriptsize{\textcolor{red}{4. maroon color chair with four leg and rest at back}} \\
\scriptsize{\textcolor{red}{5. it be a wooden chair . it be red in color .}}
}
& 
\makecell[{c}{p{\linewidth}}]{
\scriptsize{1. a wooden chair red in color} \\
\scriptsize{\textcolor{red}{2. it is a wooden chair . it is red in color .}} \\
\scriptsize{3. a wooden chair with red colour back and seat with spindle and strong four legs} \\
\scriptsize{4. a red wooden kitchen chair with detached back and slightly rounded seat} \\
\scriptsize{5. this is wooden chair with four legs and it is in red texture light weight} \\
}
\\
\midrule
\makecell[{c}{p{\linewidth}}]{ \includegraphics[trim=50 120 50 150,clip]{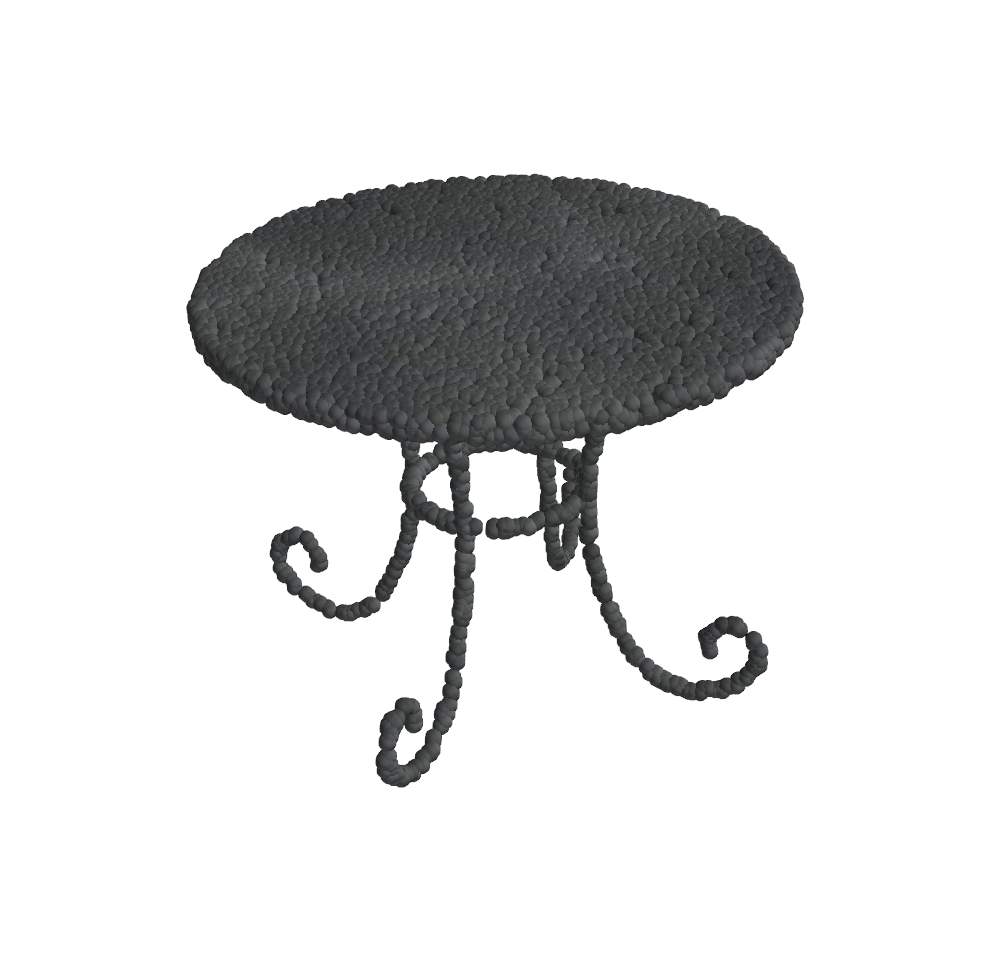}}
& 
\makecell[{c}{p{\linewidth}}]{
\scriptsize{1. marble round table with metal leg . table be black color .}  \\
\scriptsize{\textcolor{red}{2. black , round metal outdoor table . with long curl leg .}} \\
\scriptsize{\textcolor{red}{3. a center - table with black circular top and four curved leg}} \\
\scriptsize{4. a stylish black color round table for all - purpose} \\
\scriptsize{5. a circular black table .}
}

& 
\makecell[{c}{p{\linewidth}}]{
\scriptsize{1. the table is circular with three legs . the table is black and the legs stick out from the top.} \\
\scriptsize{2. a black color round shaped wooden table with three legs} \\
\scriptsize{3. a black colored round table with four slim shaped legs} \\
\scriptsize{\textcolor{red}{4. black round metal outdoor table with long curled legs .}} \\
\scriptsize{5. black round table three legs wooden material} \\
}
\\
\bottomrule
\end{tabular}
\caption{Retrieval result of the proposed Parts2Words and TriCoLo~\cite{TriCoLo}, the ground truth description are marked in \textcolor{red}{red}. Each retrieval case under our model can match 2 or 3 ground truth sentences, which is more than TriCoLo.}
\label{fig:cmp_s2t}
\end{figure*}

\begin{figure}[t]
    \centering
    \includegraphics[width=\linewidth]{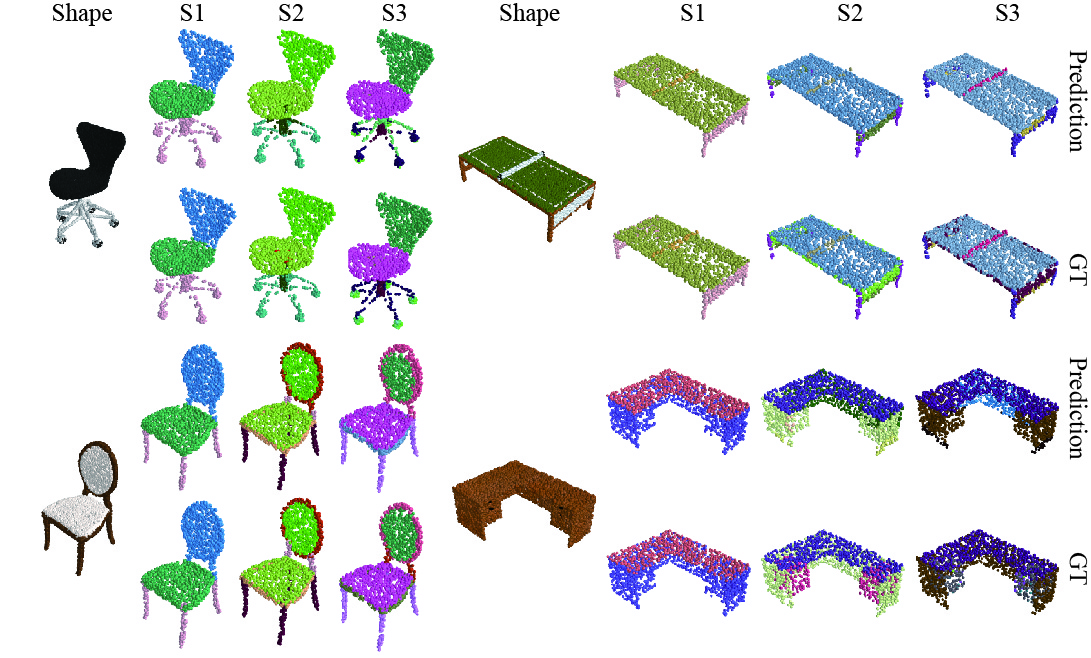}
    \caption{Segmentation results in three different granularity.}
    \label{fig:seg}
\end{figure}

\begin{figure}[htbp]
    \begin{center}
    \includegraphics[width=0.98\linewidth]{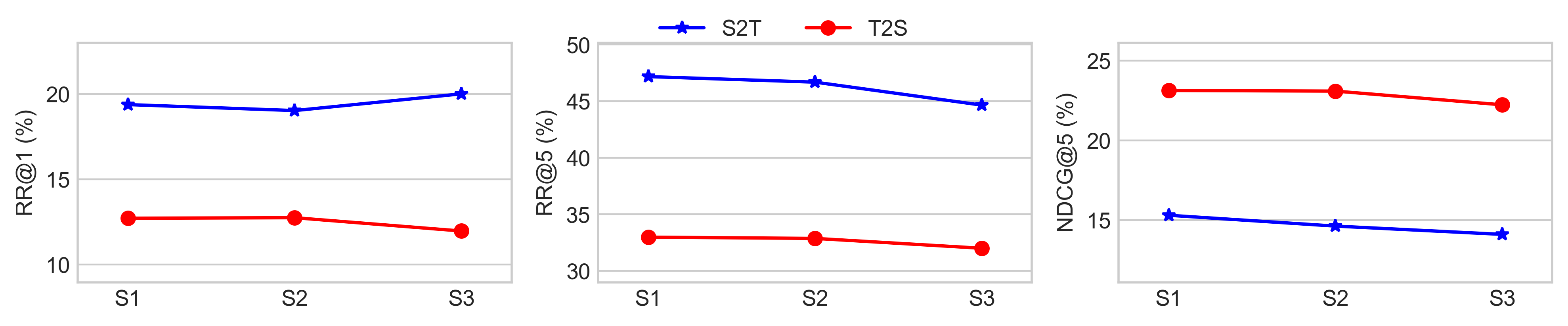}
    \end{center}
    \vspace{-0.5cm}
    \caption{Comparison in terms of Recall@1(RR@1), Recall@5(RR@5) and NDCG@5 using different segmentation granularity. S1, S2, and S3 represent the representing different segmentation granularities from coarse to fine, which have 17 classes, 72 classes, and 90 classes separately. The segmentation with the coarsest granularity \textbf{S1} achieves a better performance. }
    \label{fig:s}
    \vspace{-0.2cm}
\end{figure}

\begin{figure}[htbp]
    \begin{center}
    \includegraphics[width=0.98\linewidth]{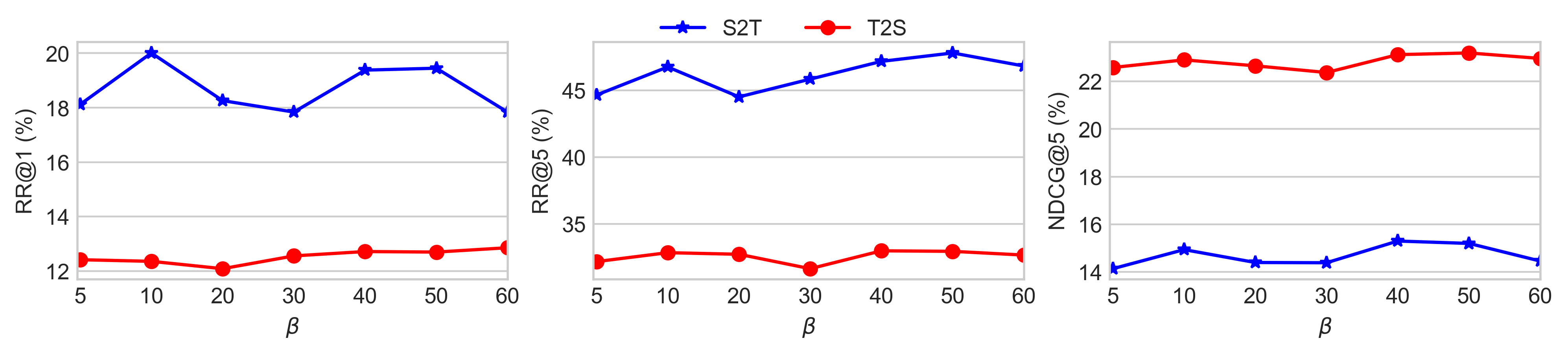}
    \end{center}
    \vspace{-0.5cm}
    \caption{Comparison in terms of Recall@1(RR@1), Recall@5(RR@5) and NDCG@5 using different loss weight $\beta$. According to the above result, we choose loss weight $\beta$ as 40.}
    \label{fig:beta}
\end{figure}

\begin{table}\centering
\caption{Component-wise analysis on text2shape, with different backbones on shape encoder and different matching modules.}
\label{cmp:ablation study 01}

\centering
\resizebox{0.98\columnwidth}{!}{%
\begin{tabular}{cc|ccc|ccc} 
\hline
\multicolumn{2}{c|}{{}} & \multicolumn{3}{c|}{S2T} & \multicolumn{3}{c}{T2S} \\
Backbones & Matching & RR@1 & RR@5 & NDCG@5 & RR@1 & RR@5 & NDCG@5 \\
\hline
PN++ & EMD  & 16.66     & 43.11    & 13.20  & 11.10     & 28.89    & 20.25 \\
PN & CD & 18.47     & 45.98    & 14.52    & 11.73     & 32.60    & 22.49    \\
PN & EMD & \textbf{19.38}     & \textbf{47.17}    & \textbf{15.30}  & \textbf{12.72}     & \textbf{32.98}    & \textbf{23.13}\\
\hline
\end{tabular}}
\end{table}
\begin{table}\centering
\caption{Effectiveness of end-to-end training and semi-hard negative mining strategies.}
\label{cmp:ablation study 02}
\vspace{-0.1cm}
\centering
\resizebox{0.98\columnwidth}{!}{%
\begin{tabular}{cc|ccc|ccc} 
\hline
\multicolumn{2}{c|}{{}} & \multicolumn{3}{c|}{S2T} & \multicolumn{3}{c}{T2S} \\
w e2e & w semi & RR@1 & RR@5 & NDCG@5 & RR@1 & RR@5 & NDCG@5 \\
\hline
 & \checkmark & 16.31    & 43.39    & 12.99    & 10.68    & 29.25    & 20.14   \\ 
\checkmark &  & 0.00     & 0.21    & 0.04  & 0.08     & 0.35    & 0.23    \\
\checkmark  & \checkmark &  \textbf{19.38}     & \textbf{47.17}    & \textbf{15.30}  & \textbf{12.72}     & \textbf{32.98}    & \textbf{23.13}\\
\hline
\end{tabular}}
\end{table}
\begin{table}\centering
\caption{Ablation study on the influence of color and segmentation supervision.}
\label{cmp:ablation study 03}
\vspace{-0.1cm}
\centering
\resizebox{0.98\columnwidth}{!}{%
\begin{tabular}{c|ccc|ccc} 
\hline
\multicolumn{1}{c|}{{}} & \multicolumn{3}{c|}{S2T} & \multicolumn{3}{c}{T2S} \\
 & RR@1 & RR@5 & NDCG@5 & RR@1 & RR@5 & NDCG@5 \\
\hline
+COLOR   & 7.77    & 26.94    & 6.91   & 5.06    & 17.21    & 11.25   \\
+SEG  & 11.62    & 29.18    & 8.53   & 7.58    & 21.93    & 14.91   \\
Parts2Words (COLOR+SEG)  & \textbf{19.38}     & \textbf{47.17}    & \textbf{15.30}  & \textbf{12.72}     & \textbf{32.98}    & \textbf{23.13}\\
\hline
\end{tabular}}
\end{table}

\subsection{Comparison results}

Table~\ref{tab:cmp_sota} presents the quantitative results on 3D-Text dataset where our method outperforms the latest approaches Text2Shape~\cite{chen2018text2shape},  $\rm{Y^{2}Seq2Seq}$~\cite{han2019y2seq2seq} and TriCoLo~\cite{TriCoLo} in all measures. We can observe that our proposed model outperforms other methods on the 3D-Text dataset. Our best result at RR@1 are 19.38 and 12.72 for shape-to-text (S2T) retrieval and text-to-shape (T2S) retrieval, 
which achieves a $12.64\%$ relative improvement compared to current SOTA methods.

Additionally, to make our experiment more comparable and convincing, we also conducted four additional experiments by modifying our proposed network to approximate other existing methods. The results are also presented in Figure~\ref{tab:cmp_sota}.
We designed two global-based matching networks, Global-Max and Global-Avg, to show the performance of point-based global matching, which simply use max pooling (Global-Max) and average pooling (Global-Avg) operation on both features of all points and embeddings of all words separately. 
Besides, we also presented a local-based matching network called LocalBaseline, by replacing our matching module with the stack cross attention module (SCAN)~\cite{lee2018stacked} in image-text matching. We train this network by two approaches: LocalBaseline is a two-stage training approach based on a pre-trained segmentation network, identical to the SCAN which freezes the detector model to produce features from image patches; LocalBaseline+ is an end-to-end training approach, identical to our proposed method.
We achieved $1.61$ and $0.78$ improvements in terms of RR@1 by comparing with LocalBaseline+. The comparison with four modified networks proved that the local-based matching approaches extract more detailed features than the global-based ones, our optimal transport-based matching module learns better joint embeddings than the popular SCAN module, and the proposed end-to-end multi-task learning approach makes a better connection between matching embeddings and segmentation prior information.

As shown in Figure~\ref{fig:topk}, we plot the curve of RR@K of 5 methods, among which Parts2Words, Parts2Words-CD, and LocalBaseline+ are local-based methods, and Global-Avg and Global-Max are global-based methods. Parts2Words-CD was obtained by replacing EMD with CD (Chamfer Distance) on the basis of Parts2Words, and further analysis was conducted on this in subsequent ablation experiments. It can be seen that the recall rate of the local matching methods with segmentation prior is obviously higher than the global matching methods. Meanwhile, we can also observe that the proposed Parts2Words achieves the best results compared to other local matching models. 

The examples of T2S and S2T retrieval results are shown in Figure~\ref{fig:s2t} and Figure~\ref{fig:t2s}. 
From Figure~\ref{fig:s2t}, we display top-5 retrieved texts, we can see that our model can match an average of 2-3 ground truth texts within top-5 best-matched texts for each case. 
From Figure~\ref{fig:t2s}, we can see the top-5 retrieval shapes have a more similar appearance, especially in details such as color and geometry. 
The results demonstrate that our method could find correspondences in details, such as color and geometric description. In particular, for complex shapes, our model can still achieve superior results. 

We also compare with with TriCoLo~\cite{TriCoLo} by presenting 3 text-to-shape retrieval results and 3 shape-to-text retrieval results, as shown in Figures~\ref{fig:cmp_t2s} and~\ref{fig:cmp_s2t}. For each query text/shape, we display top-5 retrieval shapes/texts ranked by the similarity scores from the two methods. The ground truth shapes/texts are marked in red. And as shown in Figure~\ref{fig:cmp_t2s}, our model matches the ground truth shapes within top-2 retrieval shapes. For the S2T retrieval result in Figure~\ref{fig:cmp_s2t}, our model can match 2 or 3 ground truth sentences within top-5 retrieval results, which is more than the ones that TriCoLo can match.

\subsection{Ablation Study}

\textbf{Granularity} We explore the impact of part embedding extracted under different segmentation granularities on the matching model. 
The PartNet dataset contains hierarchical segmentation annotation in three granularities from \textbf{S1} to \textbf{S3}, representing different segmentation granularities from coarse to fine, which have 17 classes, 72 classes, and 90 classes separately. The ground truths and our semantic predictions are shown in Figure~\ref{fig:seg}.
As shown in Figure~\ref{fig:s}, the results show that our method with coarse part segmentation annotation achieves the best performance. We believe that the parts learned through finer segmentation ground truth are hard to align with the simple plain words.
Therefore, the coarsest part segmentation annotation \textbf{S1} is selected.

\textbf{Loss Weight} We adopt the balance weight $\beta$ to adjust the joint loss to make our network focus more on the retrieval performance than the segmentation result. We present our retrieval performance when selecting different $\beta$ to select the most suitable parameter settings, as shown in Figure~\ref{fig:beta}. The retrieval performance improves with the increase of the weight of retrieval loss until $\beta$ is 40. 

\textbf{Backbone} We analyzed the influence of different network backbones on the retrieval results. We use PointNet and PointNet++ respectively as the backbone for the feature extraction network. As shown in Table~\ref{cmp:ablation study 01}, the comparison results show that the retrieval result gets worse after replacing the backbone with PointNet++. 

\textbf{CD/EMD} In Table~\ref{cmp:ablation study 01}, we replace the EMD for matching similarity calculation by CD (Chamfer Distance), which can be regarded as a local optimal hard matching method. In the Chamfer distance matching flow, each node only corresponds to the node with the most similar individual. The experimental results show that the EMD is better than CD. 

\textbf{Sampling} We explore the impact of different negative sample learning strategies based on the triplet ranking loss on retrieval. We compared two strategies: hardest negative mining and semi-hard negative mining. As demonstrated in Table~\ref{cmp:ablation study 02}, we find that the model using the hardest negative mining results in a collapsed model. 

\textbf{Training} We examine the effectiveness of end-to-end training by joint multi-task learning as shown in Table~\ref{cmp:ablation study 02}. In comparison, we separately train the shape encoder module and the matching module. Our results demonstrate that the end-to-end model outperforms the separate training approach.

\textbf{Color and Segmentation} As shown in Table~\ref{cmp:ablation study 03}, we separately evaluate the influence of integrating the point cloud color into the feature fusion step and supervising the training with the segmentation loss. The results demonstrate that, after removing these two components (w/o color and w/o seg loss), it is hard for the network to learn to establish a local alignment between color and geometry, which will lead to a significant decrease in accuracy. It indicates that color and part information are crucial for multi-modal retrieval.

\subsection{Visualization}

\begin{figure}[t]
    \includegraphics[width=\linewidth]{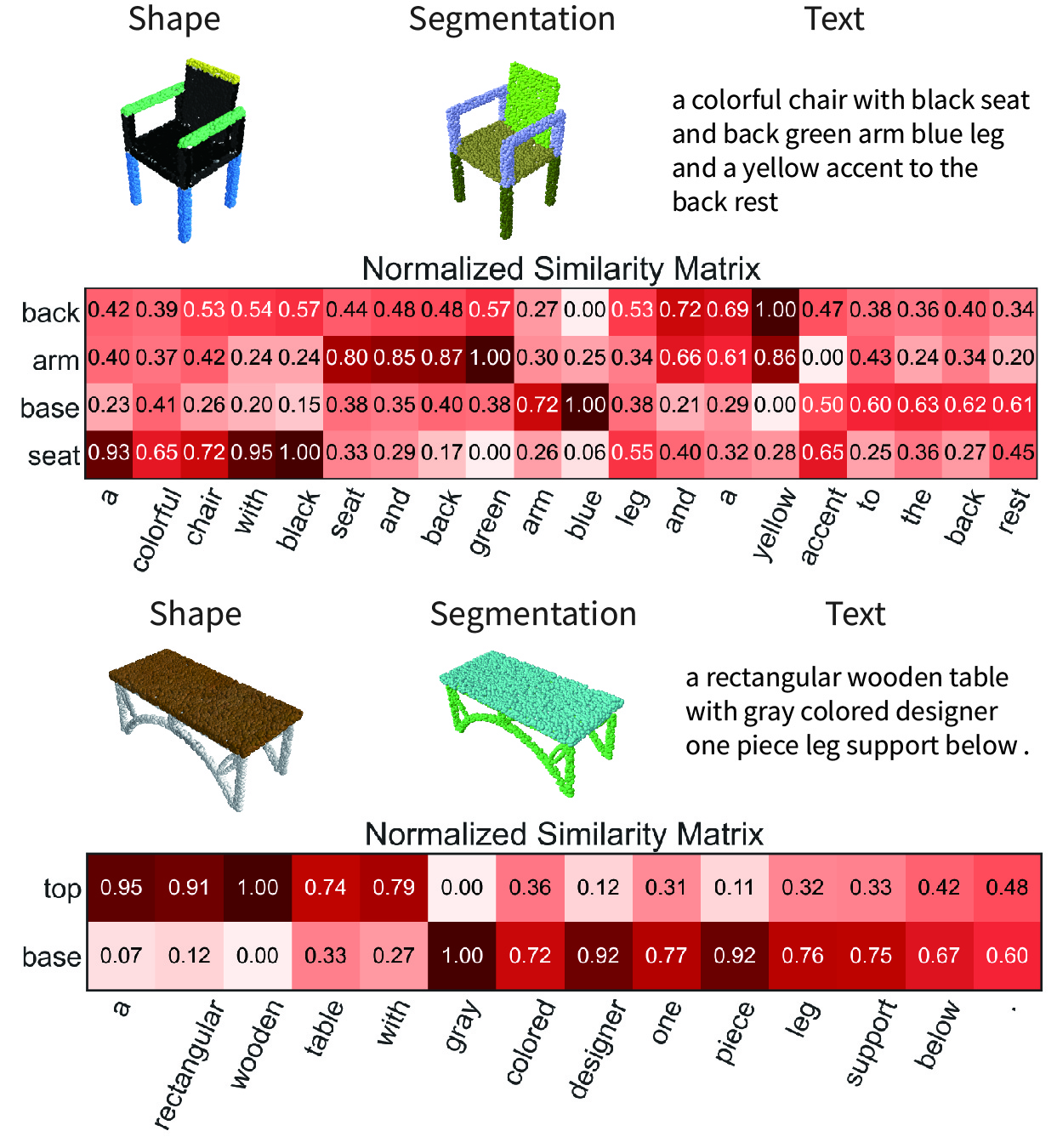}
    
    \caption{Visualization of the similarity between parts and words. We display a normalized similarity matrix between parts and words. The place with darker color indicates the higher relationship between corresponding parts and words.}
    \label{fig:att_vis}
    \vspace{-0.3cm}
\end{figure}

We visualized two examples of the best matching pair between the shape and the text, as shown in Figure~\ref{fig:att_vis}. The normalized similarity matrix is colored by calculating each pairwise distance $D=\{1-c_{ij}\}$ between parts and words. We can see that our model can accurately find the correspondence (dark red) between parts and words. For the first example, the chair is firstly segmented into 4 parts, then the black seat matches the words "black" and "seat" in the text well, and the rest also attends the words "yellow", "black" and "rest". Besides, the similarity weight between the part of blue legs and the word "blue" obtained the highest score.

\section{Conclusion}

We introduce a method to learn the joint embedding of 3D point clouds and text. Our method successively increases the ability of a joint understanding of 3D point clouds and text by learning to bidirectionally match 3D parts to words in an optimized space. We obtain the 3D parts by leveraging a 3D segmentation prior, which effectively resolves the self-occlusion issue of parts that is suffered by current multi-view based methods. We also demonstrate that matching 3D parts to words using the optimal transport is an efficient way to merge different modalities including 3D shapes and text in a common space, where the proposed cross-modal earth mover's distance is also justified to effectively capture the relationship of part-word in this matching procedure. Experimental results show that our method significantly outperforms other state-of-the-art methods.

\vspace{-0.5cm}
\paragraph{Acknowledgement.}
This work is supported by the Young Scientists Fund of the National Natural Science Foundation of China (Grant No.62206106). 

{\small
\bibliographystyle{ieee_fullname}
\bibliography{egbib}
}

\end{document}